\documentclass{article}

\usepackage{PRIMEarxiv}

\usepackage[utf8]{inputenc} 
\usepackage[T1]{fontenc}    
\usepackage{hyperref}       
\usepackage{url}            
\usepackage{booktabs}       
\usepackage{amsfonts}       
\usepackage{nicefrac}       
\usepackage{microtype}      
\usepackage{lipsum}
\usepackage{fancyhdr}       
\usepackage{graphicx}       
\graphicspath{{media/}}     
\usepackage{natbib}
\usepackage{multirow}%
\usepackage{amsmath,amssymb,amsfonts}%
\usepackage{amsthm}%
\usepackage{mathrsfs}%
\usepackage[title]{appendix}%
\usepackage{xcolor}%
\usepackage{textcomp}%
\usepackage{manyfoot}%
\usepackage{booktabs}%
\usepackage{algorithm}%
\usepackage{algorithmicx}%
\usepackage{algpseudocode}%
\usepackage{listings}%
\usepackage{hyperref}
\usepackage{url}

\usepackage{booktabs}
\usepackage{tabularx}
\usepackage{array}
\usepackage{graphicx}
\usepackage{pifont}
\usepackage[most]{tcolorbox}

\newcommand{\cmark}{\ding{51}}
\newcommand{\xmark}{\ding{55}}

\usepackage{caption}
\usepackage{subcaption}
\usepackage{mathtools}

\usepackage{multicol}
\usepackage{multirow}
\usepackage{makecell}
\definecolor{lightgrey}{gray}{0.9}
\usepackage{xcolor}
\usepackage{amssymb}
\usepackage{bbding}
\usepackage{colortbl}
\definecolor{minetable1colorx}{rgb}{0.75, 0.75, 0.75}

\usepackage{algorithm}
\usepackage{algorithmicx}
\usepackage{algpseudocode}
\usepackage{enumitem}
\providecommand{\pmark}{\raisebox{-0.1ex}{$\triangle$}}
\definecolor{BestBG}   {HTML}{FFCDCB}
\definecolor{SecondBG} {HTML}{FFE3CF}
\definecolor{ThirdBG}  {HTML}{CAE5FF}

\newcommand{\best}[1]{\cellcolor{BestBG}\textbf{#1}}
\newcommand{\second}[1]{\cellcolor{SecondBG}{#1}}
\newcommand{\third}[1]{\cellcolor{ThirdBG}{#1}}

\newcommand{\redregion}[1]{%
  \begingroup\setlength{\fboxsep}{0.2ex}\colorbox{BestBG}{#1}\endgroup}
\newcommand{\orangeregion}[1]{%
  \begingroup\setlength{\fboxsep}{0.2ex}\colorbox{SecondBG}{#1}\endgroup}
\newcommand{\blueregion}[1]{%
  \begingroup\setlength{\fboxsep}{0.2ex}\colorbox{ThirdBG}{#1}\endgroup}

\title{VehDyn: A Driving World Model Benchmark for Vehicle Dynamics
}

\author{
  Tianyi Wang$^*$, Wangsheng Du$^*$, Jiazhou Chen, Tianyi Zeng, Xiangyu Li, Jiseop Byeon, Yujin Wang, Yiming Xu, \\ \And Yangyang Wang, Bingzhao Gao, Sikai Chen, Zhaomiao Guo, Junfeng Jiao, Christian Claudel$^\dag$, Alexandre Bayen\\
  $^*$Equal Contribution\\
  $^\dag$Corresponding Author\\
}

\begin{document}
\maketitle

\begin{abstract}
Video world models are emerging as data engines, action planners, and generative simulators for autonomous driving, but existing benchmarks primarily assess visual fidelity and coarse physical plausibility, providing limited evidence on whether generated driving futures obey realistic vehicle kinematics and dynamics.
This limitation is further compounded by the lack of datasets in which vehicle, road, maneuver, and speed conditions are independently controlled, and ground-truth vehicle states are recorded in synchrony with videos.
We introduce VehDyn, a driving world model benchmark for vehicle dynamics.
VehDyn is built on a CARLA-CarSim co-simulation platform where photorealistic rendering is coupled with a validated multi-body dynamics model, and it contains 10,080 configurations from a full factorial design over five vehicle types, four tire-road friction coefficients, three maneuvers, four target speeds, 14 scenes, and three illuminations, each paired with synchronized position, velocity, and attitude sequences.
Built on this dataset, VehDyn introduces a hierarchical evaluation framework that measures trajectory alignment, kinematic consistency, and dynamic consistency, and benchmarks 12 state-of-the-art video world models. 
We further assess the video quality using two established protocols and correlate it with the VehDyn score.
Trajectory-level metrics are nearly saturated, with ten of twelve models within 20\% of ground truth, while no model reaches 92\% of ground truth on dynamic consistency, and visual-quality metrics are only weakly correlated with vehicle-dynamics fidelity. 
DrivingWorld achieves the highest VehDyn score, followed by Cosmos 3 Nano and LTX-Video 2.5, and the VehDyn score agrees closely with human judgment.
VehDyn provides a systematic foundation for developing driving world models that are physically consistent and visually realistic.
The dataset and evaluation metrics will be released after acceptance.
\end{abstract}

\section{Introduction}
\label{sec:introduction}

A video world model is a generative simulator that predicts how an environment will evolve under different action conditions \citep{ha2018recurrent}, and supports downstream tasks \citep{zhao2026edge}.
Inspired by this vision, there has been great progress in driving world models: given initial observations and optional conditions such as text prompts, control sequences, or target trajectories, driving world models predict the future evolution of the ego vehicle and its surroundings \citep{gao2024vista}.
The value of a generated future depends on whether the ego vehicle moves as a real vehicle would under the same commands and road conditions.
That motion is governed by tire-road adhesion, inertia, and suspension compliance \citep{wang2026dbpnet}, which bound the achievable acceleration, determine the stopping distance, and couple longitudinal and lateral inputs to the pitch and roll of the body. 
A generated lane change may follow a plausible planar path while exhibiting a lateral acceleration beyond the available adhesion, and an emergency-braking sequence may appear smooth while producing an incorrect stopping distance, pitch response, or deceleration profile.
Such failures are visually unremarkable but physically invalid, and any planner trained or evaluated on such futures inherits the error. 
Prior studies \citep{zhang2025worldinworld,fan2026wow} have found visual realism to be largely uncorrelated with physical understanding.
Therefore, measuring vehicle-dynamics fidelity in driving video generation represents a prerequisite for deploying driving world models as reliable simulators, especially in safety-critical scenarios. 

Existing evaluation of driving world models has progressively moved beyond conventional video quality metrics \citep{wang2026robophys}, but vehicle-dynamics fidelity remains largely unmeasured. 
Early work relied primarily on Fréchet inception distance (FID), Fréchet video distance (FVD), and downstream perception or planning performance \citep{wu2026omniworldbench,ying2026wbench,shang2026worldarena}. 
ACT-Bench \citep{arai2024act} introduced action fidelity by recovering the ego trajectory from generated video with monocular visual odometry and comparing it with the commanded trajectory through action-match rate, average displacement error (ADE), and final displacement error (FDE).
DrivingGen \citep{zhou2026drivinggen} broadened the scope to distribution realism, driving-specific image quality, temporal consistency, and a trajectory-consistency score that rewards smooth velocity and acceleration.
WorldLens \citep{liang2026worldlens} added reconstruction quality, action following, downstream utility, and human-preference assessment, and What-If World \citep{cai2026if} tested whether a model's output changes in the direction physics predicts when a single variable in the prompt is altered.
Consequently, no prior benchmark compares generated motion with synchronized ground-truth vehicle states in both kinematic and dynamic aspects.

Existing benchmarks have introduced different datasets covering diverse and challenging scenarios. 
Real-world datasets such as nuScenes \citep{caesar2020nuscenes}, nuPlan \citep{caesar2021nuplan}, NAVSIM \citep{dauner2024navsim}, DrivingDojo \citep{wang2024drivingdojo}, and StyleDrive \citep{hao2026styledrive} provided rich traffic diversity and useful pose or trajectory annotations, but their physical conditions are observational rather than experimentally controlled.
Simulation-based datasets \citep{dosovitskiy2017carla,li2022metadrive} offered greater controllability, but they were primarily designed for closed-loop policy evaluation and rely on a game-engine vehicle model that is not intended as a dynamics reference. 
High-fidelity simulators such as CarSim model tire forces, suspension kinematics, and load transfer with validated multi-body formulations and are the standard tool in vehicle dynamics analysis \citep{zeng2023wheel,zeng2026damper}, but they have not been coupled with photorealistic rendering to construct evaluation data for video world models.
In all, current resources (Table \ref{tab:overall_comparison}) offered either rich visual diversity or controlled driving tasks, but not a physics-grounded testbed in which generated videos can be compared with synchronized vehicle-dynamics ground truth under systematically varied conditions.
A detailed literature review is provided in Appendix \ref{appendix:literature}.

\begin{table*}[t!]
\centering
\caption{Comparison of existing driving world model benchmarks and autonomous driving datasets. \cmark: supported; \pmark: partial; \xmark: not supported.}
\label{tab:overall_comparison}
\setlength{\tabcolsep}{4pt}
\renewcommand{\arraystretch}{1.08}
\small
(a) Comparison of evaluation dimensions in driving world model benchmarks. 
\vspace{6pt}
\resizebox{\textwidth}{!}{%
\begin{tabular}{lcccccccc}
\toprule
\multirow{2}{*}{\textbf{Benchmark}}
& \multirow{2}{*}{\textbf{Venue \& Year}}
& \multirow{2}{*}{\textbf{Data}}
& \multicolumn{2}{c}{\textbf{World Model Types}}
& \multicolumn{3}{c}{\textbf{Vehicle Motion Evaluation}}
& \multirow{2}{*}{\textbf{Human}} \\
\cmidrule(lr){4-5} \cmidrule(lr){6-8}
& & & \textbf{General} & \textbf{Driving}
& \textbf{Trajectory} & \textbf{Kinematics} & \textbf{Dynamics} & \\
\midrule
ACT-Bench
& ICLRW '25
& nuScenes
& \xmark
& \cmark
& \cmark
& \xmark
& \xmark
& \xmark \\
WorldSimBench
& ICML '25
& CARLA
& \cmark
& \xmark
& \pmark
& \xmark
& \xmark
& \cmark \\
DrivingGen
& ICLR '26
& Real-World
& \cmark
& \cmark
& \cmark
& \pmark
& \xmark
& \cmark \\
WorldLens
& CVPR '26
& nuScenes
& \xmark
& \cmark
& \cmark
& \pmark
& \pmark
& \cmark \\
What-If World
& arXiv '26
& nuScenes
& \cmark
& \xmark
& \xmark
& \xmark
& \pmark
& \cmark \\
\midrule
\textbf{VehDyn (Ours)}
& --
& CARLA-CarSim
& \cmark
& \cmark
& \cmark
& \cmark
& \cmark
& \cmark \\
\bottomrule
\end{tabular}%
}
(b) Comparison of data controllability in autonomous driving datasets.
\vspace{2pt}
\resizebox{\textwidth}{!}{%
\begin{tabular}{lcccccccc}
\toprule
\multirow{2}{*}{\textbf{Dataset}}
& \multirow{2}{*}{\textbf{Venue \& Year}}
& \multirow{2}{*}{\textbf{Visual Ctrl.}}
& \multicolumn{4}{c}{\textbf{Controlled Dynamics Factors}}
& \multirow{2}{*}{\textbf{Dynamics GT}}
& \multirow{2}{*}{\textbf{Paired Analysis}} \\
\cmidrule(lr){4-7}
& & & \textbf{Vehicle} & \textbf{Speed} & \textbf{Maneuver} & \textbf{Friction} & & \\
\midrule
\multicolumn{9}{l}{\textit{Real-World Datasets}} \\
nuScenes
& CVPR '20
& \pmark
& \xmark
& \pmark
& \xmark
& \xmark
& \pmark
& \xmark \\
nuPlan
& CVPRW '21
& \pmark
& \xmark
& \pmark
& \pmark
& \xmark
& \pmark
& \xmark \\
NAVSIM
& NeurIPS '24
& \pmark
& \xmark
& \pmark
& \pmark
& \xmark
& \pmark
& \xmark \\
DrivingDojo
& NeurIPS '24
& \pmark
& \xmark
& \pmark
& \pmark
& \xmark
& \pmark
& \xmark \\
StyleDrive
& AAAI '26
& \pmark
& \xmark
& \pmark
& \xmark
& \xmark
& \pmark
& \xmark \\
DrivingGen
& ICLR '26
& \pmark
& \xmark
& \pmark
& \pmark
& \xmark
& \pmark
& \xmark \\
\midrule
\multicolumn{9}{l}{\textit{CARLA-Based Datasets}} \\
CARLA (Leaderboard)
& CoRL '17
& \cmark
& \xmark
& \pmark
& \cmark
& \xmark
& \pmark
& \xmark \\
Bench2Drive
& NeurIPS '24
& \cmark
& \xmark
& \pmark
& \cmark
& \xmark
& \pmark
& \xmark \\
Bench2Drive-Speed
& arXiv '26
& \cmark
& \xmark
& \cmark
& \cmark
& \xmark
& \pmark
& \pmark \\
Bench2Drive-Robust
& arXiv '26
& \cmark
& \xmark
& \pmark
& \cmark
& \xmark
& \pmark
& \pmark \\
Fail2Drive
& arXiv '26
& \cmark
& \xmark
& \pmark
& \cmark
& \xmark
& \pmark
& \pmark \\
\midrule
\textbf{VehDyn (Ours)}
& --
& \cmark
& \cmark
& \cmark
& \cmark
& \cmark
& \cmark
& \cmark \\
\bottomrule
\end{tabular}%
}
\end{table*}

\begin{figure*}[t!]
  \centering
  \includegraphics[width=\linewidth]{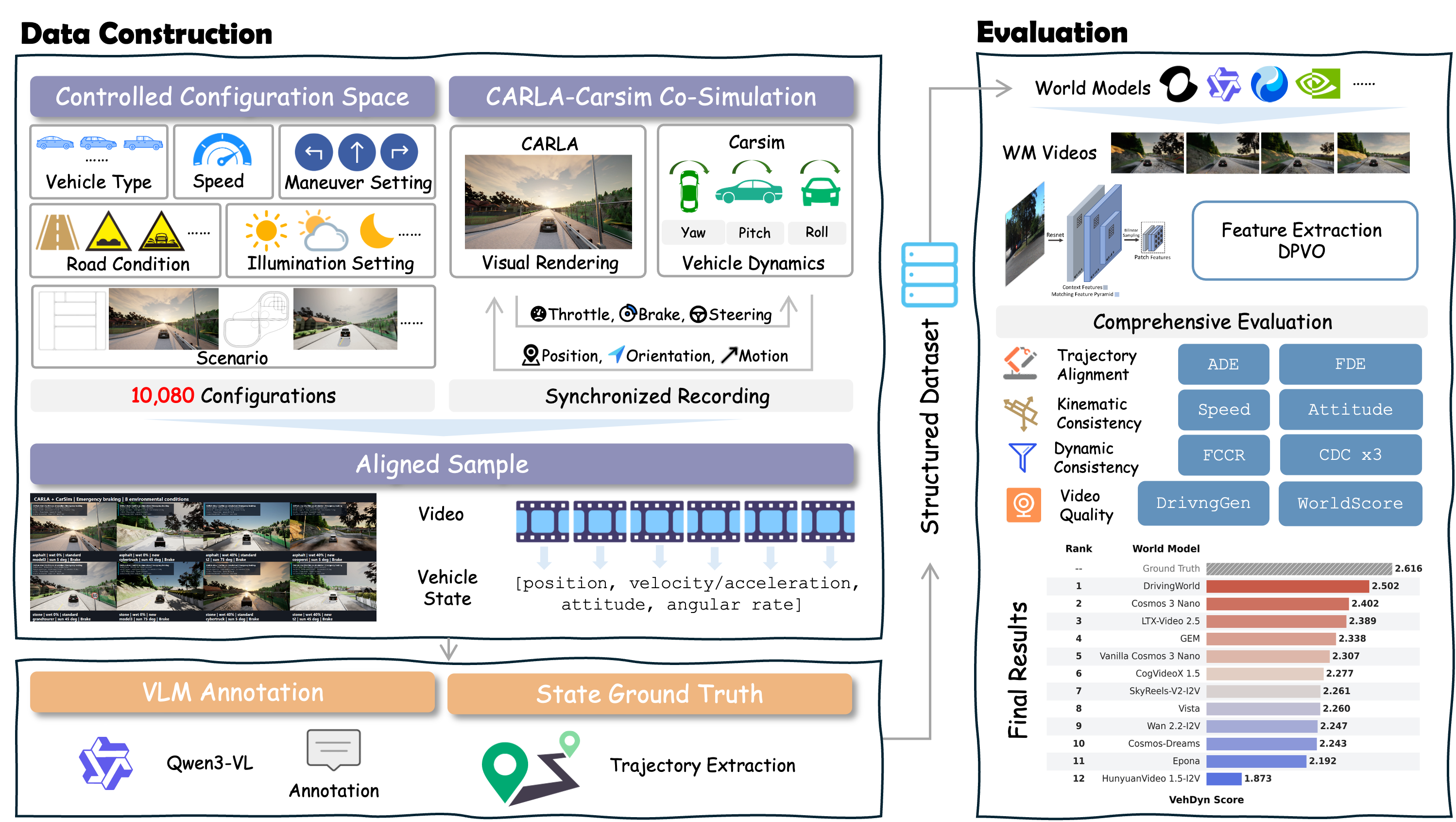}
  \caption{Overview of the proposed VehDyn benchmark. Left: data construction. A full factorial design over vehicle type, target speed, maneuver, road condition, scene, and illumination is instantiated on a CARLA-CarSim co-simulation platform. Right: evaluation. Vehicle states are recovered from the generated videos by DPVO and compared with ground truth at three levels, trajectory alignment, kinematic consistency, and dynamic consistency, and aggregated into the VehDyn score.
  }
  \label{fig1}
\end{figure*}

To address this gap, we propose \textbf{VehDyn} (Figure \ref{fig1}), a comprehensive driving world model benchmark for evaluating vehicle dynamics with diverse data distributions and novel evaluation metrics.
\textbf{VehDyn} is built on a CARLA-CarSim co-simulation platform in which CARLA provides visual rendering, while CarSim computes the vehicle response with a high-fidelity multi-body model, with both simulators synchronized on a unified clock.
The dataset comprises 10,080 configurations generated from a full factorial design over five vehicle types, four tire-road friction coefficients, three maneuvers, four target speeds, 14 scenes, and three sun angles, each paired with ground-truth position, velocity, and attitude sequences. 
Built on this dataset, \textbf{VehDyn} defines a hierarchical evaluation protocol comprising trajectory alignment, kinematic consistency, and dynamic consistency. 
We benchmark 12 video world models, additionally score every generated video under the DrivingGen \citep{zhou2026drivinggen} and WorldScore \citep{duan2025worldscore} protocols, and verify the agreement between the VehDyn score and human judgment.
The main contributions are summarized as follows:
\begin{itemize}
\item \textbf{A physics-grounded dataset}, which includes 10,080 CARLA-CarSim co-simulation configurations generated from a full factorial design over vehicle type, maneuver, speed, friction coefficient, scene, and sun angle, paired with synchronized ground-truth vehicle states.
\item \textbf{A vehicle-dynamics-aware evaluation protocol}, which extends beyond visual plausibility and planar trajectory alignment to six-degree-of-freedom (6-DoF) kinematic consistency, and condition-aware dynamic consistency measured against ground-truth responses.
\item \textbf{An extensive benchmark}, which evaluates 12 representative video world models, characterizing the gap between visual fidelity and vehicle-dynamics consistency, and identifies failure modes that visual metrics do not expose.
\end{itemize}

\section{VehDyn Dataset}
\label{sec:dataset}

\subsection{Data Construction}

VehDyn is constructed in three stages (Figure \ref{fig1}). 
We first build a CARLA-CarSim co-simulation platform in which visual rendering and vehicle dynamics are computed by separate simulators on a shared clock. 
We then define a controllable scenario space over six factors and instantiate every combination in a full factorial design. 
For each instance, we synchronously record a video clip, the complete vehicle state sequence, and the scenario configuration, and we finally annotate each clip with a motion-centric caption. 
Each configuration contains ten independently generated episodes.
The resulting dataset contains 100,800 samples.

\paragraph{Co-simulation platform.} 

CARLA renders road geometry, scene assets, traffic participants, and lighting, while CarSim computes the vehicle response with a validated multi-body model that includes tire force generation, suspension kinematics, and longitudinal and lateral load transfer. 
At each simulation step, the reference-path controller in CARLA issues throttle, brake, and steering commands; CarSim integrates the vehicle dynamics under the configured vehicle and road parameters and returns the updated position, orientation, and motion states; CARLA then places the vehicle at the returned pose and renders the next frame. 
CarSim integrates at 2,000 Hz and CARLA renders at 30 Hz in synchronous fixed-step mode, with the CarSim state sampled at every rendered frame, so that each frame corresponds to an exactly known dynamic state.
Through this closed-loop interaction, visual observations and high-fidelity vehicle motion states are collected synchronously.

\paragraph{Scenario space.} 

The scenario space comprises six factors (Table \ref{tab:dataset_item_type} in Appendix \ref{appendix:dataset}). 
Four dynamic factors define the physical conditions to which a world model must respond (vehicle type, target speed, driving maneuver, and tire-road friction coefficient), and the two visual factors provide replicates under which the dynamic response should be invariant (urban scene and sun angle). 
Vehicle appearance, scene, and sun angle are configured at CARLA initialization, the target speed is imposed by the path-following controller, and driving maneuvers are realized as predefined reference paths or control sequences. 
Vehicle mass, moment of inertia, and wheelbase are specified in CarSim.
In addition, road adhesion conditions are kept semantically consistent: CARLA visually represents the road conditions through road material and surface wetness, and CarSim applies the corresponding tire-road friction coefficient.
Route details are given in Figure \ref{fig:route_diversity} in Appendix \ref{appendix:dataset}.

\paragraph{Data acquisition.} 

Visual observations are recorded at 30 Hz for 5 s at \(1920 \times 1080\) resolution from an RGB camera rigidly mounted at the vehicle reference point and moving with the body. 
This third-person viewpoint is chosen in preference to a front-facing dashcam view for two reasons.
First, the vehicle body is visible in the frame, so pitch, yaw, and roll are directly observable in the generated video. 
Second, because the camera is rigidly attached to the body, its 6-DoF pose differs from the vehicle pose only by a fixed transform, which allows vehicle attitude to be recovered from camera-pose estimates. 
Vehicle states are recorded from CarSim at every rendered frame and include position, orientation, velocity and rate, control inputs, and progress along the route. 
All modalities are timestamped on the shared simulation clock.

\paragraph{Scene annotation.} 

Each clip is captioned with Qwen3-VL \citep{bai2025qwen3} under a prompt that foregrounds visually observable motion rather than static appearance or abstract physical laws. 
Because the conditioning frame already conveys scene layout, lighting, and vehicle identity, the caption is allocated to the temporal dynamics that a generative model must synthesize: the maneuver, its onset, direction, and extent, the resulting change in vehicle attitude, and the camera's rigid attachment to the vehicle. 
A manual review pass by three annotators discards captions containing hallucinated or physically ambiguous content. 
The full prompt is given in Appendix \ref{appendix:dataset}.

\begin{figure*}[t!]
    \centering
    \includegraphics[width=\textwidth]
    {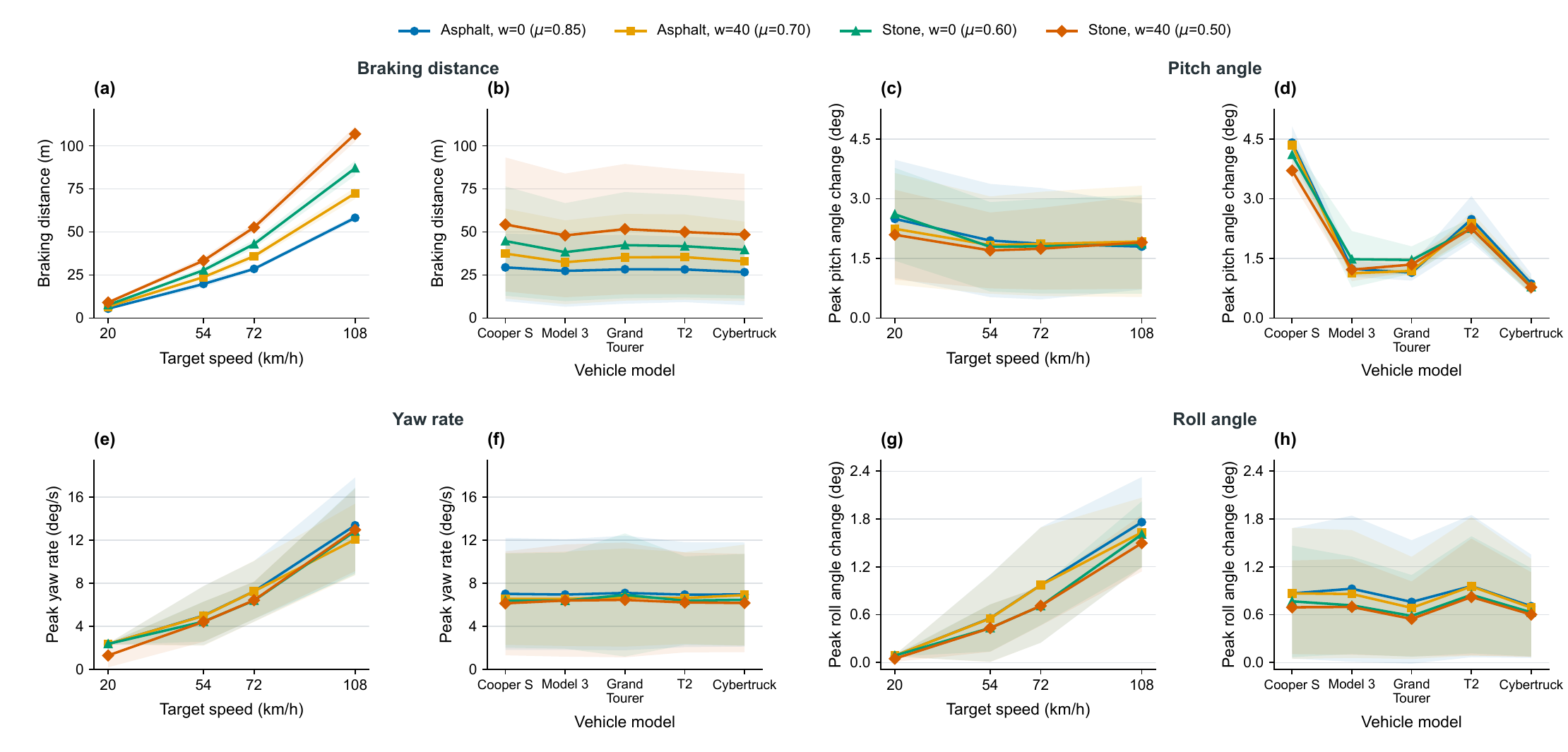}
    \caption{
        Dynamic characteristics of the VehDyn dataset under different target speeds, road conditions and vehicle types. 
        (a)--(d), emergency-braking condition; (e)--(h), single-lane-changing condition.
        Curves represent the mean values, and shaded bands denote $\pm1$ standard deviation across scenes and illumination settings.
    }
    \label{fig2:dynamic_characteristics}
\end{figure*}

\subsection{Data Analysis}

\paragraph{Dataset composition.} 

Each of the 100,800 samples consists of a 5 s, 150-frame RGB clip, a synchronized ground-truth state sequence, the scenario configuration, and a caption. 
The dataset totals 140 h of video and 2 TB.
Because the design is fully crossed, every level of every factor occurs with every level of the others, and any two samples that differ in a single factor form a matched pair. 
For each target-speed sweep, friction contrast, or vehicle contrast, the ground-truth change in dynamic response is available directly from the recorded states. 
Recorded variables, data sources, and acquisition settings are provided in Table \ref{tab:variable_definition} in Appendix \ref{appendix:dataset}.

\paragraph{Dynamic characteristics.} 

To verify that the factorial design induces physically meaningful variation, we analyze the ground-truth states across the dataset. 
Emergency-braking samples characterize longitudinal motion and pitch responses, while single-lane-changing samples characterize lateral motion, yaw, and roll responses. 
Figure \ref{fig2:dynamic_characteristics} shows the responses across target speeds, vehicle models, and the four road conditions.
In the emergency-braking scenario, braking distance increases nonlinearly with target speed, more markedly under lower-friction conditions, and increases as friction decreases for every vehicle model, while differences among models at fixed conditions remain small. 
Peak pitch-angle changes, by contrast, vary more across vehicle models than across target speeds, potentially reflecting differences in mass distribution and suspension characteristics.
In the single-lane-changing scenario, yaw-angle changes are relatively similar across conditions, whereas yaw rates are more sensitive to speed, making them more suitable for characterizing turning dynamics. 
Both the peak yaw rate and the peak roll-angle changes increase with target speed. 
Yaw rate is nearly identical across vehicle models, while roll-angle changes exhibit more pronounced differences among models.
The full dynamic characteristics analysis is provided in Appendix \ref{appendix:dataset}.

\section{VehDyn Evaluation}
\label{sec:evaluation}

VehDyn evaluates a generated video by recovering the ego-vehicle state sequence from the video and comparing it with the synchronized ground-truth states at three levels (Figure \ref{fig1}): (1) trajectory alignment, which measures planar position; (2) kinematic consistency, which measures translational and rotational motion states; and (3) dynamic consistency, which measures whether the generated motion is physically feasible and whether it responds to changed physical conditions as the real vehicle does.
Per-metric implementation details are provided in Appendix \ref{appendix:evaluation}.

\subsection{Feature Extraction}

We employ deep patch visual odometry (DPVO) \citep{teed2023deep} to estimate a sequence of 6-DoF relative camera poses $\{\hat{\mathbf{T}}_t\}_{t=1}^{T}$ from the video frames and the known camera intrinsics. 
DPVO tracks sparse image patches through recurrent correspondence updates and jointly refines camera poses by differentiable bundle adjustment. 
Because the camera is rigidly attached to the vehicle body, the body pose is obtained from the camera pose by the fixed extrinsic transform, so that roll, pitch, and yaw of the vehicle follow directly from the estimated camera rotation.
A comparison with trajectory-reconstruction methods adopted by other benchmarks is provided in Appendix \ref{appendix:quantitative}.

\subsection{Metric Design}

\paragraph{Trajectory alignment.}

Planar position is compared using ADE and FDE, computed over all maneuvers.
Lower ADE and FDE indicate better trajectory alignment.

\paragraph{Kinematic consistency.} 

Two trajectories with similar planar paths may differ substantially in velocity and attitude.
Therefore, we compare both the translational states (longitudinal and lateral velocity, $v_x$ and $v_y$), and the rotational states (roll, pitch, yaw) by root mean square error (RMSE) over the corresponding state sequence. 
Since RMSE penalizes large deviations more heavily than mean absolute error, it reflects not only overall accuracy, but also severe local inconsistencies.
All kinematic-consistency metrics are lower-is-better.

\paragraph{Dynamic consistency.} 

Trajectory and kinematic errors quantify agreement with one recorded outcome but do not test physical feasibility or causal sensitivity.
We employ the friction-circle constraint rate (FCCR), which uses the adhesion bound on the resultant planar acceleration and reports the fraction of active-maneuver frames in which the recovered acceleration satisfies the bound for the friction coefficient of the sample's road condition, within a fixed tolerance.
Then we construct samples under different physical conditions and perform cross-instance evaluation. 
We evaluate condition-aware dynamic consistency (CDC) from three aspects: response direction consistency (CDC$_{\mathrm{Dir}}$) measures whether the model correctly predicts the direction of vehicle-state changes induced by variations in physical conditions, response magnitude consistency (CDC$_{\mathrm{Mag}}$) evaluates the accuracy with which the model captures the strength of such response changes, and overall trend consistency (CDC$_{\mathrm{Trend}}$) measures whether, as a continuous physical condition varies across multiple levels, the model response follows the same variation pattern as the real vehicle. 

\paragraph{VehDyn score.}

Each metric $m$ is first mapped to a normalized score $s_m \in [0, 1]$ with higher-is-better orientation.
Metrics are then averaged within each level and the level scores are summed:
\begin{equation}
\mathrm{VehDyn} = \sum_{\ell \in \mathcal{L}} \underbrace{\frac{1}{|M_\ell|}\sum_{m\in M_\ell}\tilde{s}_m}_{\displaystyle S_\ell},
\label{eq:vehdyn}
\end{equation}
where $\mathcal{L}$ indexes the three levels, $M_\ell$ is the set of metrics in level $\ell$, and $S_\ell$ is the level score. 
Averaging within $M_\ell$ makes a level score independent of how many metrics instantiate it. 
Summing over $\mathcal{L}$ gives each level equal weight, so that the relative importance of trajectory, kinematics, and dynamics is an explicit design choice.
Scores are computed per maneuver family and macro-averaged with equal weight.
The normalization process is described in Appendix \ref{appendix:evaluation}.

\subsection{Complementary Visual Evaluation}

We also score every generated video under two established protocols without modification, i.e., DrivingGen \citep{zhou2026drivinggen} and WorldScore \citep{duan2025worldscore} to quantify the gap between visual and dynamic scores.

\paragraph{DrivingGen.}

We adopt Distribution (FVD) and Temporal Consistency (Video Consistency and Trajectory Consistency).
FVD measures clip-level distributional realism between inflated 3D ConvNet \citep{carreira2017quo} feature distributions of generated and ground-truth videos. 
Video Consistency measures frame-to-frame DINOv3 \citep{simeoni2025dinov3} feature similarity after motion-adaptive temporal subsampling with an optical-flow model \citep{wang2024sea}.
Trajectory Consistency computes how stable a trajectory’s velocity and acceleration are over time. 

\begin{figure*}[t!]
  \centering
  \includegraphics[width=\linewidth]{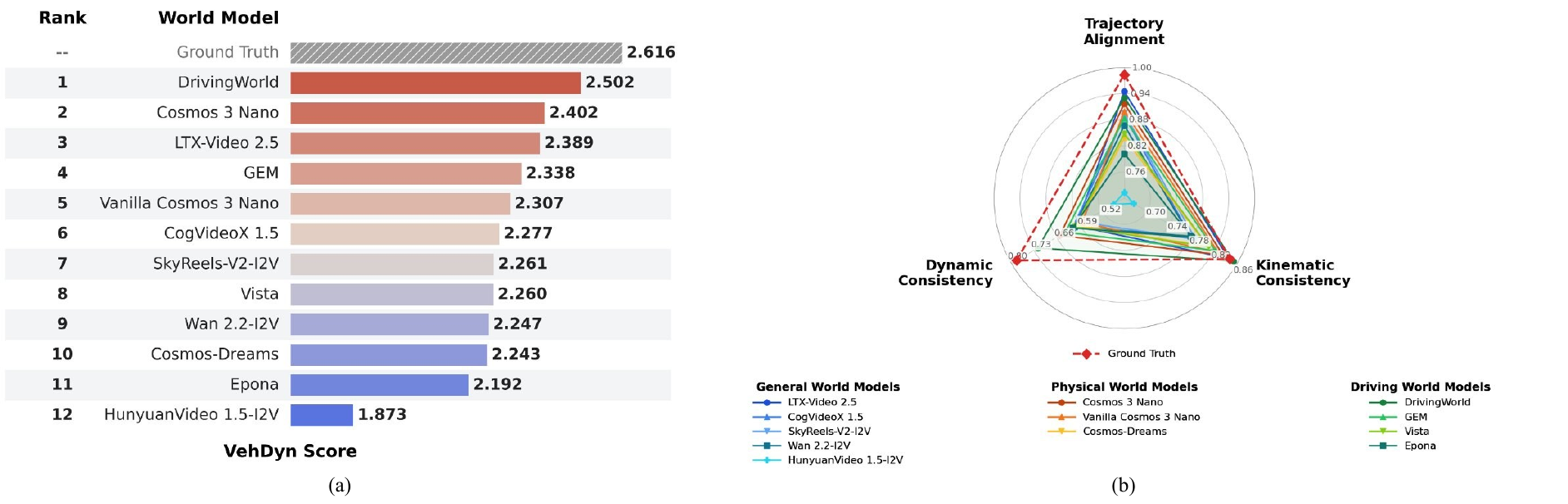}
  \caption{Overall results of the proposed VehDyn benchmark. (a) VehDyn score of the 12 evaluated models ordered by rank, with the ground-truth video scored through the same pipeline as the reference ceiling (dashed line). (b) Normalized level scores of the 12 evaluated models and ground truth.
  }
  \label{fig3}
\end{figure*}

\paragraph{WorldScore.}

We adopt Quality (3D Consistency, Photometric Consistency, Style Consistency, and Subjective Quality).
3D Consistency uses DROID-SLAM \citep{teed2021droid} to compute the reprojection error between co-visible pixels in consecutive frames. 
Photometric Consistency measures the average end-point error (AEPE) of optical flow between consecutive frames. 
Style Consistency measures Gram-matrix style drift \citep{gatys2015neural}.
Subjective Quality uses the arithmetic mean of CLIP-Aesthetic \citep{schuhmann2022clip+} and CLIP-IQA+ \citep{wang2023exploring}.

\begin{table*}[t!]
\caption{Detailed quantitative results of the VehDyn benchmark for 12 representative video world models. Top: VehDyn metrics. Rank is by VehDyn score. Bottom: Visual evaluation under the WorldScore and DrivingGen benchmarks. Rank is by the sum of the normalized level scores. Best results are in \redregion{red region}, second best in \orangeregion{orange region}, and third best in \blueregion{blue region}. Models fall into three categories: general world models, physical world models, and driving world models.}
\label{tab:evaluation_results}
\centering
\resizebox{\textwidth}{!}{%
\small
\setlength{\tabcolsep}{4pt}
\begin{tabular}{@{}l|*{2}{c}|*{5}{c}|*{4}{c}|c@{}}
\toprule
\multirow{3}{*}{\makecell[l]{VehDyn Metrics\\\\Models}}
& \multicolumn{2}{c|}{Trajectory Alignment} &
\multicolumn{5}{c|}{Kinematic Consistency} &
\multicolumn{4}{c|}{Dynamic Consistency} &
\multirow{3}{*}{\makecell[c]{Rank}}\\
\cmidrule(lr){2-3} \cmidrule(lr){4-8} \cmidrule(lr){9-12}
& \multicolumn{2}{c|}{All} & \multicolumn{2}{c}{Emergency-Braking} & \multicolumn{3}{c|}{Single-Lane-Changing} & \multicolumn{1}{c}{All} & \multicolumn{3}{c|}{All} & \\
\cmidrule(lr){2-3} \cmidrule(lr){4-5} \cmidrule(lr){6-8} \cmidrule(lr){9-9} \cmidrule(lr){10-12}
& ADE $\downarrow$ & FDE $\downarrow$ &
$v_x$ $\downarrow$ & Pitch $\downarrow$ &
$v_y$ $\downarrow$ & Roll $\downarrow$ & Yaw $\downarrow$ &
FCCR $\uparrow$ & CDC$_{\mathrm{Dir}}$ $\uparrow$ & CDC$_{\mathrm{Mag}}$ $\uparrow$ & CDC$_{\mathrm{Trend}}$ $\uparrow$ & \\
\midrule
\textcolor{gray}{Ground Truth} & \textcolor{gray}{0.514} & \textcolor{gray}{0.814} & \textcolor{gray}{0.897} & \textcolor{gray}{2.522} & \textcolor{gray}{0.697} & \textcolor{gray}{0.602} & \textcolor{gray}{3.885} & \textcolor{gray}{0.900} & \textcolor{gray}{0.722} & \textcolor{gray}{0.713} & \textcolor{gray}{0.803} & \textcolor{gray}{--}\\
\midrule
\multicolumn{13}{@{}l}{\textit{General World Models}} \\
LTX-Video 2.5 & \best{1.665} & \best{2.310} & 3.217 & \best{2.077} & \best{0.932} & \best{0.558} & 3.330 & \second{0.686} & 0.563 & \third{0.610} & 0.539 & 3 \\
CogVideoX 1.5 & 2.804 & 5.368 & 5.683 & 3.018 & \third{1.270} & 0.971 & 3.543 & \best{0.738} & 0.582 & 0.557 & 0.565 & 6 \\
SkyReels-V2-I2V & 2.853 & 4.757 & 7.243 & 2.415 & \second{1.127} & 0.828 & 3.291 & 0.569 & 0.577 & 0.584 & 0.590 & 7 \\
Wan 2.2-I2V & 3.620 & 5.902 & 9.063 & 2.161 & 1.552 & 0.658 & \third{3.283} & 0.595 & 0.571 & 0.602 & 0.643 & 9 \\
HunyuanVideo 1.5-I2V & 6.826 & 14.437 & 7.471 & 3.283 & 4.403 & 1.872 & 4.334 & 0.488 & 0.528 & 0.436 & 0.485 & 12 \\
\midrule
\multicolumn{13}{@{}l}{\textit{Physical World Models}} \\
Cosmos 3 Nano & \third{2.432} & \third{3.517} & \second{1.866} & 2.173 & 1.847 & 0.772 & 3.556 & 0.591 & \third{0.646} & \second{0.617} & 0.746 & 2 \\
Vanilla Cosmos 3 Nano & 2.736 & 4.476 & 4.415 & 2.203 & 1.288 & \third{0.590} & 3.335 & 0.601 & 0.581 & 0.599 & 0.543 & 5 \\
Cosmos-Dreams & 3.699 & 8.599 & 5.959 & 2.158 & 1.740 & \second{0.588} & 3.325 & 0.675 & 0.566 & 0.572 & 0.576 & 10 \\
\midrule
\multicolumn{13}{@{}l}{\textit{Driving World Models}} \\
DrivingWorld & \second{1.914} & \second{3.214} & \best{1.418} & \third{2.141} & 1.313 & 0.741 & \best{3.250} & \third{0.685} & \best{0.708} & \best{0.664} & \second{0.817} & 1 \\
GEM & 3.110 & 5.347 & 3.166 & 2.398 & 1.971 & 0.802 & \second{3.280} & 0.549 & 0.624 & 0.591 & \third{0.769} & 4 \\
Vista & 4.176 & 6.663 & 3.653 & 2.282 & 2.328 & 0.869 & 3.476 & 0.533 & 0.604 & 0.568 & 0.706 & 8 \\
Epona & 5.775 & 7.805 & \third{2.474} & \second{2.127} & 5.269 & 1.275 & 3.624 & 0.394 & \second{0.672} & 0.528 & \best{0.844} & 11 \\
\bottomrule
\end{tabular}%
}
\vspace{4pt}
\resizebox{\textwidth}{!}{%
\small
\setlength{\tabcolsep}{9.5pt}
\begin{tabular}{@{}l|*{4}{c}|*{3}{c}|c@{}}
\toprule
\multirow{3}{*}{\makecell[l]{Other Benchmarks\\\\Models}}
& \multicolumn{4}{c|}{WorldScore} &
\multicolumn{3}{c|}{DrivingGen} &
\multirow{3}{*}{\makecell[c]{Rank}}\\
\cmidrule(lr){2-5} \cmidrule(lr){6-8}
& \multicolumn{4}{c|}{Quality} & \multicolumn{1}{c}{Distribution} & \multicolumn{2}{c|}{Temporal Consistency} & \\
\cmidrule(lr){2-5} \cmidrule(lr){6-6} \cmidrule(lr){7-8}
& \makecell{3D\\Consistency $\uparrow$} & \makecell{Photometric\\Consistency $\uparrow$} & \makecell{Style\\Consistency $\uparrow$} & \makecell{Subjective\\Quality $\uparrow$} &
\makecell{FVD $\downarrow$} & \makecell{Video\\Consistency $\uparrow$} & \makecell{Trajectory\\Consistency $\uparrow$} & \\
\midrule
\textcolor{gray}{Ground Truth} & \textcolor{gray}{79.32} & \textcolor{gray}{19.88} & \textcolor{gray}{94.71} & \textcolor{gray}{42.84} & \textcolor{gray}{206.0} & \textcolor{gray}{0.9102} & \textcolor{gray}{0.4469} & \textcolor{gray}{--}\\
\midrule
\multicolumn{9}{@{}l}{\textit{General World Models}} \\
LTX-Video 2.5 & \best{75.51} & \second{27.36} & {94.46} & {29.99} & \best{224.0} & \third{0.9156} & {0.4709} & 1\\
SkyReels-V2-I2V & \second{71.90} & \best{60.36} & {88.78} & \second{40.85} & {277.2} & {0.9094} & \best{0.5130} & 2\\
Wan 2.2-I2V & {66.37} & \third{18.58} & {95.55} & {38.71} & \third{243.8} & \second{0.9206} & {0.3936} & 3\\
HunyuanVideo 1.5-I2V & {58.39} & {16.01} & {94.50} & {33.43} & {316.0} & {0.8771} & {0.3726} & 6\\
CogVideoX 1.5 & {58.33} & {0.00} & {94.86} & {31.90} & {559.2} & {0.9030} & \second{0.4752} & 8\\
\midrule
\multicolumn{9}{@{}l}{\textit{Physical World Models}} \\
Cosmos 3 Nano & \third{67.61} & {9.92} & \second{96.17} & \third{39.40} & {286.8} & {0.8742} & \third{0.4744} & 4\\
Cosmos-Dreams & {66.23} & {12.51} & {76.28} & {5.78} & \second{233.8} & {0.7747} & {0.3981} & 5\\
Vanilla Cosmos 3 Nano & {64.45} & {7.52} & {95.62} & \best{40.92} & {372.3} & {0.9125} & {0.4374} & 7\\
\midrule
\multicolumn{9}{@{}l}{\textit{Driving World Models}} \\
DrivingWorld & {50.79} & {14.81} & \best{99.32} & {10.08} & {506.9} & {0.8982} & {0.3637} & 9\\
Epona & {48.94} & {0.72} & \third{96.05} & {33.00} & {599.4} & {0.9126} & {0.3701} & 10\\
GEM & {54.73} & {7.39} & {72.92} & {16.62} & {599.8} & {0.8958} & {0.4138} & 11\\
Vista & {49.16} & {5.96} & {86.14} & {22.13} & {772.2} & \best{0.9281} & {0.4475} & 12\\
\bottomrule
\end{tabular}%
}
\end{table*}

\section{Experiments}

\subsection{Experimental Setup}

We evaluate 12 open-source video world models in three categories: five general world models (Wan 2.2-I2V \citep{wan2025wan}, LTX-Video 2.5 \citep{hacohen2025ltx}, SkyReels-V2-I2V \citep{chen2025skyreels}, HunyuanVideo 1.5-I2V \citep{wu2025hunyuanvideo}, and CogVideoX 1.5 \citep{yang2025cogvideox}), three physical world models (Cosmos 3 Nano \citep{agarwal2026cosmos}, Vanilla Cosmos 3 Nano \citep{agarwal2026cosmos}, and Cosmos-Dreams \citep{basant2026nvidia}), and four driving world models (Vista \citep{gao2024vista}, DrivingWorld \citep{hu2026drivingworld}, GEM \citep{hassan2025gem}, and Epona \citep{zhang2025epona}). 
Their architectures, conditioning, and inference costs are summarized in Table \ref{tab5} in Appendix \ref{appendix:evaluation}.
All models except Vanilla Cosmos 3 Nano, which is evaluated zero-shot to isolate the effect of post-training, are post-trained on the VehDyn dataset following their official implementations on eight A100 80 GB GPUs. 
Episodes are split sequence-wise into training, validation, and test sets at an 80\%/10\%/10\% ratio.
General and physical world models predict future frames from the initial observation and the caption, while driving world models from the initial observation and the trajectories/poses.
Full quantitative and qualitative results are given in Appendices \ref{appendix:quantitative} and \ref{appendix:qualitative}.

\begin{figure*}[t!]
  \centering
  \includegraphics[width=\linewidth]{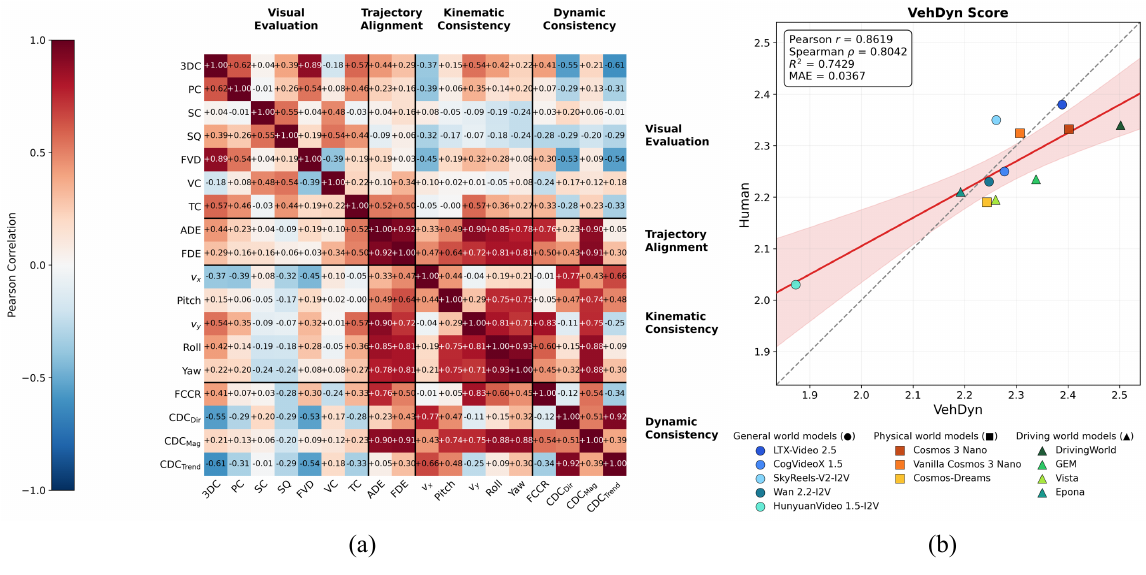}
  \caption{Cross-dimension correlation analysis. (a) Pearson correlation among the 18 metrics, with black lines separating the visual, trajectory, kinematic, and dynamic groups. (b) Pearson correlation between VehDyn score and human evaluation.}
  \label{fig4}
\end{figure*}

\subsection{Quantitative Results}

\paragraph{Per-Dimension Analysis.}

Overall VehDyn results are reported in Figure \ref{fig3} and Table \ref{tab:evaluation_results}, with per-level normalized scores in Tables \ref{tab:normalized_trajectory_alignment}--\ref{tab:vehicle_dynamic_score} in Appendix \ref{appendix:quantitative}.
DrivingWorld achieves the strongest performance among the evaluated models, with a VehDyn score of $2.502$ ($95.6\%$ of Ground Truth), followed by Cosmos 3 Nano ($2.402$), LTX-Video 2.5 ($2.389$), and GEM ($2.338$), with HunyuanVideo 1.5-I2V ranking last ($1.873$).
Post-training helps Cosmos 3 Nano exceed its zero-shot base by $0.095$, with the gain concentrated in dynamic consistency.
\textit{Trajectory alignment} is nearly saturated: ten of the twelve models fall within $20\%$ of Ground Truth. 
The top three are: LTX-Video 2.5 at $0.945$, DrivingWorld at $0.931$, and Cosmos 3 Nano at $0.917$.
\textit{Kinematic consistency} is comparatively insensitive to model choice at the level score, where DrivingWorld ($0.853$) exceeds Ground Truth ($0.847$), but the per-state errors separate the models: longitudinal velocity spans an order of magnitude (DrivingWorld $1.418$ m/s vs. Wan 2.2-I2V $9.063$ m/s), whereas yaw varies only between $0.571$ and $0.682$.
\textit{Dynamic consistency} shows that DrivingWorld retains $91.6\%$ of Ground Truth, Cosmos 3 Nano and GEM retain $82.9\%$ and $80.7\%$, and every other model retains $78\%$ or less.
CogVideoX 1.5 attains the highest friction-circle compliance ($0.738$) but ranks sixth on dynamics, and LTX-Video 2.5, second in compliance ($0.686$), ranks eighth, indicating that feasible motion does not imply a correct response to changed friction or speed.
Conversely, Epona has the highest trend consistency ($0.844$) and the second-highest direction consistency ($0.672$) but the lowest compliance ($0.394$), so its responses have the correct ordering while exceeding the adhesion limit.
Only DrivingWorld is strong on all four sub-metrics, which accounts for its overall rank.

\paragraph{Cross-Dimension Analysis.}

Figure \ref{fig4} reports Pearson correlations among the metrics and the agreement between VehDyn score and human judgments.
Visual evaluation under the WorldScore and DrivingGen protocols is only weakly related to vehicle-dynamics fidelity.
Following the human-alignment protocol \citep{huang2024vbench}, 50 annotators scored trajectory alignment, kinematic consistency, and dynamic consistency on a [0, 1] scale.
The VehDyn score exhibits strong agreement with human evaluation, achieving Pearson $r=0.8619$ and Spearman $\rho=0.8042$. 
More details about human alignment are provided in Appendix \ref{appendix:human}.

\begin{figure*}[ht!]
  \centering
  \includegraphics[width=\linewidth]{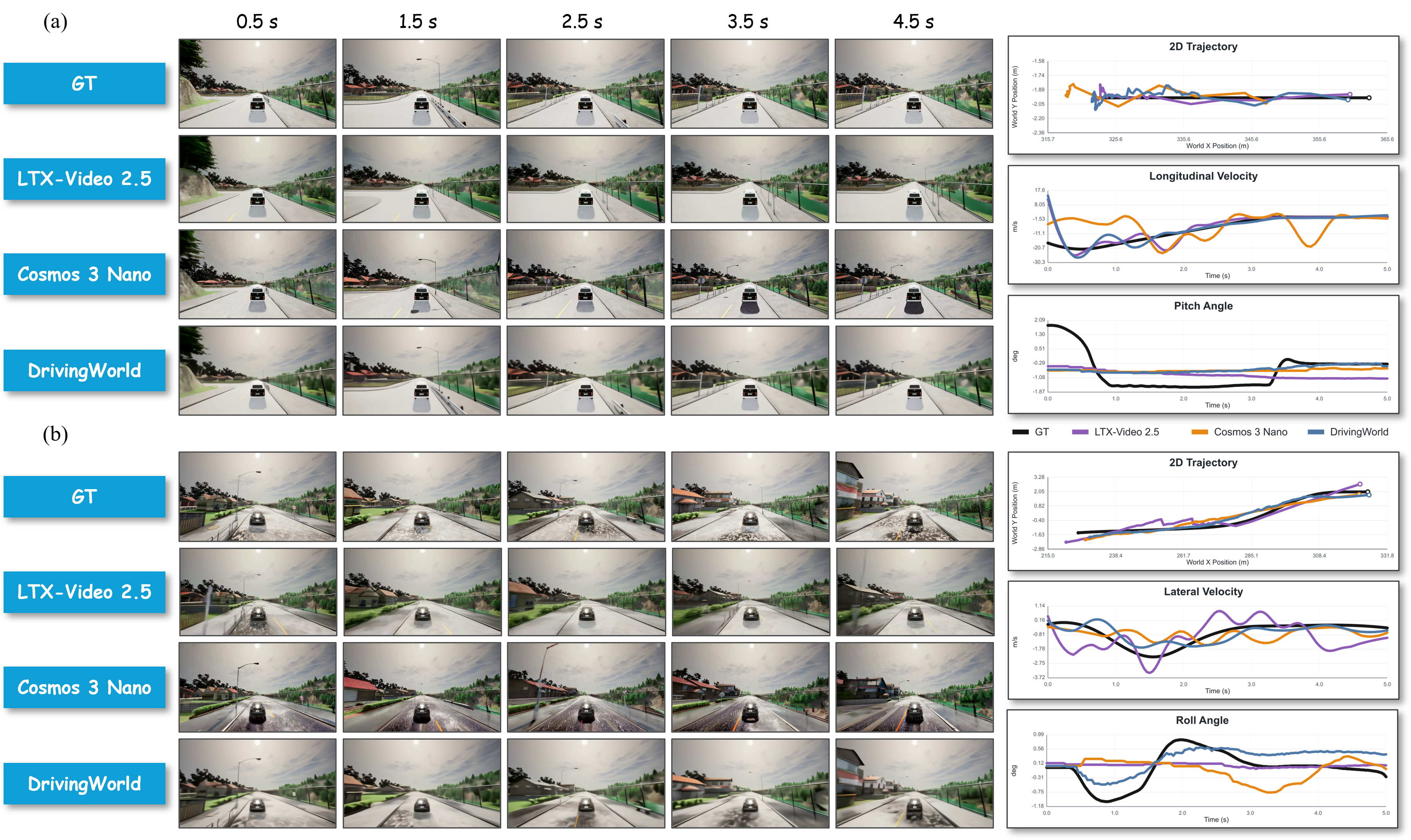}
  \caption{Qualitative comparison of the top-three models with ground truth. (a) Emergency-braking scenario. (b) Single-lane-changing scenario. Frames are shown at 0.5 s to 4.5 s. The right-hand panels plot the trajectory, velocity, and attitude.}
  \label{fig5}
\end{figure*}

\subsection{Qualitative Results}

Figure \ref{fig5} compares the three highest-ranked models with ground truth under identical scenario configurations. 
In the emergency-braking scenario, all three models successfully completed the maneuver but exhibited distinct dynamic behaviors. 
LTX-Video 2.5 produced temporally coherent visual transitions and a relatively smooth deceleration, but failed to capture the transient pitch response. 
Cosmos 3 Nano preserved richer scene details, but its deceleration showed noticeable deviations and fluctuations. 
DrivingWorld, despite lower visual fidelity, reproduced the ground-truth motion more accurately, with both its velocity and trajectory closely following the ground truth.
In the single-lane-changing scenario, all three models also complete the maneuver.
LTX-Video 2.5 and Cosmos 3 Nano retain better visual appearance while exhibiting larger motion discrepancies, whereas DrivingWorld sacrifices some visual details but achieves better consistency with ground-truth states. 

\section{Conclusion}

We introduce VehDyn, a driving world model benchmark that evaluates whether generated driving futures obey vehicle kinematics and dynamics.
VehDyn couples CARLA visual rendering with CarSim vehicle dynamics to produce 10,080 configurations from a full factorial design over six factors, each paired with synchronized ground-truth vehicle states.
A hierarchical protocol evaluates trajectory alignment, kinematic consistency, and dynamic consistency.
The resulting VehDyn score agrees closely with human judgment. 
Across 12 video world models, DrivingWorld achieves the highest VehDyn score, followed by Cosmos 3 Nano and LTX-Video 2.5. 
The results reveal a systematic gap between plausible and physically consistent generation: trajectory-level metrics are nearly saturated, kinematic errors of the best models approach the extraction floor, and no model reaches 92\% of ground truth on dynamic consistency.
Visual-quality metrics are only weakly related to these outcomes.
This work has limitations.
VehDyn is simulation-based, evaluates a single ego vehicle without traffic interaction, and relies on monocular state recovery whose error bounds the final score. 
Future work will extend the benchmark to multi-agent interaction, multi-view observation, and closed-loop planning, and will validate the protocol against real-world driving scenarios.

\bibliographystyle{unsrt}  
\bibliography{references}  

\newpage

\appendix
\section{Appendix}

\subsection{Supplementary Details about Literature Review}
\label{appendix:literature}

\subsubsection{Driving World Models}

Large-scale video diffusion and autoregressive architectures have shifted the paradigm from prediction in low-dimensional latent spaces to direct prediction in pixel spaces \citep{hacohen2025ltx,wan2025wan,yang2025cogvideox}. 
Driving world models have followed this shift with progressively richer conditioning.
Early representative systems such as GAIA-1 \citep{hu2023gaia} formulated driving video generation as autoregressive prediction over discrete tokens conditioned on text and ego actions, and the subsequent model GAIA-2 \citep{russell2025gaia} adopted a latent diffusion formulation with structured control over ego motion, surrounding agents, and multi-camera layout.
GenAD \citep{yang2024generalized} and Vista \citep{gao2024vista} scaled training to internet-sourced driving videos and introduced trajectory and command conditioning, while GEM \citep{hassan2025gem} extended control to object dynamics and scene composition.
More recent approaches, such as Epona \citep{zhang2025epona} and DrivingWorld \citep{hu2026drivingworld}, coupled video generation with planning-oriented driving tasks.
In parallel, general-purpose and physics-oriented foundation models, most notably the Cosmos family \citep{agarwal2025cosmos,agarwal2026cosmos} and its derivatives \citep{ren2025cosmos,basant2026nvidia}, are increasingly applied to driving tasks without domain-specific training. 
These advances have substantially improved visual fidelity, temporal coherence, and controllability. 

\subsubsection{Driving World Model Benchmarks}

\paragraph{Evaluation.} 
General video generation benchmarks, such as the VBench series \citep{huang2024vbench,zheng2025vbench,huang2025vbench++}, WorldScore \citep{duan2025worldscore}, and WorldModelBench \citep{li2026worldmodelbench} evaluated perceptual quality, temporal coherence, and prompt adherence, with physical plausibility scored by vision-language model (VLM)-based judges \citep{sun2026drivejudge,schafer2026egodyn}.
Evaluation of driving world models initially relied on FID and FVD \citep{wang2024drivedreamer,wang2024driving}, supplemented by downstream perception or planning performance on nuScenes \citep{caesar2020nuscenes}. 
ACT-Bench \citep{arai2024act} introduced action fidelity, estimating the ego trajectory from generated video with monocular visual odometry and comparing it with the commanded trajectory, and reported action match rate, ADE, and FDE. 
DrivingGen \citep{zhou2026drivinggen} broadened this scope to distribution realism, driving-specific image quality, scene- and agent-level temporal consistency, and trajectory alignment, and added a trajectory consistency score that measures the smoothness of the velocity and acceleration recovered by SLAM. 
WorldLens \citep{liang2026worldlens} further added reconstruction, action following, downstream-task utility, and human-preference assessment. 
What-If World \citep{cai2026if} addressed a different but complementary limitation by testing whether a model's output changes correctly when a single physical variable in the prompt is altered.
Despite this rapid progress, existing benchmarks primarily assess perceptual physical plausibility, trajectory-level behavior, and qualitative responses to interventions. 

\paragraph{Data.} 
The ability to evaluate vehicle dynamics of driving world models is also constrained by the available data. 
Existing driving world model benchmarks draw evaluation data from real-world recordings, principally nuScenes \citep{caesar2020nuscenes}, nuPlan \citep{caesar2021nuplan}, NAVSIM \citep{dauner2024navsim}, DrivingDojo \citep{wang2024drivingdojo}, and StyleDrive \citep{hao2026styledrive}, or curated from internet videos \citep{zhou2026drivinggen}. 
These datasets are well suited for evaluating visual realism and generalization across natural driving distributions, but their physical factors are observational rather than experimentally controlled. 
In particular, naturalistic datasets generally do not provide matched samples in which vehicle type, speed, maneuver, and tire-road friction are independently varied while all relevant vehicle states are synchronously recorded.
Simulation-based driving datasets such as the CARLA Leaderboard \citep{dosovitskiy2017carla}, Fail2Drive \citep{gerstenecker2026fail2drive}, and the Bench2Drive family \citep{jia2024bench2drive,shao2026can,zhang2026bench2drive} offered controllable scenarios, but they are designed for closed-loop evaluation of autonomous driving policies rather than measuring the physical fidelity of generated videos. 
High-fidelity vehicle dynamics simulators such as CarSim model tire forces, suspension kinematics, and load transfer with validated multi-body formulations and are the standard tool for vehicle dynamics analysis \citep{zeng2023wheel,zeng2026damper}.
However, such simulators have rarely been coupled with photorealistic environment simulation to construct systematic evaluation data for video world models.
Table \ref{tab:overall_comparison} summarizes the comparison with existing benchmarks and datasets.

\subsection{Supplementary Details about VehDyn Dataset}
\label{appendix:dataset}

\subsubsection{Dataset Statistics}

The key statistics of the VehDyn dataset are summarized in Table \ref{tab:dataset_item_type}.

\begin{table}[!htbp]
\centering
\caption{Configuration of the VehDyn dataset.}
\label{tab:dataset_item_type}
\small
\renewcommand{\arraystretch}{1.15}
\setlength{\tabcolsep}{4pt}
\begin{tabularx}{\textwidth}{@{}l >{\hsize=0.85\hsize\raggedright\arraybackslash}X c >{\hsize=1.15\hsize\raggedright\arraybackslash}X c@{}}
\toprule
\textbf{Factor} & \textbf{Definition} & \textbf{Unit} & \textbf{Levels} & \textbf{Number} \\
\midrule
\multicolumn{5}{@{}l}{\textit{Dynamic Factors}} \\
Vehicle type         & Ego vehicle model          & --   & Cooper S; Model 3, Grand Tourer; T2, Cybertruck & 5 \\
Friction coefficient & Tire-road friction coefficient       & --   & Asphalt: 0.85 (dry), 0.70 (wet); stone: 0.60 (dry), 0.50 (wet) & 4 \\
Driving maneuver     & Prescribed ego maneuver                      & --   & Single-lane-changing (left), single-lane-changing (right), emergency-braking & 3 \\
Target speed         & Cruise speed before maneuver            & km/h & 20, 54, 72, 108 & 4 \\
\midrule
\multicolumn{5}{@{}l}{\textit{Visual Factors}} \\
Scene                & Urban route in CARLA Towns 1-7              & --   & Two routes per town & 14 \\
Sun angle            & Solar elevation angle                        & deg  & 75, 45, 5 & 3 \\
\bottomrule
\end{tabularx}
\end{table}

The physical parameters of the five vehicle models are summarized in Table \ref{tab:vehicle_parameters}. 
Vehicle dimension data are derived from the bounding boxes of CARLA actors, while dynamic parameters are derived from representative CarSim configurations.

\begin{table}[ht!]
\centering
\caption{Physical parameters of the five vehicle models.}
\label{tab:vehicle_parameters}
\small
\setlength{\tabcolsep}{4.5pt}
\renewcommand{\arraystretch}{1.08}
\begin{tabular}{@{}l c c c c c c@{}}
\toprule
\textbf{Parameter}
& \textbf{Unit}
& \textbf{Cooper S}
& \textbf{Model 3}
& \textbf{Grand Tourer}
& \textbf{T2}
& \textbf{Cybertruck} \\
\midrule
\multicolumn{7}{@{}l}{\textit{Mass and Geometry}} \\[1pt]
Sprung mass
& kg
& 1220
& 1445
& 1370
& 1250
& 1800 \\
Length
& m
& 3.81
& 4.79
& 4.61
& 4.48
& 6.27 \\
Width
& m
& 1.97
& 2.16
& 2.24
& 2.07
& 2.39 \\
Height
& m
& 1.48
& 1.49
& 1.67
& 2.04
& 2.10 \\
Wheelbase
& m
& 2.13
& 2.68
& 2.58
& 2.51
& 3.20 \\
Front/rear track
& m
& 1.68
& 1.55
& 1.55
& 1.60
& 1.80 \\
Wheel radius
& m
& 0.38
& 0.37
& 0.36
& 0.35
& 0.54 \\
\addlinespace[2pt]
\midrule
\multicolumn{7}{@{}l}{\textit{Rotational Inertia}} \\[1pt]
Roll inertia
& kg$\cdot$m$^2$
& 536.6
& 671.3
& 761.3
& 928.1
& 1250.1 \\
Pitch inertia
& kg$\cdot$m$^2$
& 1536.7
& 1972.8
& 2100.0
& 2300.0
& 3650.0 \\
Yaw inertia
& kg$\cdot$m$^2$
& 1536.7
& 2821.0
& 2200.0
& 2437.0
& 4100.6 \\
\bottomrule
\end{tabular}%
\end{table}

\subsubsection{Route Details}

\begin{figure*}[!htbp]
    \centering
    \includegraphics[width=1\textwidth]
    {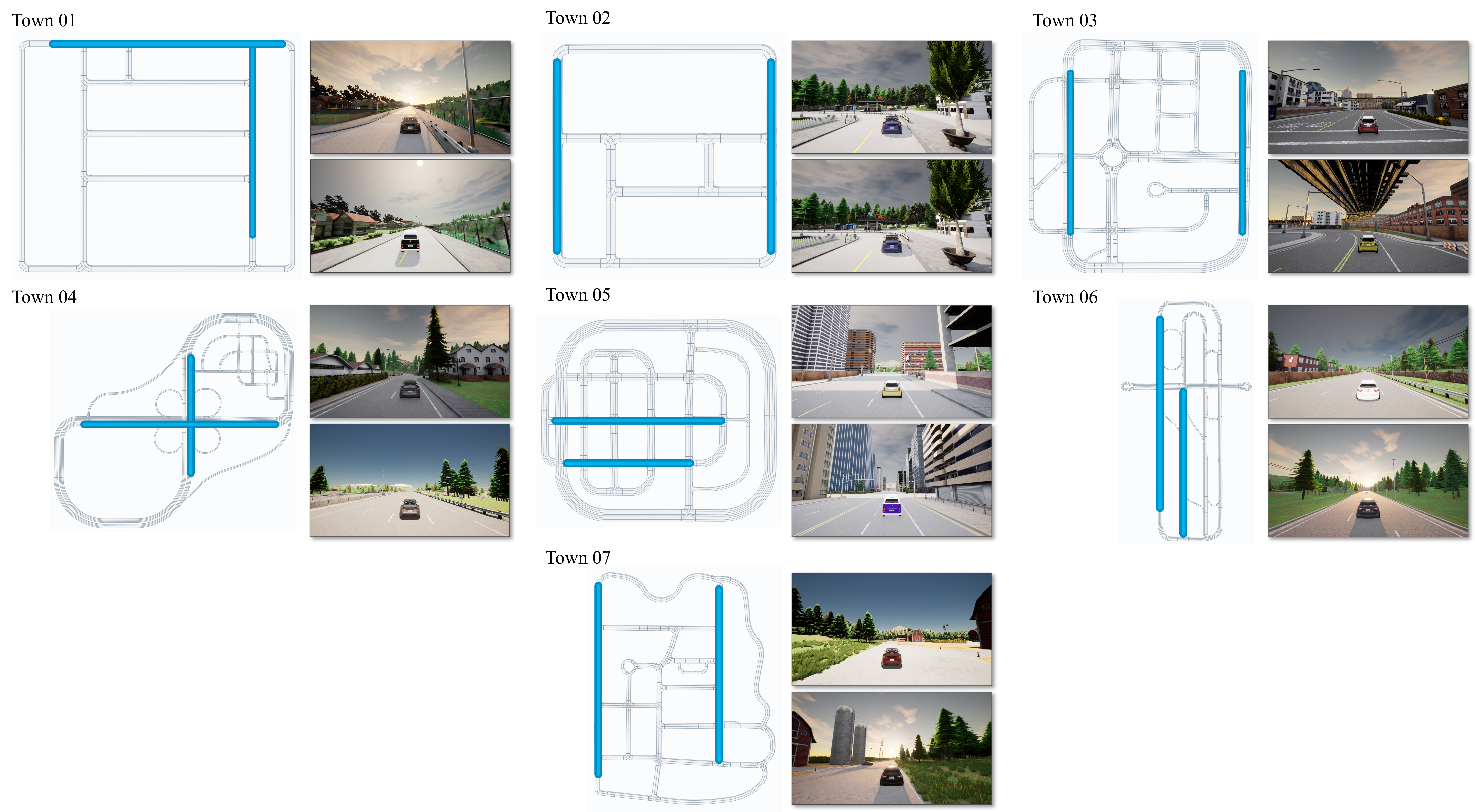}
    \caption{
    Route diversity. For each of CARLA Towns 1–7, the map shows two straight road segments used as scenes (blue) and a representative frame from each, illustrating the variation in surroundings, building density, and horizon visibility across scenes at fixed road geometry.
    }
    \label{fig:route_diversity}
\end{figure*}

VehDyn contains 14 evaluation routes sampled across CARLA Towns 1--7 (Figure \ref{fig:route_diversity}). 
Each scene is a straight road segment selected under three criteria. 
First, the segment is long enough to contain the entire maneuver at the highest target speed: at 108 km/h the vehicle covers 150 m during a 5 s clip, and the emergency-braking distance on the lowest-friction surface exceeds 90 m, so segments are at least 150 m long.
Second, the segment has at least two lanes in the direction of travel with a standard lane width of 3.5 m, so that left and right lane changes are both feasible and geometrically identical across scenes. 
Third, the segment contains no intersections, traffic signals, or road-surface transitions, so that the recorded response is attributable to the controlled factors alone. 
Towns 1 and 2 provide small-town streets with low buildings and roadside vegetation, Town 3 a dense urban layout with tall buildings, Town 4 a highway-style road with open surroundings, Town 5 a grid of multi-lane urban blocks, Town 6 wide multi-lane roads, and Town 7 a rural setting with fields and sparse structures.
The 14 scenes span the range of background clutter, occluding structures, and horizon visibility that a state-recovery pipeline and a generative model encounter, while holding the physical task fixed. 
The road-surface material and wetness that define the friction coefficient are applied uniformly along each segment and rendered consistently in CARLA. 

\subsubsection{Prompt for Video Caption} 

Below is the full prompt used to generate captions for video segments.

\begin{tcolorbox}[colback=gray!10, colframe=gray!20, left=2mm, right=2mm, top=1mm, bottom=1mm, boxrule=0pt]
\small
You are an expert in video understanding and captioning for driving videos. You are given \{n\} frames sampled at \{fps\} fps from a \{dur\}\,s third-person chase-camera clip of a single vehicle in the CARLA simulator, together with its ground-truth labels: maneuver \textbf{\textit{\{maneuver label\}}}, target speed \textbf{\textit{\{target speed\} km/h}}, vehicle \textbf{\textit{\{vehicle model\}}}, and road surface \textbf{\textit{\{surface material\}, \{wetness level\}}}.

\textbf{PURPOSE}: The caption is the text condition for an image-to-video model that receives only the initial frame plus your caption and must regenerate the clip. The initial frame already supplies scene, lighting, surface appearance, and vehicle appearance, so do not describe them. Spend the caption on what the frame cannot convey: the maneuver, its onset, direction and extent, the resulting change in vehicle attitude, and the final state.

\textbf{VIEWPOINT}: The camera is mounted above and behind the vehicle and rigidly attached to it, so it moves with the vehicle. The vehicle stays in a fixed region of the frame while the road moves past; attitude changes appear as a dip or rise of the body and a corresponding shift of the horizon. Always describe the view as a chase camera rigidly following the vehicle, never as a dashboard or static camera.

Before writing, reason through these steps internally (do not output the reasoning):

\textbf{STEP 1 - Parse the labels}. Derive the expected motion: \textit{emergency-braking} is a steady cruise followed by hard deceleration to a stop or near-stop; \textit{single-lane-changing left/right} is a steady cruise followed by a one-lane lateral shift in the stated direction and a return to straight travel. Use the label wording for vehicle, speed, and surface; do not invent any other number, unit, or physical parameter.

\textbf{STEP 2 - Identify the initial state}. From the first frame, record the lane position, heading, road geometry ahead, and brake-light state. Motion is described relative to this start.

\textbf{STEP 3 - Trace the maneuver over time}. Track ordered phases: cruise $\rightarrow$ onset (brake lights, start of lateral displacement) $\rightarrow$ transient $\rightarrow$ completion (stopped, or settled in the new lane). Record direction, extent (e.g., shifts one lane to the left; stops before the clip ends), and manner (abrupt, smooth). Describe the visible attitude response qualitatively (slight, moderate, pronounced): nose dip and rebound under braking; body roll and yaw during a lane change. Note surface cues such as spray only if visible.

\textbf{STEP 4 - Cross-check and compose}. Verify that the motion matches the maneuver label and the initial frame; describe only observable events, without driver intent, unseen agents, or physical laws. Then convey, in order: the chase-camera viewpoint moving with the vehicle, the vehicle model and surface condition, the cruise at the target speed, the maneuver from onset to completion, the attitude response, and the final state.

Caption requirements:
\begin{enumerate}
\item Chase-camera viewpoint rigidly attached to the vehicle; objective and neutral, no first-person pronouns.
\item Consistent with the maneuver label and direction; only visually observable events.
\item Temporally complete: cruise $\rightarrow$ onset $\rightarrow$ transient $\rightarrow$ completion, with direction and extent of the maneuver and attitude response.
\item State vehicle, target speed, and surface exactly as supplied; no other numbers or physical parameters; no re-description of static appearance.
\item Concise but motion-dense: target \{MAX\_WORDS, e.g. 50--80\} words.
\end{enumerate}
The output must follow exactly the format specified below:
\begin{verbatim}
{"maneuver_label": "text",
"caption": "text"}
\end{verbatim}
Return only the JSON object and no additional text.
\end{tcolorbox}

\subsubsection{Parameter Definition}

Every variable, its source, unit, and acquisition settings are provided in Table \ref{tab:variable_definition}.

\begin{table}[ht!]
\centering
\caption{Recorded variables, data sources, and acquisition settings of the VehDyn dataset.}
\label{tab:variable_definition}
\small
\renewcommand{\arraystretch}{1.15}
\setlength{\tabcolsep}{3pt}
\begin{tabularx}{\textwidth}{@{}
  l
  >{\hsize=0.70\hsize\raggedright\arraybackslash}X
  >{\hsize=1.22\hsize\raggedright\arraybackslash}X
  c
  c
  >{\hsize=1.08\hsize\raggedright\arraybackslash}X
@{}}
\toprule
\textbf{Category} & \textbf{Variable} & \textbf{Definition} & \textbf{Source} & \textbf{Unit} & \textbf{Setting} \\
\midrule
\multirow{2}{*}{Vehicle pose}
& $(x, y, z)$ & Position of the vehicle reference point in the CARLA world frame & CarSim & m & 30 Hz \\
& $(\phi, \theta, \psi)$ & Roll, pitch, yaw of the body & CarSim & deg & 30 Hz \\
\midrule
\multirow{3}{*}{Vehicle state}
& $(v_x, v_y)$ & Planar velocity recorded in the world frame and rotated into the body frame & CarSim & m/s & 30 Hz \\
& $r$ & Yaw rate about the body & CarSim & deg/s & 30 Hz \\
& $(u_\mathrm{thr}, u_\mathrm{brk}, u_\mathrm{str})$ & Throttle, brake, and steering commands issued by the controller & CARLA & -- & Throttle and brake in $[0,1]$, steering in $[-1,1]$ \\
\midrule
Scenario & Route progress & Distance traveled along the reference path & CARLA & m & 30 Hz \\
\midrule
\multirow{4}{*}{Camera}
& Mounting & Rigid RGB camera attached to the body & CARLA & -- & $(-8.5, 0, 2.5)$ m from the reference point, pitch ($8^\circ$ downward) \\
& Resolution & RGB image dimensions & CARLA & px & $1920 \times 1080$ \\
& Field of view & Horizontal field of view & CARLA & deg & $110^\circ$ \\
& Frame rate & Video sampling rate & CARLA & Hz & 30 Hz, 150 frames per clip \\
\midrule
\multirow{2}{*}{Simulation}
& Integration & CarSim dynamics integration between CARLA ticks. & CarSim & Hz & 2,000 Hz \\
& Synchronization & Frames and state rows share the CARLA tick index & CARLA/CarSim & -- & Synchronous fixed-step mode \\
\bottomrule
\end{tabularx}
\end{table}

\subsubsection{Full Dynamic Characteristics Analysis}

\begin{figure*}[t!]
    \centering
    \includegraphics[width=\textwidth]
    {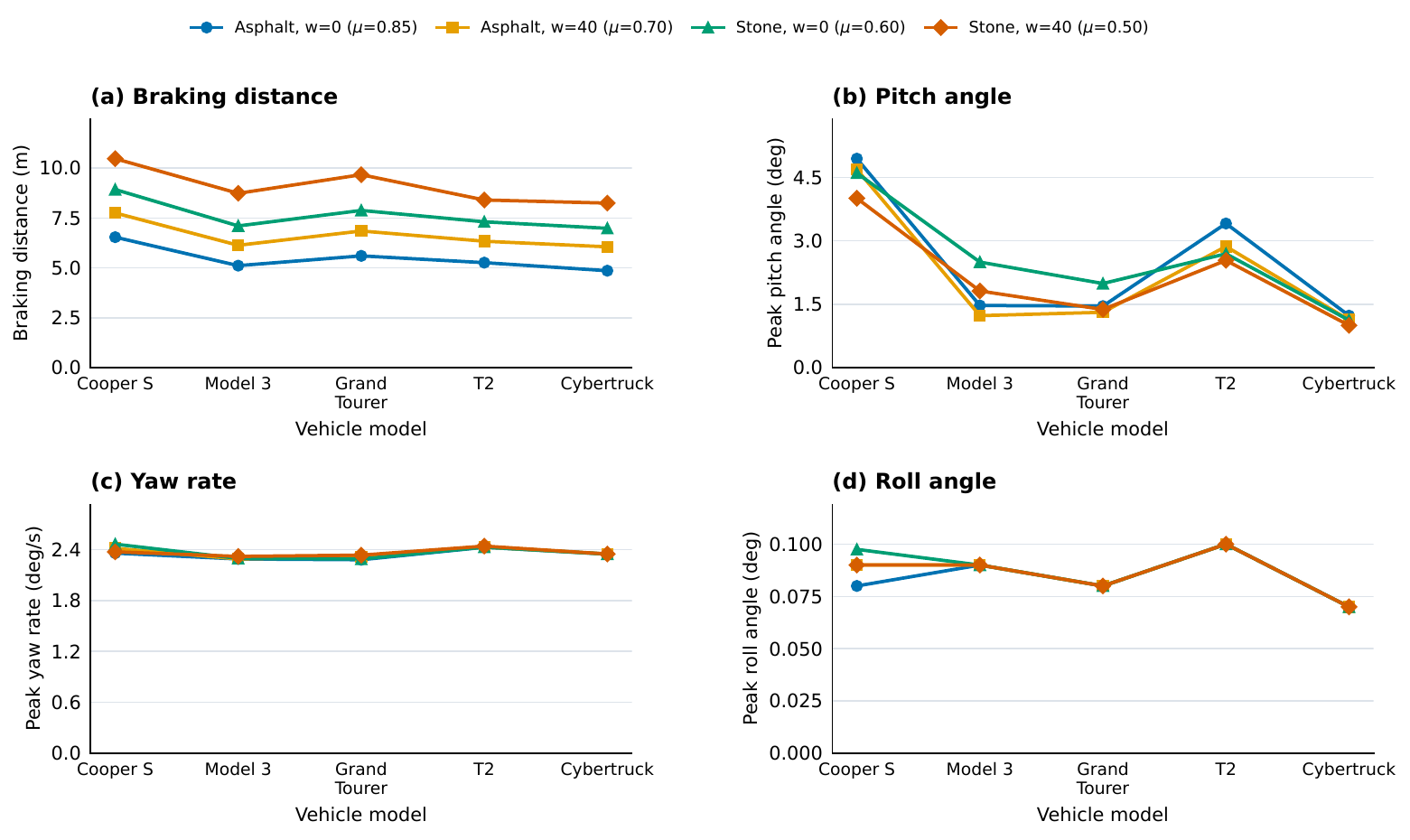}
    \caption{
        Ground-truth vehicle responses at 20~km/h.
        (a) Braking distance during emergency-braking; (b) Peak pitch angle change during emergency-braking;
        (c) Peak yaw rate during single-lane-changing; (d) Peak roll angle change during single-lane-changing.
    }
    \label{fig:gt_response_20kmh}
\end{figure*}

\begin{figure*}[t!]
    \centering
    \includegraphics[width=\textwidth]
    {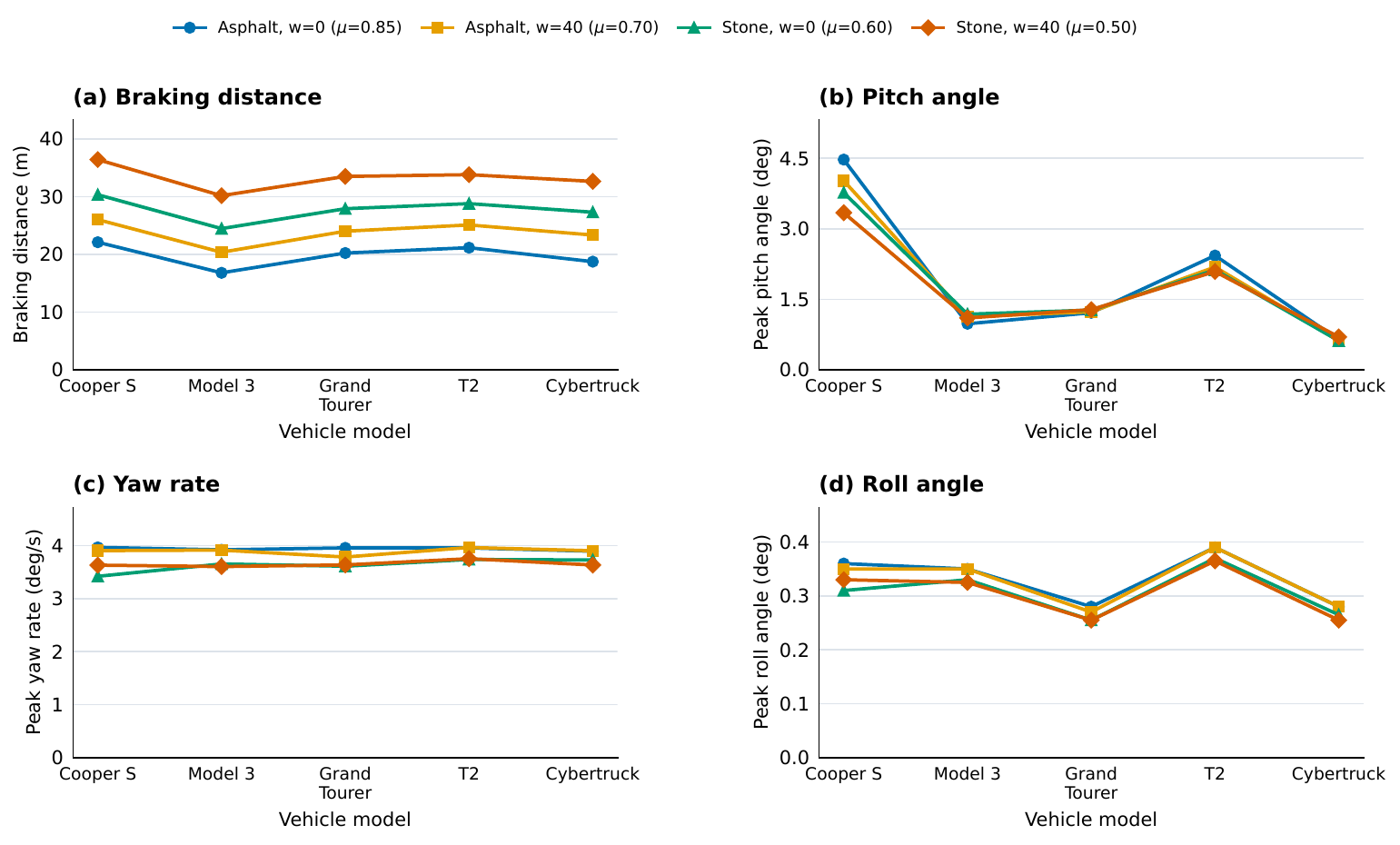}
    \caption{
        Ground-truth vehicle responses at 54~km/h.
        (a) Braking distance during emergency-braking; (b) Peak pitch angle change during emergency-braking;
        (c) Peak yaw rate during single-lane-changing; (d) Peak roll angle change during single-lane-changing.
    }
    \label{fig:gt_response_54kmh}
\end{figure*}

\begin{figure*}[t!]
    \centering
    \includegraphics[width=\textwidth]
    {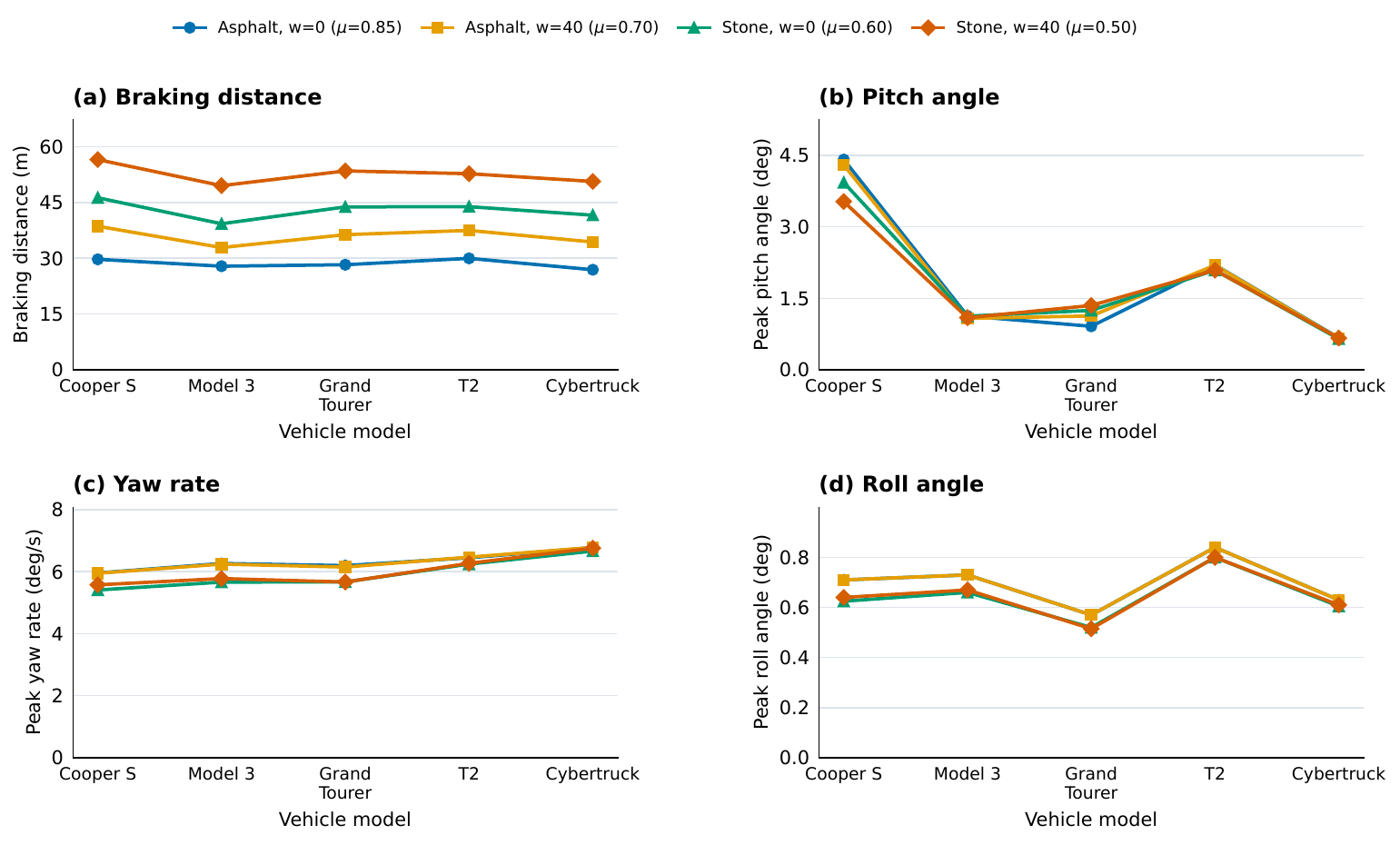}
    \caption{
        Ground-truth vehicle responses at 72~km/h.
        (a) Braking distance during emergency-braking; (b) Peak pitch angle change during emergency-braking;
        (c) Peak yaw rate during single-lane-changing; (d) Peak roll angle change during single-lane-changing.
    }
    \label{fig:gt_response_72kmh}
\end{figure*}

\begin{figure*}[t!]
    \centering
    \includegraphics[width=\textwidth]
    {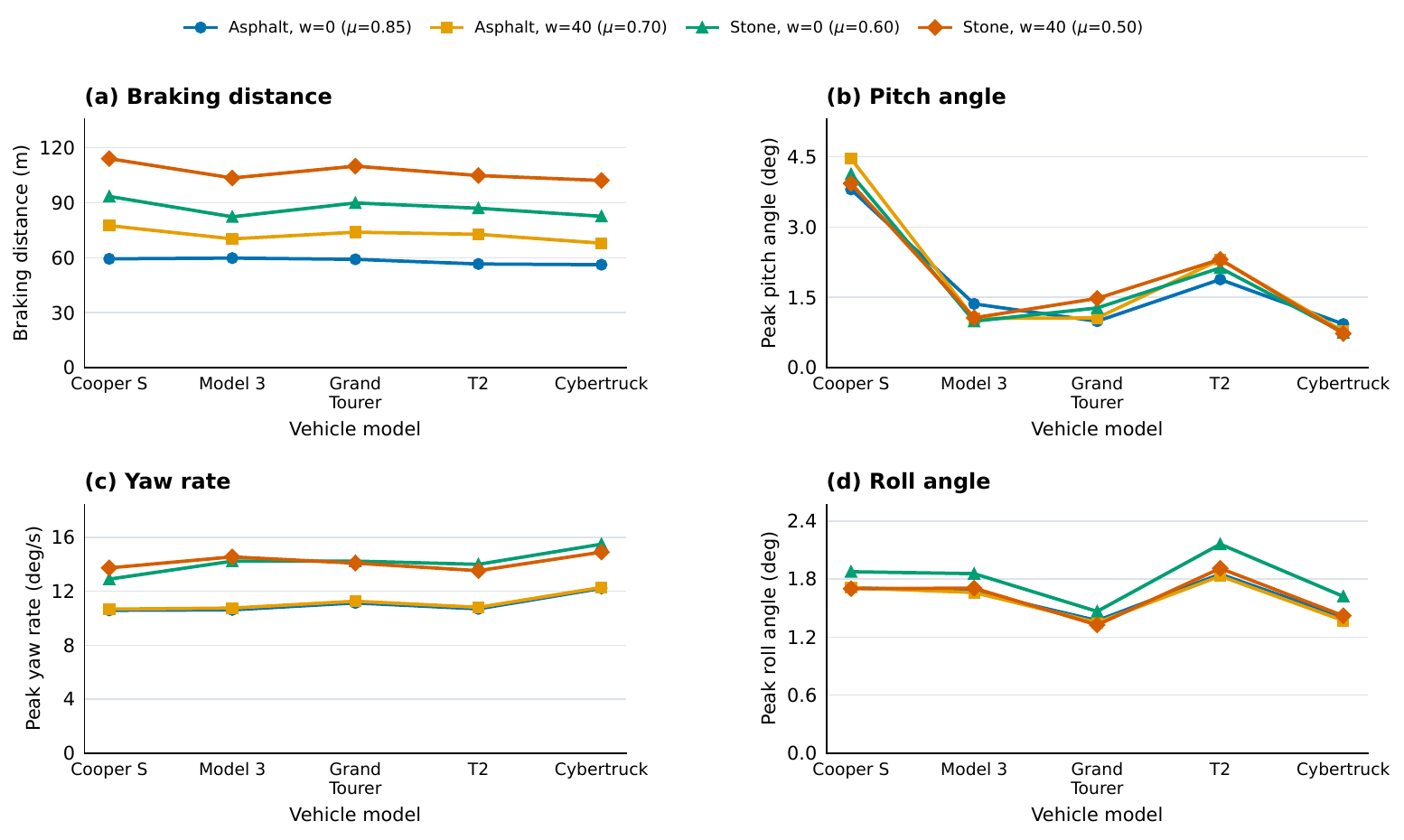}
    \caption{
        Ground-truth vehicle responses at 108~km/h.
        (a) Braking distance during emergency-braking; (b) Peak pitch angle change during emergency-braking;
        (c) Peak yaw rate during single-lane-changing; (d) (d) Peak roll angle change during single-lane-changing.
    }
    \label{fig:gt_response_108kmh}
\end{figure*}

Figures \ref{fig:gt_response_20kmh}--\ref{fig:gt_response_108kmh} present the ground-truth vehicle states across four target speeds. 
To ensure consistent comparisons, the emergency-braking results are averaged over different scenes and lighting conditions, while the single-lane-changing results are additionally averaged over left- and right-lane-changing conditions. 
The four plots characterize the physical variations in the dataset from the perspectives of longitudinal braking distance, pitch, yaw, and roll responses.

In the emergency-braking scenario, braking distance increases substantially with target speed. 
At each speed, lower tire-road friction is consistently associated with a longer braking distance, and the differences induced by road conditions become more pronounced at higher speeds. 
For a given road condition and speed, the braking distances of different vehicle types remain relatively close. 
In contrast, the peak pitch-angle variation exhibits clear and consistent differences across vehicle types. 
These observations indicate that the dataset captures both longitudinal responses primarily governed by vehicle speed and road conditions and vehicle-dependent attitude responses, rather than forcing all dynamic states to vary in the same manner with the same physical factors.

The single-lane-changing results show that both the peak yaw rate and the roll-angle variation generally increase with vehicle speed. 
At lower speeds, the responses under different road conditions remain relatively close, whereas the differences among certain road conditions and vehicle types become more evident at higher speeds. 
Unlike braking distance, the yaw and roll responses do not exhibit a simple monotonic ordering with respect to the friction coefficient, suggesting that these states cannot be explained by tire-road friction alone. 
These condition-dependent ground-truth responses provide quantitative references for evaluating whether the vehicle motion in generated videos is consistent with the specified target speed, vehicle type, maneuver, and road condition.

\subsection{Supplementary Details about VehDyn Evaluation}
\label{appendix:evaluation}

\subsubsection{Model Configurations}

\paragraph{Video world models.} 

Table \ref{tab5} summarizes the configurations of the 12 evaluated video world models.
The five general world models are image-to-video diffusion transformers conditioned on the first frame and the VehDyn caption. 
The three physical world models share the Cosmos 3 Nano backbone or its driving derivative: Cosmos 3 Nano is post-trained on VehDyn and conditioned on the first frame and the caption; Vanilla Cosmos 3 Nano is the released checkpoint evaluated zero-shot with the same conditioning, included to isolate the effect of post-training. 
The four driving world models are conditioned on ego motion in their native formats: Vista and GEM receive the first frame and the ground-truth ego trajectory, and DrivingWorld and Epona receive the first frame and per-step ego pose. 
All models except Vanilla Cosmos 3 Nano are post-trained on the VehDyn training split following their official implementations and recommended hyperparameters on eight A100 80 GB GPUs. 
Because native output lengths range from 31 to 151 frames, every generated video is resampled spanning the 5 s clip and resized to 1280$\times$720 before scoring, and the ground-truth video is processed identically. 

\begin{table}[ht!]
\centering
\caption{Characteristics of the evaluated video world models.}
\label{tab5}
\small
\setlength{\tabcolsep}{5pt}
\renewcommand{\arraystretch}{1.15}
\resizebox{\linewidth}{!}{%
\begin{tabular}{@{}lllrcrr@{}}
\toprule
\textbf{Method} & \textbf{Architecture} & \textbf{Conditioning} & \textbf{Parameters} & \textbf{Resolution} & \textbf{Frames} & \makecell[c]{\textbf{Inference}\\\textbf{(s/video)}} \\
\midrule
\multicolumn{7}{@{}l}{\textit{General World Models}} \\
LTX-Video 2.5          & DiT      & First frame + Text                          & 35.57 B & 1280$\times$704 & 121 & 173.9   \\
CogVideoX 1.5          & 3D DiT   & First frame + Text                          & 10.55 B & 1280$\times$720 & 81  & 1082.6  \\
SkyReels-V2-I2V        & 3D DiT   & First frame + Text                          & 20.26 B & 1280$\times$720 & 77  & 1875.6 \\
Wan 2.2-I2V            & 3D DiT   & First frame + Text                          & 11.39 B & 1280$\times$704 & 77  & 246.2  \\
HunyuanVideo 1.5-I2V   & MMDiT    & First frame + Text                          & 17.98 B & 1280$\times$720 & 77  & 2130.6 \\
\midrule
\multicolumn{7}{@{}l}{\textit{Physical World Models}} \\
Cosmos 3 Nano          & MoT DiT  & First frame + Text            & 16.45 B & 1280$\times$720 & 77  & 138.2   \\
Vanilla Cosmos 3 Nano  & MoT DiT  & First frame + Text             & 16.37 B & 1280$\times$720 & 81  & 185.0  \\
Cosmos-Dreams          & Causal DiT & First frame + Text & 10.48 B & 1280$\times$720 & 151 & 507.3 \\
\midrule
\multicolumn{7}{@{}l}{\textit{Driving World Models}} \\
DrivingWorld           & \makecell[l]{Autoregressive\\Transformer} & \makecell[l]{First frame + Ego pose} & 1.08 B & 1280$\times$720 & 151 & 343.6 \\
GEM                    & 3D U-Net & First frame + Ego trajectory           & 2.09 B  & 1280$\times$720 & 77  & 468.5  \\
Vista                  & 3D U-Net & First frame + Ego trajectory           & 2.52 B  & 1024$\times$576 & 51  & 182.7   \\
Epona                  & \makecell[l]{Autoregressive\\Transformer + flow DiT} & \makecell[l]{First frame + Ego pose} & 2.92 B & 1024$\times$512 & 31 & 36.2 \\
\bottomrule
\end{tabular}}
\end{table}

\begin{table}[ht!]
\centering
\caption{Architectural comparison of the trajectory-reconstruction pipelines.}
\label{tab6}
\small
\setlength{\tabcolsep}{5pt}
\renewcommand{\arraystretch}{1.15}
\resizebox{\linewidth}{!}{%
\begin{tabular}{@{}lllll@{}}
\toprule
\textbf{Method} & \makecell[l]{\textbf{Vision Encoder /}\\\textbf{Depth Prior}} & \textbf{Feature Dimensions} & \textbf{Geometric Module} & \textbf{Trajectory Estimation} \\
\midrule
DPVO
& \makecell[l]{Two 4-block \\CNN encoders}
& \makecell[l]{128 matching; \\384 context}
& \makecell[l]{Patch graph +\\differentiable BA}
& \makecell[l]{Recurrent sparse patch matching \\with joint pose/depth refinement} \\
\addlinespace[3pt]
DrivingGen
& SIFT + UniDepthV2
& \makecell[l]{128-D SIFT descriptor;\\dense metric depth}
& RANSAC + PnP
& \makecell[l]{Relative-pose chaining; \\constant-velocity extrapolation \\with orientation perturbation} \\
\addlinespace[3pt]
GEM
& \makecell[l]{DROID learned CNN +\\Depth Anything V2}
& \makecell[l]{128 matching; 256 context;\\dense depth}
& \makecell[l]{Recurrent dense update +\\dense BA}
& \makecell[l]{Dense correspondence optimization\\with depth-assisted scale estimation} \\
\bottomrule
\end{tabular}}
\end{table}

\paragraph{Trajectory-reconstruction pipelines.}

DPVO is run with its released weights and default settings on the resampled frames, using the reported camera intrinsics. 
Recovered camera poses are converted to body poses through the fixed camera-to-body transform, aligned to the vehicle-centric frame of the first frame, and scaled. 
Table \ref{tab6} compares DPVO with the trajectory-extraction pipelines used by DrivingGen and GEM. 
DPVO is the most compact, and the only one that jointly refines pose and depth without a separate depth network.
We fit a ground plane to the point cloud reconstructed by DPVO using RANSAC. 
The perpendicular distance from the first-frame camera center to this plane gives the estimated camera height \(\hat h\) in reconstruction units. 
The acquisition camera has a known height \(h=2.5\,\mathrm{m}\) above the ground. 
We then apply the scale factor \(s=h/\hat h\) uniformly to the trajectory translations and map points. 
This scale-recovery step uses only the point cloud jointly optimized by VO and the single calibrated height \(h\).

\subsubsection{Metric Calculation}

\paragraph{Problem formulation.}

Let $\mathbf{s}_{n,1:T_n}$ denote the ground-truth state sequence of sample $n$, and $\hat{\mathbf{s}}_{n,1:T_n}$ the state sequence recovered from its generated video by a common state extractor (DPVO). 
Both sequences are resampled at the same timestamps and expressed in the vehicle-centric coordinate system defined at the first frame of the clip.
The state at time $t$ is:
\begin{equation}
\mathbf{s}_t = (x_t, y_t, v_{x,t}, v_{y,t}, a_{x,t}, a_{y,t}, \phi_t, \theta_t, \psi_t),
\end{equation}
where $(x_t,y_t)$ is the planar position, $(v_{x,t},v_{y,t})$ is the longitudinal and lateral velocity, $(a_{x,t},a_{y,t})$ denotes the longitudinal and lateral acceleration, and $(\phi_t,\theta_t,\psi_t)$ denotes roll angle, pitch angle, and yaw angle.

\paragraph{Trajectory alignment.}

We measure trajectory alignment using ADE and FDE:
\begin{align}
\mathrm{ADE} &= \frac{1}{N} \sum_{n=1}^{N} \frac{1}{T_n} \sum_{t=1}^{T_n} \left\| \hat{\mathbf{p}}_{n,t}-\mathbf{p}_{n,t} \right\|_2, \\ 
\mathrm{FDE} &= \frac{1}{N} \sum_{n=1}^{N} \left\| \hat{\mathbf{p}}_{n,T_n}-\mathbf{p}_{n,T_n} \right\|_2,
\end{align}
where $\mathbf{p}_t=(x_t,y_t)$. 
Lower ADE and FDE indicate better trajectory alignment.

\paragraph{Kinematic consistency.}

For each scalar state $z \in \{v_x, v_y, \phi, \theta, \psi\}$, we compute the root mean square error (RMSE) over the valid evaluation frames:
\begin{equation}
\mathrm{RMSE}(z)= \sqrt{ \frac{1}{\sum_{n=1}^{N}T_n} \sum_{n=1}^{N}\sum_{t=1}^{T_n} d_z(\hat z_{n,t},z_{n,t})^2 },
\end{equation}
where $d_z(\hat z,z)=\hat z-z$ for velocities.
For an attitude angle $\alpha$, we use the wrapped angular error:
\begin{equation}
d_{\mathrm{ang}}(\hat\alpha,\alpha)= \operatorname{wrap}_{[-\pi,\pi)} (\hat\alpha-\alpha).
\end{equation}
We evaluate $v_x$ and pitch only for emergency-braking samples, and $v_y$, roll, and yaw only for single-lane-changing samples. 
All kinematic-consistency metrics are lower-is-better.

\paragraph{Dynamic consistency.}

\textit{Friction-circle constraint rate (FCCR)} measures whether the generated motion satisfies the planar acceleration constraint associated with the road friction condition.
For sample $n$, let $\mu_n$ be the friction coefficient associated with its road condition, and let $\mathcal{T}^{\mathrm{act}}_n$ denote the labeled active-maneuver frames. 
The recovered resultant planar acceleration is:
\begin{equation}
\hat a_{\mathrm{res},n,t} = \sqrt{\hat a_{x,n,t}^{\,2}+\hat a_{y,n,t}^{\,2}}.
\end{equation}
A frame is considered feasible if:
\begin{equation}
\hat a_{\mathrm{res},n,t} \leq \mu_n g+\delta_{\mathrm{fc}},
\end{equation}
where $g=9.81~\mathrm{m/s^2}$, and $\delta_{\mathrm{fc}}$ is a fixed tolerance that absorbs aerodynamic and grade effects neglected by the friction-circle approximation and the differentiation noise of the recovered accelerations, which is set to $0.5~\mathrm{m/s^2}$ on the validation scenes. 
FCCR is defined as the mean over samples of the fraction of active-maneuver frames that are feasible:
\begin{equation}
\mathrm{FCCR} = \frac{1}{N} \sum_{n=1}^{N} \frac{1}{|\mathcal{T}^{\mathrm{act}}_n|} \sum_{t\in\mathcal{T}^{\mathrm{act}}_n}
\mathbb{I} \left[ \hat a_{\mathrm{res},n,t} \leq \mu_n g+\delta_{\mathrm{fc}} \right].
\end{equation}
A higher FCCR indicates that a larger fraction of generated maneuver states satisfies the road-condition-dependent dynamic constraint.

\textit{Condition-aware dynamic consistency (CDC)} is also crucial because that responds correctly to changed physical conditions should reproduce not only each recorded outcome but also the change in outcome between conditions.
We further evaluate the direction, magnitude, and trend of changes in the generated motion in response to variations in physical conditions. 
For each sequence, we extract the dynamic response vector:
\begin{equation}
\mathcal{R} = \{ D_{\mathrm{brake}}, T_{\mathrm{brake}}, P_{\mathrm{pitch}}, \Omega_{\mathrm{pitch}}, A_{\mathrm{lat}}, \Psi_{\mathrm{yaw}}, \Omega_{\mathrm{yaw}}, R_{\mathrm{roll}}, \Omega_{\mathrm{roll}} \},
\end{equation}
including braking distance $D_{\mathrm{brake}}$, braking duration $T_{\mathrm{brake}}$, peak pitch excursion $P_{\mathrm{pitch}}$, peak pitch rate $\Omega_{\mathrm{pitch}}$, peak lateral acceleration $A_{\mathrm{lat}}$, peak yaw excursion $\Psi_{\mathrm{yaw}}$, peak yaw rate $\Omega_{\mathrm{yaw}}$, peak roll excursion $R_{\mathrm{roll}}$, and peak roll rate $\Omega_{\mathrm{roll}}$, each measured over the active-maneuver window with the same procedure for recovered and ground-truth sequences (Table \ref{tab:cdc_response_scales}).

A matched pair $p=(i,j)$ consists of two samples that differ in exactly one dynamic factor (vehicle type, friction coefficient, or target speed), and agree in the other five factors.
The ground-truth and recovered changes of response $r\in\mathcal{R}$ are:
\begin{equation}
\Delta_{p,r}=r_j-r_i, \qquad \hat{\Delta}_{p,r}=\hat r_j-\hat r_i.
\end{equation}

To make response changes comparable across quantities with different units,
we calibrate a response-specific scale from a fixed ground-truth reference
set $\mathcal{D}_{\mathrm{ref}}$:
\begin{equation}
s_r =
\max\left(
\operatorname{Std}\left(\{r_n^{\mathrm{GT}}\}_{n\in\mathcal{D}_{\mathrm{ref}}}\right),
10^{-6}
\right),
\qquad
\tau_r = \eta s_r,
\end{equation}
where $\eta=0.05$. The response-specific values of $s_r$ and $\tau_r$ are
reported in Table~\ref{tab:cdc_response_scales}.

We retain pairs with a meaningful ground-truth response difference:
\begin{equation}
\mathcal{P}_r^{\star}
=
\left\{
p:\left|\Delta_{p,r}\right|>\tau_r
\right\}.
\end{equation}

\begin{table*}[t]
\centering
\caption{
Ground-truth response scales $s_r$ and meaningful-change thresholds
$\tau_r$ used for condition-aware dynamic consistency.
}
\label{tab:cdc_response_scales}
\small
\begin{tabular}{llccc}
\toprule
Maneuver & Response $r$ & $s_r$ & $\tau_r$ & Unit \\
\midrule
\multirow{4}{*}{Emergency-braking}
& Braking distance $D_{\mathrm{brake}}$ & 29.173 & 1.459 & m \\
& Braking duration $T_{\mathrm{brake}}$ & 1.469 & 0.073 & s \\
& Peak pitch excursion $P_{\mathrm{pitch}}$ & 1.248 & 0.062 & deg \\
& Peak pitch rate $\Omega_{\mathrm{pitch}}$ & 3.384 & 0.169 & deg/s \\
\midrule
\multirow{5}{*}{Single-lane-changing}
& Peak lateral acceleration $A_{\mathrm{lat}}$ & 1.872 & 0.094 & $\mathrm{m/s^2}$ \\
& Peak yaw excursion $\Psi_{\mathrm{yaw}}$ & 1.639 & 0.082 & deg \\
& Peak yaw rate $\Omega_{\mathrm{yaw}}$ & 5.518 & 0.276 & deg/s \\
& Peak roll excursion $R_{\mathrm{roll}}$ & 0.707 & 0.035 & deg \\
& Peak roll rate $\Omega_{\mathrm{roll}}$ & 2.504 & 0.125 & deg/s \\
\bottomrule
\end{tabular}
\end{table*}

The direction-consistency score is:
\begin{equation}
\mathrm{CDC}_{\mathrm{Dir}} = \frac{1}{\sum_{r\in\mathcal{R}}|\mathcal{P}_r^{*}|} \sum_{r\in\mathcal{R}} \sum_{p\in\mathcal{P}_r^{*}} \mathbb{I} \left[ \operatorname{sign}(\hat{\Delta}_{p,r}) = \operatorname{sign}(\Delta_{p,r}) \right].
\end{equation}

The magnitude-consistency score is:
\begin{equation}
\mathrm{CDC}_{\mathrm{Mag}} = \frac{1}{\sum_{r\in\mathcal{R}}|\mathcal{P}_r^{*}|} \sum_{r\in\mathcal{R}} \sum_{p\in\mathcal{P}_r^{*}} \exp\left( -\frac{ |\hat{\Delta}_{p,r}-\Delta_{p,r}| }{s_r} \right),
\end{equation}
where $s_r$ is a response-specific scale estimated from the development split, such as the standard deviation of the corresponding ground-truth response.

Finally, for an ordered condition sweep $g$, such as the four target-speed levels under fixed vehicle, road, scene, and illumination settings, let $\mathbf{r}_{g,r}$ and $\hat{\mathbf{r}}_{g,r}$ denote the ground-truth and generated response vectors. 
The trend consistency score is:
\begin{equation}
\mathrm{CDC}_{\mathrm{Trend}} = \frac{1}{\sum_{r\in\mathcal{R}}|\mathcal{G}_r|} \sum_{r\in\mathcal{R}} \sum_{g\in\mathcal{G}_r} \frac{ 1+\rho_{\mathrm{S}} (\hat{\mathbf{r}}_{g,r},\mathbf{r}_{g,r}) }{2},
\end{equation}
where $\rho_{\mathrm{S}}$ is Spearman's rank correlation coefficient. 
Higher values of all CDC metrics indicate better agreement with the dynamic changes induced by the conditions.

\paragraph{Score normalization.}

Each raw metric value $x_m$ is mapped to a normalized score $s_m \in [0, 1]$ with higher-is-better orientation.
The four dynamic-consistency metrics are natively in $[0, 1]$ and are used unchanged.
The other seven metrics are normalized by the clipped affine map using the bounds in Table \ref{tab7}.
For metrics with explicit lower and upper bounds $a_m$ and $b_m$, respectively, we use:
\begin{equation}
s_m =
\begin{cases}
\mathrm{clip}\!\left(
\dfrac{x_m-a_m}{b_m-a_m},0,1
\right),
& \text{higher is better}, \\
\mathrm{clip}\!\left(
\dfrac{b_m-x_m}{b_m-a_m},0,1
\right),
& \text{lower is better}.
\end{cases}
\label{eq:norm}
\end{equation}

For metrics whose original benchmarks do not provide usable normalization bounds, we follow the percentile strategy of WorldArena \citep{shang2026worldarena}: the 1st and 99th percentiles of the raw values over the VehDyn evaluation pool serve as $a_m$ and $b_m$, respectively. 
This strategy is applied to ADE, FDE, $v_x$, $v_y$, pitch, yaw, and roll.
After these transformations, every metric is expressed on a common $[0,1]$ higher-is-better scale.
The corresponding numerical bounds are reported in Table \ref{tab7}.

\begin{table}[ht!]
\caption{Bounds used for score normalization.}
\label{tab7}
\centering
\small
\begin{tabular}{l l c c}
\toprule
Metric & Strategy & $a_m$ & $b_m$ \\
\midrule
ADE & VehDyn empirical & 0.137 & 23.912 \\
FDE & VehDyn empirical & 0.061 & 49.116 \\
$v_x$ & VehDyn empirical & 0.249 & 20.324 \\
Pitch & VehDyn empirical & 0.525 & 9.854 \\
$v_y$ & VehDyn empirical & 0.030 & 22.919 \\
Roll & VehDyn empirical & 0.042 & 5.386 \\
Yaw & VehDyn empirical & 0.153 & 9.893 \\
FCCR & Native [0, 1] & 0.000 & 1.000 \\
CDC$_{\mathrm{Dir}}$ & Native [0, 1] & 0.000 & 1.000 \\
CDC$_{\mathrm{Mag}}$ & Native [0, 1] & 0.000 & 1.000 \\
CDC$_{\mathrm{Trend}}$ & Native [0, 1] & 0.000 & 1.000 \\
\bottomrule
\end{tabular}
\end{table}

\subsection{Supplementary Details about Quantitative Results}
\label{appendix:quantitative}

\subsubsection{Comparison of Different Trajectory-Reconstruction Pipelines}

\begin{figure*}[t!]
  \centering
  \includegraphics[width=\linewidth]{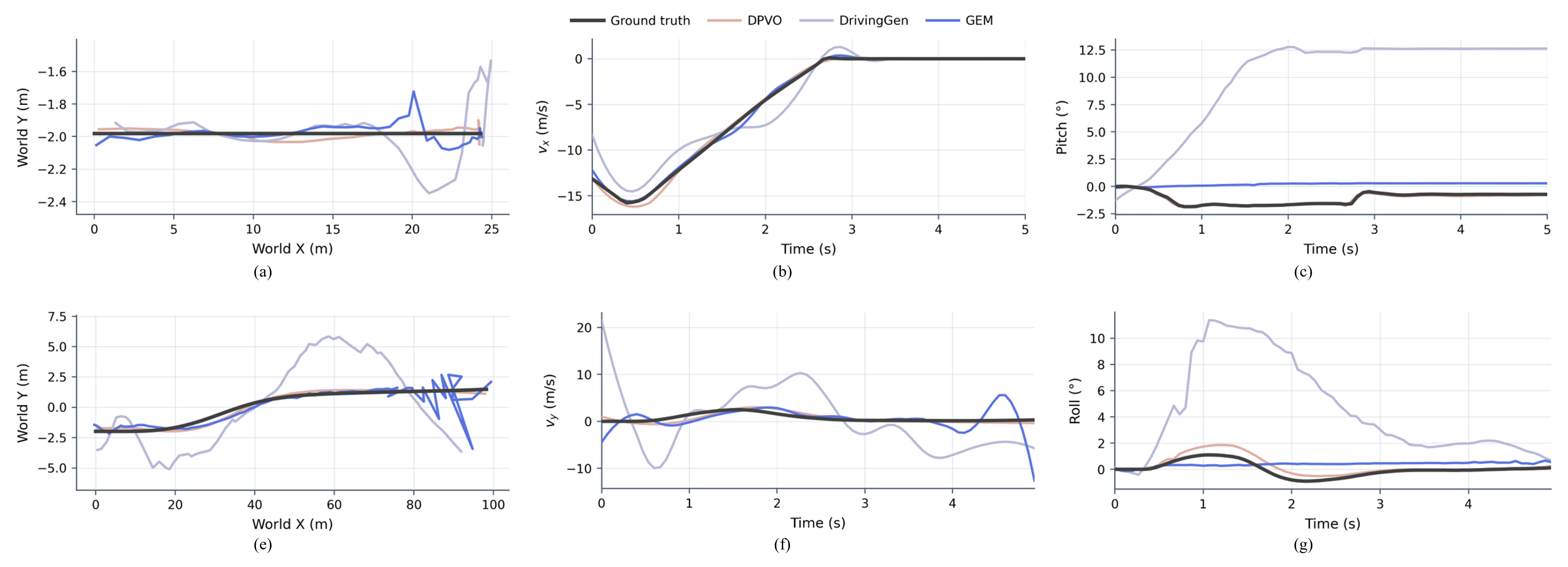}
    \caption{Qualitative comparison of vehicle motion under two representative driving scenarios.
    Top row: Emergency-braking, (a) vehicle trajectory, (b) longitudinal velocity $v_x$, and (c) pitch angle.
    Bottom row: Single-lane-changing, (d) vehicle trajectory, (e) lateral velocity $v_y$, and (f) roll angle.}
  \label{fig12_dpvo_comparsion}
\end{figure*}

Table \ref{tab:pipeline_extraction_comparison} compares DPVO, DrivingGen, and GEM under a unified evaluation protocol. 
DPVO achieves the lowest macro-averaged error across all reported metrics. 
Its ADE/FDE are 0.514/0.814 m, corresponding to reductions of 76.7\%/82.1\% over DrivingGen and 12.7\%/24.0\% over GEM. 
This indicates more accurate and stable trajectory reconstruction over complete driving sequences.
The same advantage is observed in maneuver-specific state recovery. 
In emergency braking, DPVO achieves the lowest longitudinal-velocity and pitch RMSEs of 0.897 m/s and $2.522^\circ$. 
In the single-lane-changing scenario, it also provides the lowest lateral-velocity and roll RMSEs of 0.697 m/s and $0.602^\circ$, while its yaw RMSE ($3.885^\circ$) is marginally lower than that of GEM ($3.888^\circ$). 
Although GEM shows slightly lower median errors for a few metrics, DPVO provides the best overall accuracy across the full evaluation set.
Figure \ref{fig12_dpvo_comparsion} further illustrates these differences. 
DPVO accurately reproduces the ground-truth deceleration, pitch response, lateral-velocity pulse, and roll-yaw evolution. 
In contrast, DrivingGen exhibits pronounced motion and attitude drift, while GEM tends to suppress part of the transient or coupled dynamic response.
The quantitative results are computed over 10,080 valid sequences, whereas the figures show representative cases for intuitive interpretation. 
Overall, these results support the use of DPVO as the final trajectory-reconstruction pipeline, as it achieves stronger geometric accuracy while better preserving the motion states required for downstream vehicle-dynamics analysis.

\begin{table*}[ht!]
\caption{Comparison of trajectory-reconstruction pipelines on the VehDyn dataset. Entries are sequence-level macro means, with medians in parentheses. All metrics are errors, so lower is better. The best mean value in each column is highlighted.}
\label{tab:pipeline_extraction_comparison}
\centering
\resizebox{\textwidth}{!}{%
\small
\setlength{\tabcolsep}{4.2pt}
\begin{tabular}{@{}l|*{2}{c}|*{2}{c}|*{3}{c}@{}}
\toprule
\multirow{2}{*}{\makecell[l]{Methods}}
& \multicolumn{2}{c|}{All Maneuvers}
& \multicolumn{2}{c|}{Emergency-Braking}
& \multicolumn{3}{c}{Single-Lane-Changing} \\
\cmidrule(lr){2-3} \cmidrule(lr){4-5} \cmidrule(lr){6-8}
& ADE $\downarrow$ & FDE $\downarrow$
& $v_x$ RMSE $\downarrow$ & Pitch RMSE $\downarrow$
& $v_y$ RMSE $\downarrow$ & Roll RMSE $\downarrow$ & Yaw RMSE $\downarrow$ \\
\midrule
DPVO
& \best{0.514 (0.301)} & \best{0.814 (0.346)}
& \best{0.897 (0.469)} & \best{2.522 (2.162)}
& \best{0.697 (0.521)} & \best{0.602 (0.468)} & \best{3.885 (3.813)} \\
DrivingGen
& 2.202 (1.591) & 4.536 (2.238)
& 1.572 (1.248) & 8.660 (6.726)
& 3.552 (2.736) & 9.141 (5.250) & 9.095 (6.823) \\
GEM
& 0.589 (0.283) & 1.071 (0.346)
& 0.982 (0.288) & 2.765 (2.037)
& 1.045 (0.661) & 0.624 (0.496) & 3.888 (3.863) \\
\bottomrule
\end{tabular}%
}
\end{table*}

\subsubsection{VehDyn Score}

\paragraph{Different maneuvers.}

The maneuver-specific results in Tables \ref{tab:ebrake_results}--\ref{tab:rlc_right_results} reveal that vehicle-dynamics fidelity is strongly dependent on the type of motion. 
DrivingWorld ranks first in emergency-braking and right-lane-changing and second in left-lane-changing, and is the only model in the top two of all three tables.
Cosmos 3 Nano ranks third in emergency-braking and right-lane-changing but seventh in left-lane-changing, where its trajectory and lateral-velocity errors are markedly larger than on the right.
In the emergency-braking scenario, Epona ranks second, supported by competitive trajectory errors and strong condition-aware dynamic responses, including the highest $CDC_{\mathrm{Dir}}$ and $CDC_{\mathrm{Trend}}$ among the models. 
However, its performance drops to eleventh in both single-lane-changing scenarios, where the trajectory and lateral-state errors increase substantially and $FCCR$ decreases to 0.282 and 0.315. 
This contrast shows that preserving the correct ordering of responses across physical conditions does not necessarily imply that the generated motion remains within the feasible lateral dynamic envelope.
In addition, LTX-Video 2.5 rises from sixth in emergency-braking to first in left-lane-changing and second in right-lane-changing, with the lowest ADE, FDE, lateral-velocity, and roll errors on the left. 
Wan 2.2-I2V and SkyReels-V2-I2V similarly perform considerably better in the lane-changing scenario than in emergency-braking. 
Separating the maneuver families exposes model-specific failure modes, feasible but condition-insensitive motion on one side and condition-sensitive but infeasible motion on the other, that a single aggregate score conceals.

\begin{table*}[ht!]
\caption{Emergency-braking results of the VehDyn benchmark. The rank is obtained from the maneuver-specific VehDyn score: normalized trajectory-alignment, kinematic-consistency, and dynamic-consistency scores are summed to a value in $[0,3]$. Best results are in \redregion{red region}, second best in \orangeregion{orange region}, and third best in \blueregion{blue region}. Models fall into three categories: general world models, physical world models, and driving world models.}
\label{tab:ebrake_results}
\centering
\resizebox{\textwidth}{!}{%
\small
\setlength{\tabcolsep}{4pt}
\begin{tabular}{@{}l|*{2}{c}|*{2}{c}|*{4}{c}|c@{}}
\toprule
\multirow{3}{*}{\makecell[l]{Models}} & \multicolumn{2}{c|}{Trajectory Alignment} & \multicolumn{2}{c|}{Kinematic Consistency} & \multicolumn{4}{c|}{Dynamic Consistency} & \multirow{3}{*}{\makecell[c]{Rank}}\\
\cmidrule(lr){2-3} \cmidrule(lr){4-5} \cmidrule(lr){6-6} \cmidrule(lr){7-9}
& ADE $\downarrow$ & FDE $\downarrow$ & $v_x$ $\downarrow$ & Pitch $\downarrow$ & FCCR $\uparrow$ & CDC$_{\mathrm{Dir}}$ $\uparrow$ & CDC$_{\mathrm{Mag}}$ $\uparrow$ & CDC$_{\mathrm{Trend}}$ $\uparrow$ & \\
\midrule
\textcolor{gray}{Ground Truth} & \textcolor{gray}{0.471} & \textcolor{gray}{0.584} & \textcolor{gray}{0.897} & \textcolor{gray}{2.522} & \textcolor{gray}{0.890} & \textcolor{gray}{0.740} & \textcolor{gray}{0.780} & \textcolor{gray}{0.716} & \textcolor{gray}{--}\\
\midrule
\multicolumn{10}{@{}l}{\textit{General World Models}} \\
LTX-Video 2.5 & 1.884 & 2.362 & 3.217 & \best{2.077} & 0.623 & 0.649 & 0.675 & 0.711 & 6 \\
CogVideoX 1.5 & 3.642 & 7.462 & 5.683 & 3.018 & \best{0.846} & 0.699 & 0.633 & 0.734 & 9 \\
SkyReels-V2-I2V & 4.415 & 9.118 & 7.243 & 2.415 & 0.658 & 0.692 & 0.654 & 0.736 & 10 \\
Wan 2.2-I2V & 6.410 & 8.761 & 9.063 & 2.161 & 0.556 & 0.684 & 0.710 & \second{0.788} & 11 \\
HunyuanVideo 1.5-I2V & 5.448 & 7.330 & 7.471 & 3.283 & 0.526 & 0.624 & 0.582 & 0.659 & 12 \\
\midrule
\multicolumn{10}{@{}l}{\textit{Physical World Models}} \\
Cosmos 3 Nano & \second{0.881} & \third{0.980} & \second{1.866} & 2.173 & 0.685 & \third{0.735} & \third{0.754} & 0.766 & 3 \\
Vanilla Cosmos 3 Nano & 2.244 & 3.220 & 4.415 & 2.203 & 0.682 & 0.721 & 0.714 & 0.740 & 7 \\
Cosmos-Dreams & 3.244 & 7.161 & 5.959 & 2.158 & \third{0.744} & 0.659 & 0.641 & 0.735 & 8 \\
\midrule
\multicolumn{10}{@{}l}{\textit{Driving World Models}} \\
DrivingWorld & \best{0.729} & \second{0.921} & \best{1.418} & \third{2.141} & \second{0.751} & \second{0.782} & \best{0.789} & \third{0.771} & 1 \\
Epona & \third{0.892} & \best{0.892} & \third{2.474} & \second{2.127} & 0.657 & \best{0.794} & \second{0.760} & \best{0.842} & 2 \\
GEM & 1.422 & 2.039 & 3.166 & 2.398 & 0.615 & 0.705 & 0.704 & 0.746 & 4 \\
Vista & 2.137 & 3.156 & 3.653 & 2.282 & 0.713 & 0.720 & 0.709 & 0.692 & 5 \\
\bottomrule
\end{tabular}%
}
\end{table*}

\begin{table*}[t!]
\caption{Left-lane-changing results of the VehDyn benchmark. All metrics are evaluated only on left single-lane-changing sequences. The final rank is obtained from the maneuver-specific VehDyn score: normalized trajectory-alignment, kinematic-consistency, and dynamic-consistency scores are summed to a value in $[0,3]$. Best results are in \redregion{red region}, second best in \orangeregion{orange region}, and third best in \blueregion{blue region}.}
\label{tab:llc_left_results}
\centering
\resizebox{\textwidth}{!}{%
\small
\setlength{\tabcolsep}{4pt}
\begin{tabular}{@{}l|*{2}{c}|*{3}{c}|*{4}{c}|c@{}}
\toprule
\multirow{2}{*}{\makecell[l]{VehDyn Metrics\\Models}}
& \multicolumn{2}{c|}{Trajectory Alignment}
& \multicolumn{3}{c|}{Kinematic Consistency}
& \multicolumn{4}{c|}{Dynamic Consistency}
& \multirow{2}{*}{\makecell[c]{Rank}}\\
\cmidrule(lr){2-3}
\cmidrule(lr){4-6}
\cmidrule(lr){7-10}
& ADE $\downarrow$
& FDE $\downarrow$
& $v_y$ $\downarrow$
& Roll $\downarrow$
& Yaw $\downarrow$
& FCCR $\uparrow$
& CDC$_{\mathrm{Dir}}$ $\uparrow$
& CDC$_{\mathrm{Mag}}$ $\uparrow$
& CDC$_{\mathrm{Trend}}$ $\uparrow$
& \\
\midrule
\textcolor{gray}{Ground Truth}
& \textcolor{gray}{0.586}
& \textcolor{gray}{1.103}
& \textcolor{gray}{0.812}
& \textcolor{gray}{0.652}
& \textcolor{gray}{4.275}
& \textcolor{gray}{0.884}
& \textcolor{gray}{0.711}
& \textcolor{gray}{0.650}
& \textcolor{gray}{0.853}
& \textcolor{gray}{--}\\
\midrule
\multicolumn{11}{@{}l}{\textit{General World Models}} \\
LTX-Video 2.5
& \best{1.365}
& \best{1.837}
& \best{0.939}
& \best{0.547}
& 3.751
& \best{0.723}
& 0.526
& \best{0.586}
& 0.538
& 1 \\
Wan 2.2-I2V
& 2.754
& 5.010
& 1.909
& \third{0.657}
& 3.746
& 0.610
& 0.510
& \third{0.553}
& 0.642
& 3 \\
SkyReels-V2-I2V
& \second{1.862}
& \second{2.225}
& \second{1.114}
& 0.710
& 3.771
& 0.528
& 0.505
& \second{0.556}
& 0.486
& 4 \\
CogVideoX 1.5
& \third{2.097}
& \third{3.897}
& 1.377
& 0.942
& 4.137
& \second{0.699}
& 0.496
& 0.524
& 0.499
& 5 \\
HunyuanVideo 1.5-I2V
& 6.522
& 15.890
& 4.076
& 1.734
& 4.915
& 0.448
& 0.483
& 0.381
& 0.464
& 12 \\
\midrule
\multicolumn{11}{@{}l}{\textit{Physical World Models}} \\
Cosmos 3 Nano
& 3.671
& 5.398
& 2.261
& 0.873
& 4.090
& 0.529
& \third{0.592}
& 0.526
& 0.748
& 7 \\
Vanilla Cosmos 3 Nano
& 3.006
& 4.868
& \third{1.321}
& \second{0.608}
& 3.790
& 0.560
& 0.515
& 0.547
& 0.500
& 8 \\
Cosmos-Dreams
& 4.292
& 10.360
& 2.376
& 0.698
& \best{3.644}
& 0.599
& 0.533
& 0.528
& 0.593
& 10 \\
\midrule
\multicolumn{11}{@{}l}{\textit{Driving World Models}} \\
DrivingWorld
& 2.849
& 5.171
& 1.731
& 0.829
& \second{3.669}
& \third{0.623}
& \best{0.658}
& 0.552
& \best{0.834}
& 2 \\
GEM
& 3.524
& 7.104
& 1.957
& 0.863
& \third{3.740}
& 0.515
& 0.571
& 0.515
& \second{0.793}
& 6 \\
Vista
& 4.218
& 6.624
& 2.195
& 0.755
& 3.830
& 0.486
& 0.574
& 0.535
& \third{0.760}
& 9 \\
Epona
& 7.620
& 10.784
& 5.398
& 1.326
& 4.271
& 0.282
& \second{0.625}
& 0.433
& 0.820
& 11 \\
\bottomrule
\end{tabular}%
}
\end{table*}

\begin{table*}[t!]
\caption{Right-lane-changing results of the VehDyn benchmark. All metrics are evaluated only on right single-lane-changing sequences. The final rank is obtained from the maneuver-specific VehDyn score: normalized trajectory-alignment, kinematic-consistency, and dynamic-consistency scores are summed to a value in $[0,3]$. Best results are in \redregion{red region}, second best in \orangeregion{orange region}, and third best in \blueregion{blue region}.}
\label{tab:rlc_right_results}
\centering
\resizebox{\textwidth}{!}{%
\small
\setlength{\tabcolsep}{4pt}
\begin{tabular}{@{}l|*{2}{c}|*{3}{c}|*{4}{c}|c@{}}
\toprule
\multirow{2}{*}{\makecell[l]{VehDyn Metrics\\Models}}
& \multicolumn{2}{c|}{Trajectory Alignment}
& \multicolumn{3}{c|}{Kinematic Consistency}
& \multicolumn{4}{c|}{Dynamic Consistency}
& \multirow{2}{*}{\makecell[c]{Rank}}\\
\cmidrule(lr){2-3}
\cmidrule(lr){4-6}
\cmidrule(lr){7-10}
& ADE $\downarrow$
& FDE $\downarrow$
& $v_y$ $\downarrow$
& Roll $\downarrow$
& Yaw $\downarrow$
& FCCR $\uparrow$
& CDC$_{\mathrm{Dir}}$ $\uparrow$
& CDC$_{\mathrm{Mag}}$ $\uparrow$
& CDC$_{\mathrm{Trend}}$ $\uparrow$
& \\
\midrule
\textcolor{gray}{Ground Truth}
& \textcolor{gray}{0.485}
& \textcolor{gray}{0.756}
& \textcolor{gray}{0.581}
& \textcolor{gray}{0.552}
& \textcolor{gray}{3.497}
& \textcolor{gray}{0.925}
& \textcolor{gray}{0.705}
& \textcolor{gray}{0.665}
& \textcolor{gray}{0.841}
& \textcolor{gray}{--}\\
\midrule
\multicolumn{11}{@{}l}{\textit{General World Models}} \\
LTX-Video 2.5
& \best{1.781}
& \best{2.711}
& \second{0.925}
& \second{0.568}
& 2.938
& \second{0.698}
& 0.505
& \second{0.550}
& 0.483
& 2 \\
Wan 2.2-I2V
& 2.711
& 4.975
& 1.187
& 0.658
& \best{2.808}
& 0.605
& 0.526
& \third{0.540}
& 0.582
& 4 \\
CogVideoX 1.5
& 2.772
& 4.996
& 1.164
& 1.000
& 2.951
& 0.681
& 0.520
& 0.486
& 0.528
& 7 \\
SkyReels-V2-I2V
& \third{2.443}
& \second{3.387}
& 1.139
& 0.939
& 2.839
& 0.531
& 0.477
& 0.491
& 0.552
& 8 \\
HunyuanVideo 1.5-I2V
& 8.080
& 17.885
& 4.730
& 2.009
& 3.756
& 0.501
& 0.482
& 0.341
& 0.478
& 12 \\
\midrule
\multicolumn{11}{@{}l}{\textit{Physical World Models}} \\
Cosmos 3 Nano
& 2.648
& 4.017
& 1.430
& 0.670
& 3.019
& 0.565
& \third{0.587}
& 0.526
& 0.723
& 3 \\
Vanilla Cosmos 3 Nano
& 2.814
& 4.974
& 1.255
& \third{0.571}
& 2.880
& 0.583
& 0.512
& 0.533
& 0.503
& 5 \\
Cosmos-Dreams
& 3.441
& 7.902
& \third{1.076}
& \best{0.472}
& 2.993
& \best{0.700}
& 0.488
& 0.522
& 0.447
& 9 \\
\midrule
\multicolumn{11}{@{}l}{\textit{Driving World Models}} \\
DrivingWorld
& \second{2.134}
& \third{3.495}
& \best{0.896}
& 0.653
& \third{2.832}
& \third{0.682}
& \best{0.645}
& \best{0.575}
& \second{0.843}
& 1 \\
GEM
& 4.207
& 6.556
& 1.984
& 0.741
& \second{2.821}
& 0.525
& 0.569
& 0.512
& \third{0.763}
& 6 \\
Vista
& 5.147
& 8.437
& 2.453
& 0.978
& 3.140
& 0.489
& 0.553
& 0.485
& 0.658
& 10 \\
Epona
& 7.496
& 9.878
& 5.140
& 1.224
& 2.985
& 0.315
& \second{0.629}
& 0.416
& \best{0.850}
& 11 \\
\bottomrule
\end{tabular}%
}
\end{table*}

\paragraph{Trajectory alignment.}

As shown in Table \ref{tab:normalized_trajectory_alignment}, LTX-Video 2.5 obtains the highest normalized trajectory-alignment score among all models at 0.945, followed by DrivingWorld at 0.931 and Cosmos 3 Nano at 0.917, compared with 0.984 for Ground Truth. 
Most evaluated models remain relatively close in this dimension, indicating that video world models are generally capable of reproducing the coarse planar displacement of the ego vehicle. 
LTX-Video 2.5 ranks first in trajectory alignment but only third in the full VehDyn benchmark, whereas DrivingWorld ranks second in trajectory alignment but first overall. 
A generated sequence can follow the intended planar path while still producing inaccurate velocity, attitude, or condition-dependent physical responses. 
Trajectory alignment remains necessary for verifying where the vehicle moves, but it provides limited discrimination once the coarse ego-motion pattern has been reproduced.

\begin{table*}[ht!]
\caption{Normalized trajectory-alignment results. Best results are in \redregion{red region}, second best in \orangeregion{orange region}, and third best in \blueregion{blue region}. Models fall into three categories: general world models, physical world models, and driving world models.
}
\label{tab:normalized_trajectory_alignment}
\centering
\small
\setlength{\tabcolsep}{4pt}
\begin{tabular}{@{}l|*{2}{c}|c|c@{}}
\toprule
\multirow{3}{*}{\makecell[l]{Models}} & \multicolumn{2}{c|}{Trajectory Alignment} & \multirow{3}{*}{\makecell[c]{Normalized\\Score}} & \multirow{3}{*}{\makecell[c]{Rank}}\\
\cmidrule(lr){2-3}
& \multicolumn{2}{c|}{All Maneuvers} & & \\
\cmidrule(lr){2-3}
& ADE $\uparrow$ & FDE $\uparrow$ & & \\
\midrule
\textcolor{gray}{Ground Truth} & \textcolor{gray}{0.984} & \textcolor{gray}{0.985} & \textcolor{gray}{0.984} & \textcolor{gray}{--} \\
\midrule
\multicolumn{5}{@{}l}{\textit{General World Models}} \\
LTX-Video 2.5 & \best{0.936} & \best{0.954} & \best{0.945} & 1 \\
SkyReels-V2-I2V & 0.886 & 0.904 & 0.895 & 5 \\
CogVideoX 1.5 & 0.888 & 0.892 & 0.890 & 6 \\
Wan 2.2-I2V & 0.854 & 0.881 & 0.867 & 8 \\
HunyuanVideo 1.5-I2V & 0.719 & 0.707 & 0.713 & 12 \\
\midrule
\multicolumn{5}{@{}l}{\textit{Physical World Models}} \\
Cosmos 3 Nano & \third{0.903} & \third{0.930} & \third{0.917} & 3 \\
Vanilla Cosmos 3 Nano & 0.891 & 0.910 & 0.900 & 4 \\
Cosmos-Dreams & 0.850 & 0.826 & 0.838 & 10 \\
\midrule
\multicolumn{5}{@{}l}{\textit{Driving World Models}} \\
DrivingWorld & \second{0.925} & \second{0.936} & \second{0.931} & 2 \\
GEM & 0.875 & 0.892 & 0.884 & 7 \\
Vista & 0.830 & 0.865 & 0.848 & 9 \\
Epona & 0.763 & 0.842 & 0.802 & 11 \\
\bottomrule
\end{tabular}%
\end{table*}

\paragraph{Kinematic consistency.}

Table \ref{tab:normalized_kinematic_consistency} shows that DrivingWorld achieves the highest normalized kinematic-consistency score of 0.853, followed closely by LTX-Video 2.5 at 0.845 and Cosmos 3 Nano at 0.835. 
However, no single model dominates every kinematic variable. 
DrivingWorld is strongest in longitudinal velocity and yaw, whereas LTX-Video 2.5 performs best in pitch, lateral velocity, and roll. 
The state-level results also expose maneuver-specific weaknesses that are obscured by the aggregate score. 
Epona, for example, remains competitive in the emergency-braking variables, with strong normalized longitudinal-velocity and pitch scores, but degrades markedly in lateral velocity and roll during lane-changing. 
Conversely, several general and physical world models preserve lateral and rotational states well despite weaker braking dynamics. 
DrivingWorld's normalized kinematic score slightly exceeds the ground-truth score, which should consequently be interpreted as estimator effects.

\begin{table*}[ht!]
\caption{Normalized kinematic-consistency results. Best results are in \redregion{red region}, second best in \orangeregion{orange region}, and third best in \blueregion{blue region}. Models fall into three categories: general world models, physical world models, and driving world models.
}
\label{tab:normalized_kinematic_consistency}
\centering
\small
\setlength{\tabcolsep}{4pt}
\begin{tabular}{@{}l|*{5}{c}|c|c@{}}
\toprule
\multirow{3}{*}{\makecell[l]{Models}} & \multicolumn{5}{c|}{Kinematic Consistency} & \multirow{3}{*}{\makecell[c]{Normalized\\Score}} & \multirow{3}{*}{\makecell[c]{Rank}}\\
\cmidrule(lr){2-6}
& \multicolumn{2}{c}{Emergency-Braking} & \multicolumn{3}{c|}{Single-Lane-Changing} & & \\
\cmidrule(lr){2-3} \cmidrule(lr){4-6}
& $v_x$ $\uparrow$ & Pitch $\uparrow$ & $v_y$ $\uparrow$ & Roll $\uparrow$ & Yaw $\uparrow$ & & \\
\midrule
\textcolor{gray}{Ground Truth} & \textcolor{gray}{0.968} & \textcolor{gray}{0.786} & \textcolor{gray}{0.971} & \textcolor{gray}{0.895} & \textcolor{gray}{0.617} & \textcolor{gray}{0.847} & \textcolor{gray}{--} \\
\midrule
\multicolumn{8}{@{}l}{\textit{General World Models}} \\
LTX-Video 2.5 & 0.852 & \best{0.834} & \best{0.961} & \best{0.903} & 0.674 & \second{0.845} & 2 \\
SkyReels-V2-I2V & 0.652 & 0.797 & \second{0.952} & 0.853 & 0.678 & 0.786 & 8 \\
CogVideoX 1.5 & 0.729 & 0.733 & \third{0.946} & 0.826 & 0.652 & 0.777 & 10 \\
Wan 2.2-I2V & 0.561 & 0.825 & 0.933 & 0.885 & \second{0.679} & 0.777 & 11 \\
HunyuanVideo 1.5-I2V & 0.640 & 0.704 & 0.809 & 0.658 & 0.571 & 0.676 & 12 \\
\midrule
\multicolumn{8}{@{}l}{\textit{Physical World Models}} \\
Cosmos 3 Nano & \second{0.919} & 0.823 & 0.921 & 0.863 & 0.651 & \third{0.835} & 3 \\
Vanilla Cosmos 3 Nano & 0.792 & 0.820 & 0.945 & \third{0.897} & 0.673 & 0.826 & 4 \\
Cosmos-Dreams & 0.716 & 0.825 & 0.925 & \second{0.898} & 0.674 & 0.808 & 7 \\
\midrule
\multicolumn{8}{@{}l}{\textit{Driving World Models}} \\
DrivingWorld & \best{0.942} & \third{0.827} & 0.944 & 0.869 & \best{0.682} & \best{0.853} & 1 \\
GEM & 0.855 & 0.799 & 0.915 & 0.858 & \second{0.679} & 0.821 & 5 \\
Vista & 0.830 & 0.812 & 0.900 & 0.845 & 0.659 & 0.809 & 6 \\
Epona & \third{0.889} & \second{0.828} & 0.771 & 0.769 & 0.644 & 0.780 & 9 \\
\bottomrule
\end{tabular}%
\end{table*}

\paragraph{Dynamic consistency.}

The results in Table \ref{tab:normalized_dynamic_consistency} indicate that DrivingWorld ranks first with a normalized score of 0.718, followed by Cosmos 3 Nano at 0.650 and GEM at 0.633. 
The four dynamic sub-metrics capture two complementary requirements: $FCCR$ measures instantaneous physical feasibility under the friction-circle constraint, while $CDC_{\mathrm{Dir}}$, $CDC_{\mathrm{Mag}}$, and $CDC_{\mathrm{Trend}}$ measure whether the generated vehicle responds to changes in physical conditions in the same direction, magnitude, and overall trend as the reference dynamics. 
CogVideoX 1.5 achieves the highest $FCCR$ of 0.738 but only a 0.610 aggregate dynamic-consistency score because its condition-aware $CDC$ scores are substantially lower. 
In contrast, Epona achieves the highest $CDC_{\mathrm{Trend}}$ of 0.844 and the second-highest $CDC_{\mathrm{Dir}}$ of 0.672, but its $FCCR$ is only 0.394. 
Its responses preserve part of the correct cross-condition ordering while frequently violating the adhesion-based feasibility constraint. 
DrivingWorld is the only model that remains near the top across all four dynamic sub-metrics, which explains its strong overall performance. 
The comparison between Cosmos 3 Nano and Vanilla Cosmos 3 Nano further shows that post-training improves the full VehDyn score from 2.307 to 2.402, with most of the gain concentrated in dynamic consistency, which increases from 0.581 to 0.650.

\begin{table*}[ht!]
\caption{Normalized dynamic-consistency results. Best results are in \redregion{red region}, second best in \orangeregion{orange region}, and third best in \blueregion{blue region}. Models fall into three categories: general world models, physical world models, and driving world models.
}
\label{tab:normalized_dynamic_consistency}
\centering
\small
\setlength{\tabcolsep}{4pt}
\begin{tabular}{@{}l|*{4}{c}|c|c@{}}
\toprule
\multirow{3}{*}{\makecell[l]{Models}} & \multicolumn{4}{c|}{Dynamic Consistency} & \multirow{3}{*}{\makecell[c]{Normalized\\Score}} & \multirow{3}{*}{\makecell[c]{Rank}}\\
\cmidrule(lr){2-5}
& \multicolumn{1}{c}{All Maneuvers} & \multicolumn{3}{c|}{All Maneuvers} & & \\
\cmidrule(lr){2-2} \cmidrule(lr){3-5}
& FCCR $\uparrow$ & CDC$_{\mathrm{Dir}}$ $\uparrow$ & CDC$_{\mathrm{Mag}}$ $\uparrow$ & CDC$_{\mathrm{Trend}}$ $\uparrow$ & & \\
\midrule
\textcolor{gray}{Ground Truth} & \textcolor{gray}{0.900} & \textcolor{gray}{0.722} & \textcolor{gray}{0.713} & \textcolor{gray}{0.803} & \textcolor{gray}{0.784} & \textcolor{gray}{--} \\
\midrule
\multicolumn{7}{@{}l}{\textit{General World Models}} \\
CogVideoX 1.5 & \best{0.738} & 0.582 & 0.557 & 0.565 & 0.610 & 4 \\
Wan 2.2-I2V & 0.595 & 0.571 & 0.602 & 0.643 & 0.603 & 6 \\
LTX-Video 2.5 & \second{0.686} & 0.563 & \third{0.610} & 0.539 & 0.599 & 8 \\
SkyReels-V2-I2V & 0.569 & 0.577 & 0.584 & 0.590 & 0.580 & 11 \\
HunyuanVideo 1.5-I2V & 0.488 & 0.528 & 0.436 & 0.485 & 0.484 & 12 \\
\midrule
\multicolumn{7}{@{}l}{\textit{Physical World Models}} \\
Cosmos 3 Nano & 0.591 & \third{0.646} & \second{0.617} & 0.746 & \second{0.650} & 2 \\
Cosmos-Dreams & 0.675 & 0.566 & 0.572 & 0.576 & 0.597 & 9 \\
Vanilla Cosmos 3 Nano & 0.601 & 0.581 & 0.599 & 0.543 & 0.581 & 10 \\
\midrule
\multicolumn{7}{@{}l}{\textit{Driving World Models}} \\
DrivingWorld & \third{0.685} & \best{0.708} & \best{0.664} & \second{0.817} & \best{0.718} & 1 \\
GEM & 0.549 & 0.624 & 0.591 & \third{0.769} & \third{0.633} & 3 \\
Epona & 0.394 & \second{0.672} & 0.528 & \best{0.844} & 0.610 & 5 \\
Vista & 0.533 & 0.604 & 0.568 & 0.706 & 0.603 & 7 \\
\bottomrule
\end{tabular}%
\end{table*}

\paragraph{Overall score.}

It can be seen from Table \ref{tab:vehicle_dynamic_score} that DrivingWorld achieves the highest score of 2.502, corresponding to 95.6\% of the ground-truth score of 2.616. 
It is followed by Cosmos 3 Nano at 2.402, LTX-Video 2.5 at 2.389, and GEM at 2.338. 
Relative to LTX-Video 2.5, DrivingWorld is slightly lower in trajectory alignment and only marginally higher in kinematic consistency, but its substantially stronger dynamic-consistency score produces a clear advantage in the final VehDyn score. 
Similarly, Cosmos 3 Nano trails LTX-Video 2.5 in both trajectory alignment and kinematic consistency, but its stronger dynamic performance is sufficient to place it slightly higher overall.
VehDyn rewards balanced physical performance rather than specialization in a single dimension. 

\begin{table*}[ht!]
\caption{Overall VehDyn score. Best results are in \redregion{red region}, second best in \orangeregion{orange region}, and third best in \blueregion{blue region}. Models fall into three categories: general world models, physical world models, and driving world models.
}
\label{tab:vehicle_dynamic_score}
\centering
\small
\setlength{\tabcolsep}{4pt}
\begin{tabular}{@{}l|*{3}{c}|c|c@{}}
\toprule
\multirow{2}{*}{\makecell[l]{Models}} & \multicolumn{3}{c|}{VehDyn Benchmark} & \multirow{2}{*}{\makecell[c]{VehDyn\\Score}} & \multirow{2}{*}{\makecell[c]{Rank}}\\
\cmidrule(lr){2-4}
& \makecell{Trajectory\\Alignment $\uparrow$} & \makecell{Kinematic\\Consistency $\uparrow$} & \makecell{Dynamic\\Consistency $\uparrow$} & & \\
\midrule
\textcolor{gray}{Ground Truth} & \textcolor{gray}{0.984} & \textcolor{gray}{0.847} & \textcolor{gray}{0.784} & \textcolor{gray}{2.616} & \textcolor{gray}{--} \\
\midrule
\multicolumn{6}{@{}l}{\textit{General World Models}} \\
LTX-Video 2.5 & \best{0.945} & \second{0.845} & 0.599 & \third{2.389} & 3 \\
CogVideoX 1.5 & 0.890 & 0.777 & 0.610 & 2.277 & 6 \\
SkyReels-V2-I2V & 0.895 & 0.786 & 0.580 & 2.261 & 7 \\
Wan 2.2-I2V & 0.867 & 0.777 & 0.603 & 2.247 & 9 \\
HunyuanVideo 1.5-I2V & 0.713 & 0.676 & 0.484 & 1.873 & 12 \\
\midrule
\multicolumn{6}{@{}l}{\textit{Physical World Models}} \\
Cosmos 3 Nano & \third{0.917} & \third{0.835} & \second{0.650} & \second{2.402} & 2 \\
Vanilla Cosmos 3 Nano & 0.900 & 0.826 & 0.581 & 2.307 & 5 \\
Cosmos-Dreams & 0.838 & 0.808 & 0.597 & 2.243 & 10 \\
\midrule
\multicolumn{6}{@{}l}{\textit{Driving World Models}} \\
DrivingWorld & \second{0.931} & \best{0.853} & \best{0.718} & \best{2.502} & 1 \\
GEM & 0.884 & 0.821 & \third{0.633} & 2.338 & 4 \\
Vista & 0.848 & 0.809 & 0.603 & 2.260 & 8 \\
Epona & 0.802 & 0.780 & 0.610 & 2.192 & 11 \\
\bottomrule
\end{tabular}%
\end{table*}

\subsubsection{Complementary Visual Evaluation Scores}

Table \ref{tab:normalized_visual_evaluation} reports the complementary visual evaluation under WorldScore and DrivingGen with the rank obtained from the sum of the normalized Quality, Distribution, and Temporal Consistency level scores. 
In contrast to the VehDyn ranking, the conventional visual metrics favor the general world models. 
LTX-Video 2.5 ranks first (2.181), followed by SkyReels-V2-I2V (2.109) and Wan 2.2-I2V (2.050); the physical world models occupy the middle of the ranking, with Cosmos 3 Nano fourth and Cosmos-Dreams fifth; and the four driving world models occupy the last four positions, with DrivingWorld, first on VehDyn, ninth here.
This result indicates that strong visual generation quality does not necessarily coincide with strong vehicle-dynamics fidelity. 
In particular, the general world models tend to preserve appearance, temporal smoothness, and distribution-level realism well, whereas driving world models do not consistently obtain higher scores under conventional visual criteria.

\begin{table*}[ht!]
\caption{Normalized visual evaluation results. Best results are in \redregion{red region}, second best in \orangeregion{orange region}, and third best in \blueregion{blue region}. Models fall into three categories: general world models, physical world models, and driving world models.
}
\label{tab:normalized_visual_evaluation}
\centering
\resizebox{\textwidth}{!}{%
\small
\setlength{\tabcolsep}{9.5pt}
\begin{tabular}{@{}l|*{4}{c}|*{3}{c}|c@{}}
\toprule
\multirow{3}{*}{\makecell[l]{Other Benchmarks\\\\Models}}
& \multicolumn{4}{c|}{WorldScore} &
\multicolumn{3}{c|}{DrivingGen} &
\multirow{3}{*}{\makecell[c]{Rank}}\\
\cmidrule(lr){2-5} \cmidrule(lr){6-8}
& \multicolumn{4}{c|}{Quality}
& \multicolumn{1}{c}{Distribution}
& \multicolumn{2}{c|}{Temporal Consistency}
& \\
\cmidrule(lr){2-5} \cmidrule(lr){6-6} \cmidrule(lr){7-8}
& \makecell{3D\\Consistency $\uparrow$}
& \makecell{Photometric\\Consistency $\uparrow$}
& \makecell{Style\\Consistency $\uparrow$}
& \makecell{Subjective\\Quality $\uparrow$}
& \makecell{FVD\\Score $\uparrow$}
& \makecell{Video\\Consistency $\uparrow$}
& \makecell{Trajectory\\Consistency $\uparrow$}
& \\
\midrule

\textcolor{gray}{Ground Truth}
& \textcolor{gray}{0.7932}
& \textcolor{gray}{0.1988}
& \textcolor{gray}{0.9471}
& \textcolor{gray}{0.4284}
& \textcolor{gray}{1.0000}
& \textcolor{gray}{0.9102}
& \textcolor{gray}{0.4469}
& \textcolor{gray}{--}\\

\midrule
\multicolumn{9}{@{}l}{\textit{General World Models}} \\

LTX-Video 2.5
& \best{0.7551}
& \second{0.2736}
& {0.9446}
& {0.2999}
& \best{0.9196}
& \third{0.9156}
& {0.4709}
& 1\\

SkyReels-V2-I2V
& \second{0.7190}
& \best{0.6036}
& {0.8878}
& \second{0.4085}
& {0.7431}
& {0.9094}
& \best{0.5130}
& 2\\

Wan 2.2-I2V
& {0.6637}
& \third{0.1858}
& {0.9555}
& {0.3871}
& \third{0.8450}
& \second{0.9206}
& {0.3936}
& 3\\

HunyuanVideo 1.5-I2V
& {0.5839}
& {0.1601}
& {0.9450}
& {0.3343}
& {0.6519}
& {0.8771}
& {0.3726}
& 6\\

CogVideoX 1.5
& {0.5833}
& {0.0000}
& {0.9486}
& {0.3190}
& {0.3684}
& {0.9030}
& \second{0.4752}
& 8\\

\midrule
\multicolumn{9}{@{}l}{\textit{Physical World Models}} \\

Cosmos 3 Nano
& \third{0.6761}
& {0.0992}
& \second{0.9617}
& \third{0.3940}
& {0.7183}
& {0.8742}
& \third{0.4744}
& 4\\

Cosmos-Dreams
& {0.6623}
& {0.1251}
& {0.7628}
& {0.0578}
& \second{0.8811}
& {0.7747}
& {0.3981}
& 5\\

Vanilla Cosmos 3 Nano
& {0.6445}
& {0.0752}
& {0.9562}
& \best{0.4092}
& {0.5533}
& {0.9125}
& {0.4374}
& 7\\

\midrule
\multicolumn{9}{@{}l}{\textit{Driving World Models}} \\

DrivingWorld
& {0.5079}
& {0.1481}
& \best{0.9932}
& {0.1008}
& {0.4064}
& {0.8982}
& {0.3637}
& 9\\

Epona
& {0.4894}
& {0.0072}
& \third{0.9605}
& {0.3300}
& {0.3437}
& {0.9126}
& {0.3701}
& 10\\

GEM
& {0.5473}
& {0.0739}
& {0.7292}
& {0.1662}
& {0.3434}
& {0.8958}
& {0.4138}
& 11\\

Vista
& {0.4916}
& {0.0596}
& {0.8614}
& {0.2213}
& {0.2668}
& \best{0.9281}
& {0.4475}
& 12\\

\bottomrule
\end{tabular}%
}
\end{table*}

LTX-Video 2.5 achieves the strongest overall visual performance through the highest 3D consistency (0.7551) and FVD score (0.9196), together with strong style consistency (0.9446) and the third-highest video consistency (0.9156), although its subjective quality (0.2999) is below that of most general models.
SkyReels-V2-I2V ranks second with a more balanced profile: it obtains the highest photometric consistency (0.6036) and trajectory consistency (0.5130) and the second-highest 3D consistency (0.7190) and subjective quality (0.4085), but its FVD score (0.7431) is the lowest among the top three.
Wan 2.2-I2V ranks third, benefiting from high style consistency (0.9555), the third-highest FVD score (0.8450), and the second-highest video consistency (0.9206).
These results illustrate that similar overall visual rankings can arise from different combinations of spatial fidelity, appearance preservation, distribution similarity, and temporal stability.

Cosmos 3 Nano ranks fourth with a comparatively balanced result, including the second-highest style consistency (0.9617), the third-highest subjective quality (0.3940), and the third-highest trajectory consistency (0.4744).
Vanilla Cosmos 3 Nano ranks seventh and achieves the highest subjective quality among the generated videos (0.4092), but is weaker in photometric consistency (0.0752) and distribution similarity (FVD score 0.5533).
Post-training on VehDyn improves 3D consistency from 0.6445 to 0.6761, the FVD score from 0.5533 to 0.7183, and trajectory consistency from 0.4374 to 0.4744, while video consistency (0.9125 to 0.8742) and subjective quality (0.4092 to 0.3940) decrease slightly.
This suggests that adaptation toward the VehDyn data distribution improves several structural and distribution-level properties without necessarily producing a uniform gain in perceptual quality.
Cosmos-Dreams achieves the second-highest FVD score (0.8811), but its style consistency (0.7628), subjective quality (0.0578), and video consistency (0.7747) are substantially below those of the leading models.
Therefore, a strong distribution-level score does not guarantee high per-video perceptual or temporal quality. 
More generally, the considerable variation among the WorldScore and DrivingGen sub-metrics suggests that no individual visual metric is sufficient to characterize the overall quality of generated driving videos.

DrivingWorld achieves the highest style consistency of all evaluated models (0.9932), but ranks only ninth because its 3D consistency (0.5079), subjective quality (0.1008), FVD score (0.4064), and trajectory consistency (0.3637) are comparatively weak.
Epona similarly combines high style consistency (0.9605) and video consistency (0.9126) with very low photometric consistency (0.0072) and FVD score (0.3437).
Vista achieves the highest video consistency (0.9281) and a trajectory consistency of 0.4475, approximately matching the ground-truth value (0.4469), but ranks last because its spatial, photometric, and distribution-level scores remain low, with the lowest FVD score of any model (0.2668).
These cases show that temporal smoothness alone can coexist with substantial deficiencies in visual structure and distribution fidelity.
In addition, a generated sequence can be smoother or more temporally regular than the reference sequence and obtain a higher consistency score while still exhibiting less accurate vehicle motion.

SkyReels-V2-I2V ranks second under the complementary visual evaluation but seventh under VehDyn, and GEM ranks eleventh visually but fourth under VehDyn, whereas DrivingWorld ranks ninth visually but first in vehicle-dynamics fidelity.
Only LTX-Video 2.5 and Cosmos 3 Nano perform strongly under both evaluations (first and fourth visually; third and second on VehDyn), and even their relative order changes.
The Spearman correlation between the two rankings across the 12 models is $\rho_S = 0.39$.
These rank reversals demonstrate that conventional visual-quality metrics and vehicle-dynamics metrics capture fundamentally different properties of a driving world model. 
Visual metrics assess whether generated videos appear coherent, stable, and distributionally realistic, whereas VehDyn evaluates whether the underlying vehicle motion follows the intended trajectory, reproduces the corresponding kinematic states, and remains dynamically consistent under changes in physical conditions.

\subsection{Supplementary Details about Qualitative Results}
\label{appendix:qualitative}

We provide additional qualitative rollouts under controlled changes in the target driving condition. Unless varied, the reference setting is $72$~km/h, Model 3, wet stone road with $\mu=0.50$, and a $75^\circ$ sun angle. 
Each panel contains five frames from the first $5$ s and the corresponding state curves. 
For the emergency-braking scenario, we show the 2D trajectory, longitudinal velocity, and pitch angle; and for the single-lane-changing scenario, we show the 2D trajectory, lateral velocity, and roll angle.

\begin{figure*}[p]
    \centering
    \includegraphics[width=\textwidth]{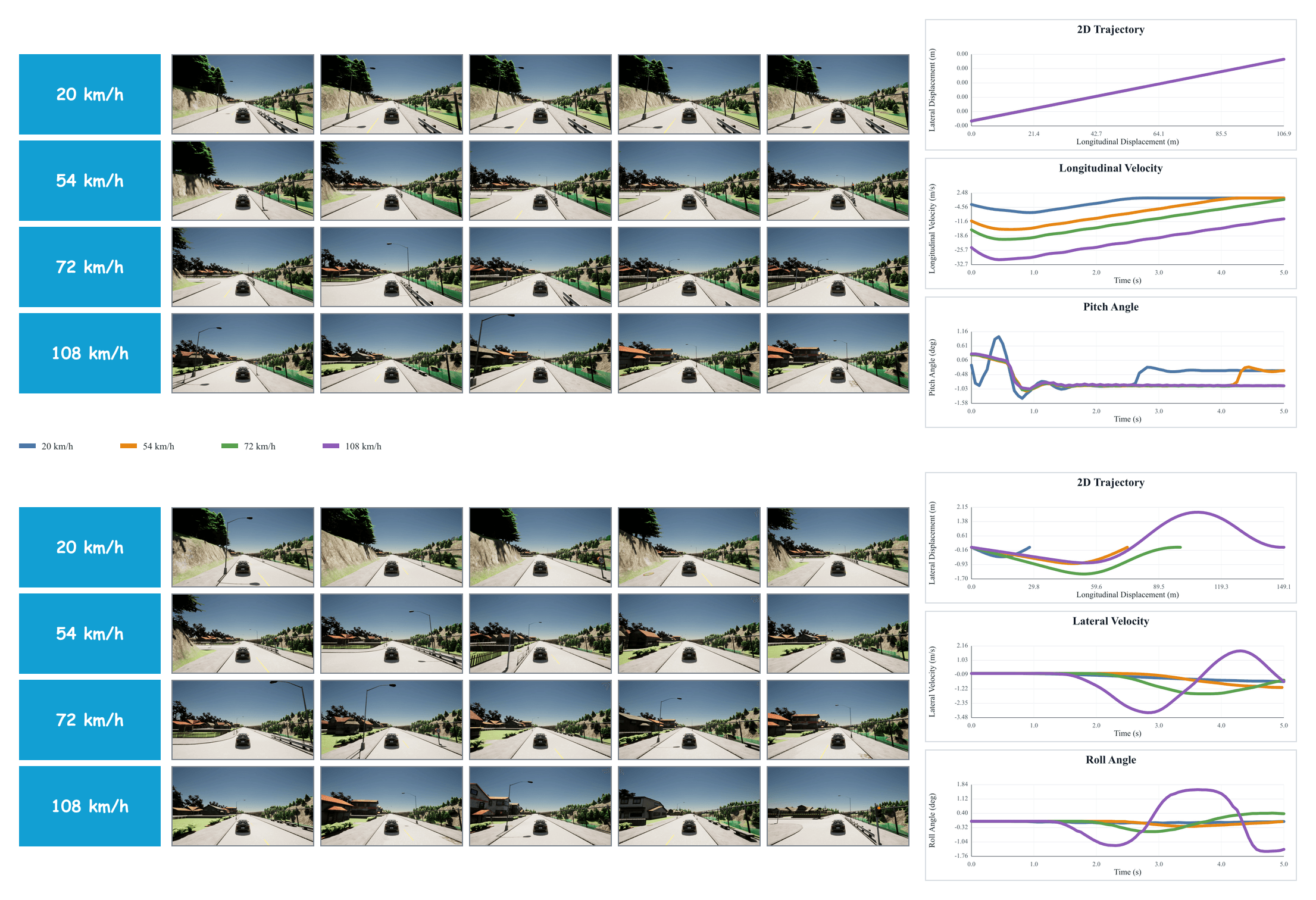}
    \caption{Qualitative target-speed sweep: Ground Truth. Top: Emergency-braking scenario; Bottom: Single-lane-changing scenario.}
    \label{fig:app-speed-gt}
\end{figure*}

\begin{figure*}[p]
    \centering
    \includegraphics[width=\textwidth]{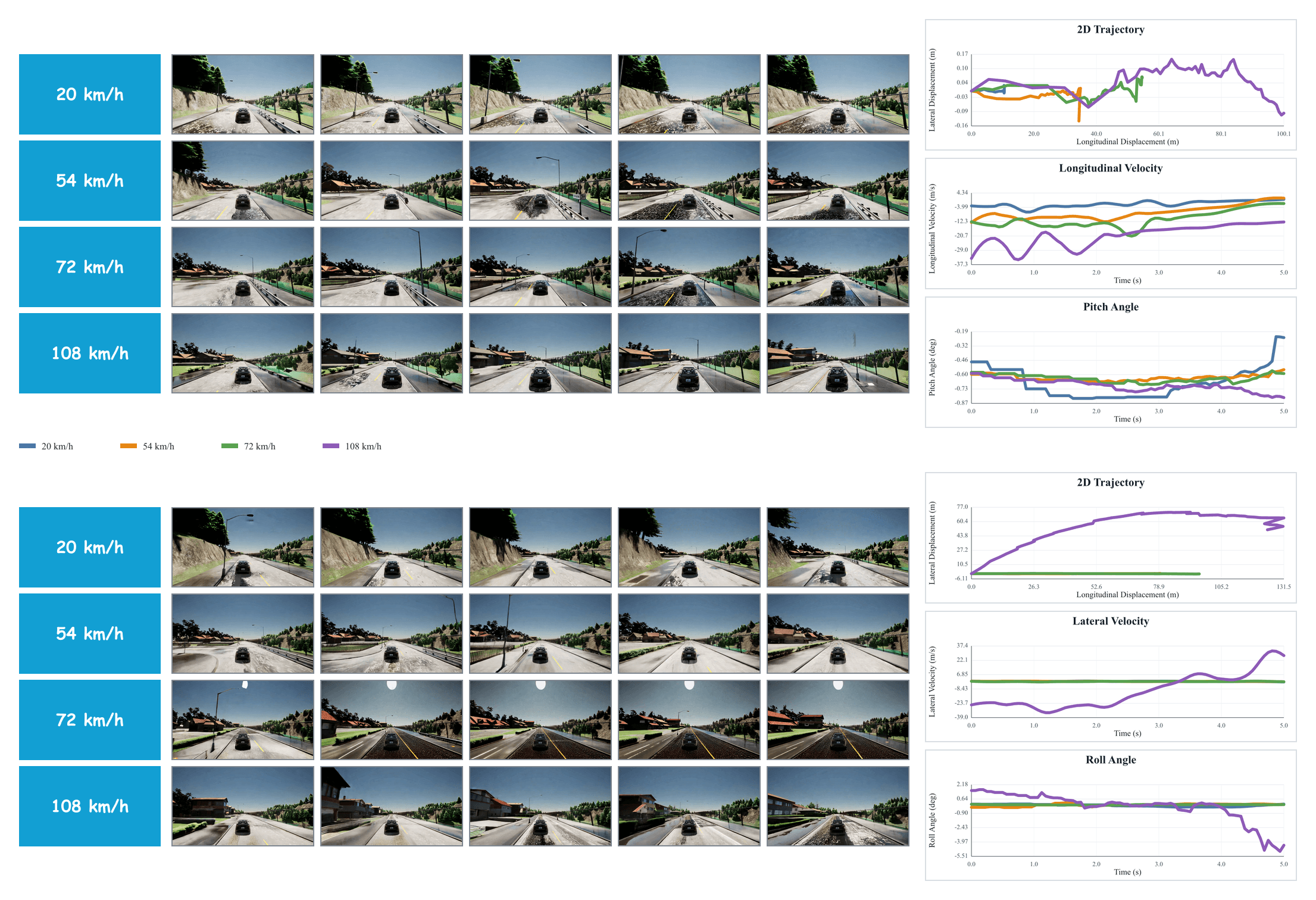}
    \caption{Qualitative target-speed sweep: Cosmos 3 Nano. Top: Emergency-braking scenario; Bottom: Single-lane-changing scenario.}
    \label{fig:app-speed-cosmos}
\end{figure*}

\begin{figure*}[p]
    \centering
    \includegraphics[width=\textwidth]{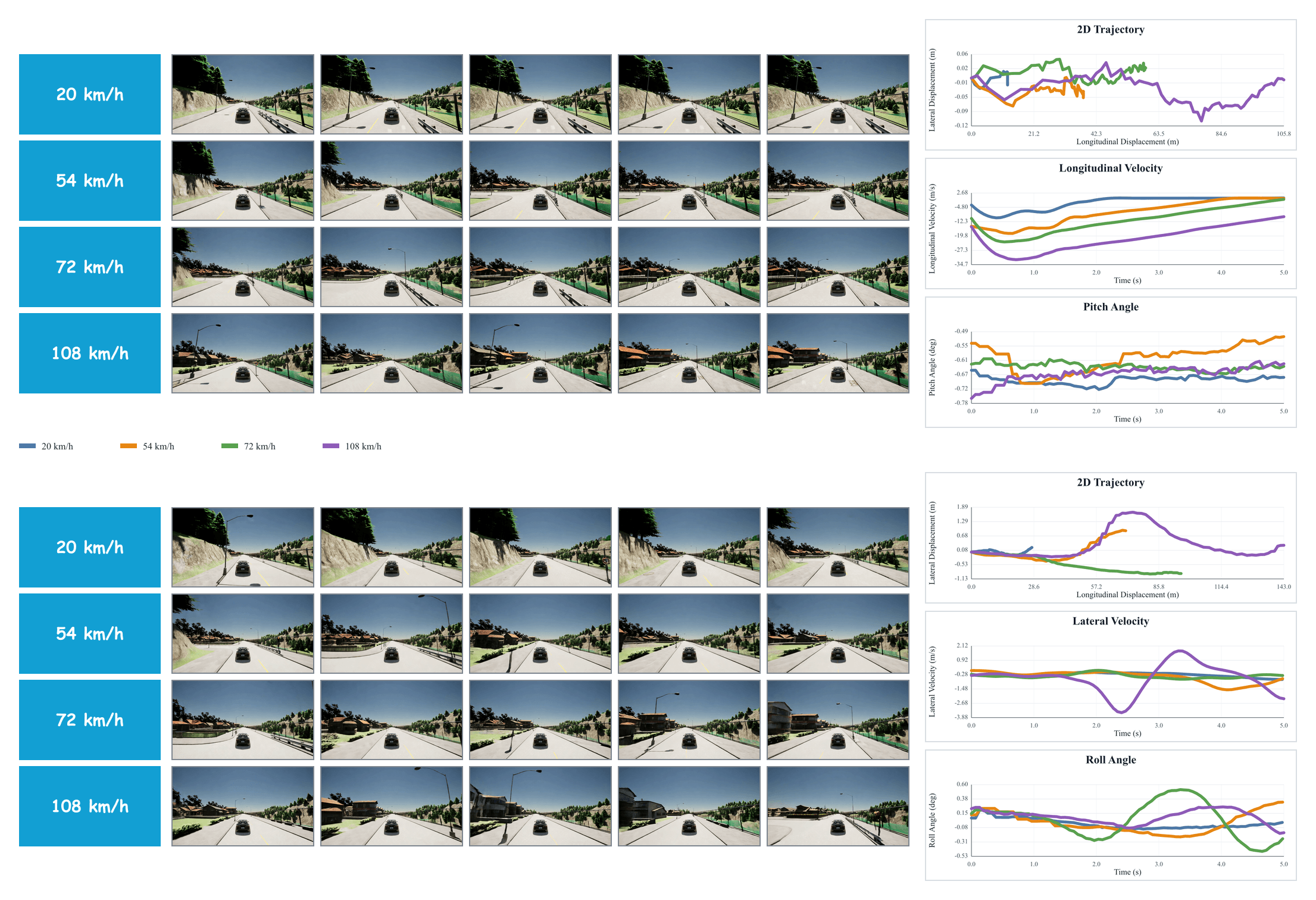}
    \caption{Qualitative target-speed sweep: Vanilla Cosmos 3 Nano. Top: Emergency-braking scenario; Bottom: Single-lane-changing scenario.}
    \label{fig:app-speed-cosmos-post}
\end{figure*}

\begin{figure*}[p]
    \centering
    \includegraphics[width=\textwidth]{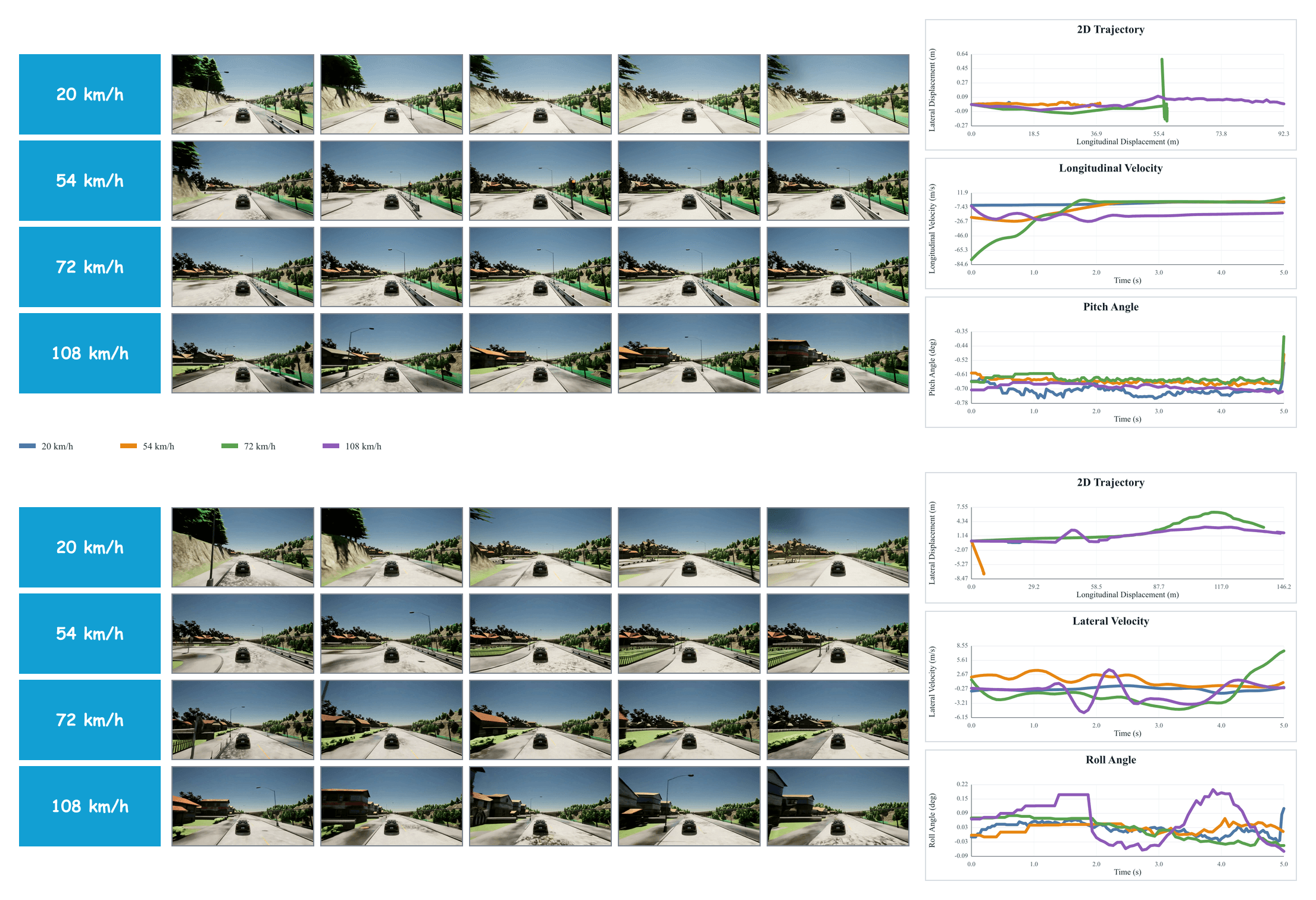}
    \caption{Qualitative target-speed sweep: Cosmos-Dreams. Top: Emergency-braking scenario; Bottom: Single-lane-changing scenario.}
    \label{fig:app-speed-cosmos-dreams}
\end{figure*}

\begin{figure*}[p]
    \centering
    \includegraphics[width=\textwidth]{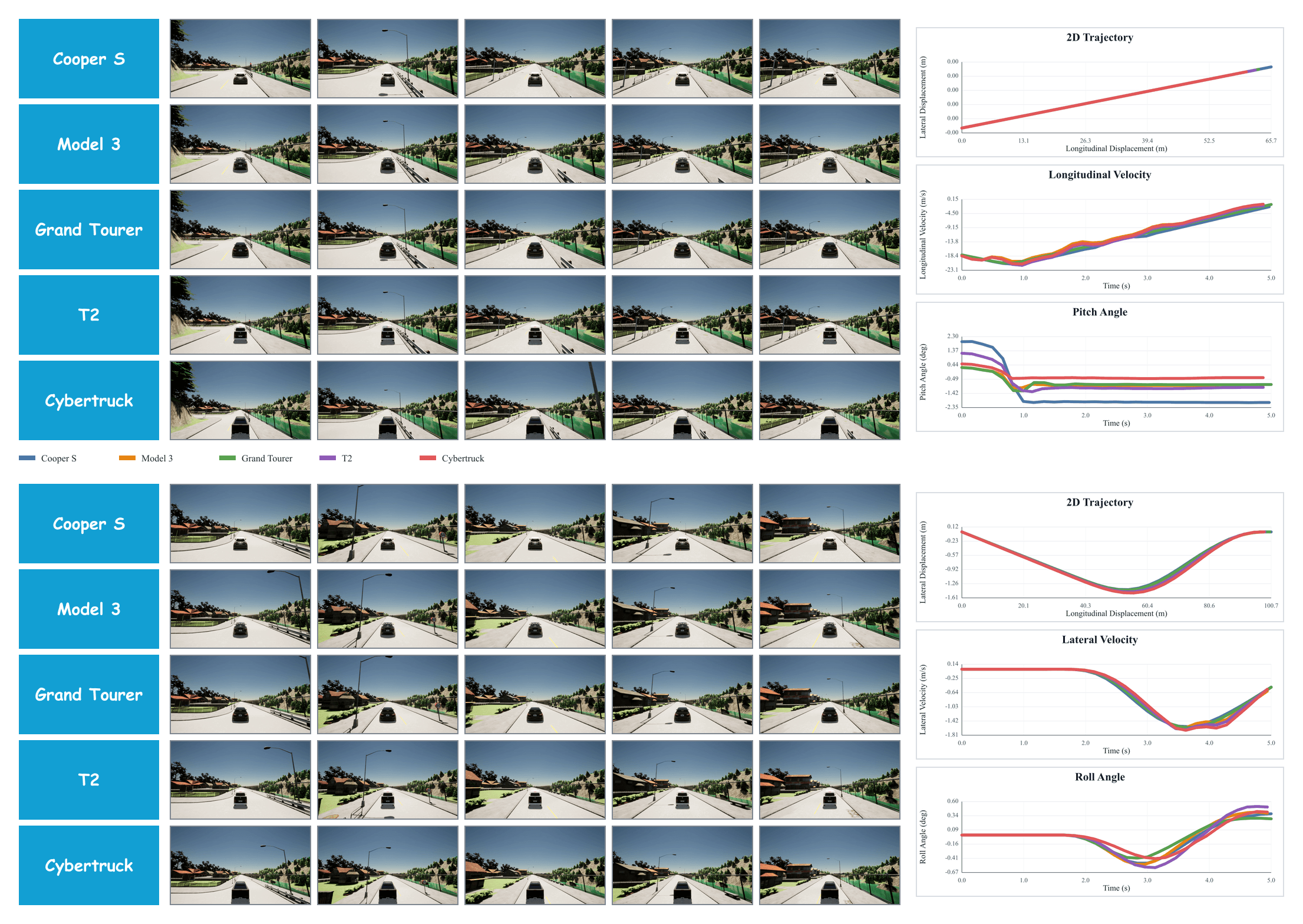}
    \caption{Qualitative vehicle-type sweep: Ground Truth. Top: Emergency-braking scenario; Bottom: Single-lane-changing scenario.}
    \label{fig:app-vehicle-gt}
\end{figure*}

\begin{figure*}[p]
    \centering
    \includegraphics[width=\textwidth]{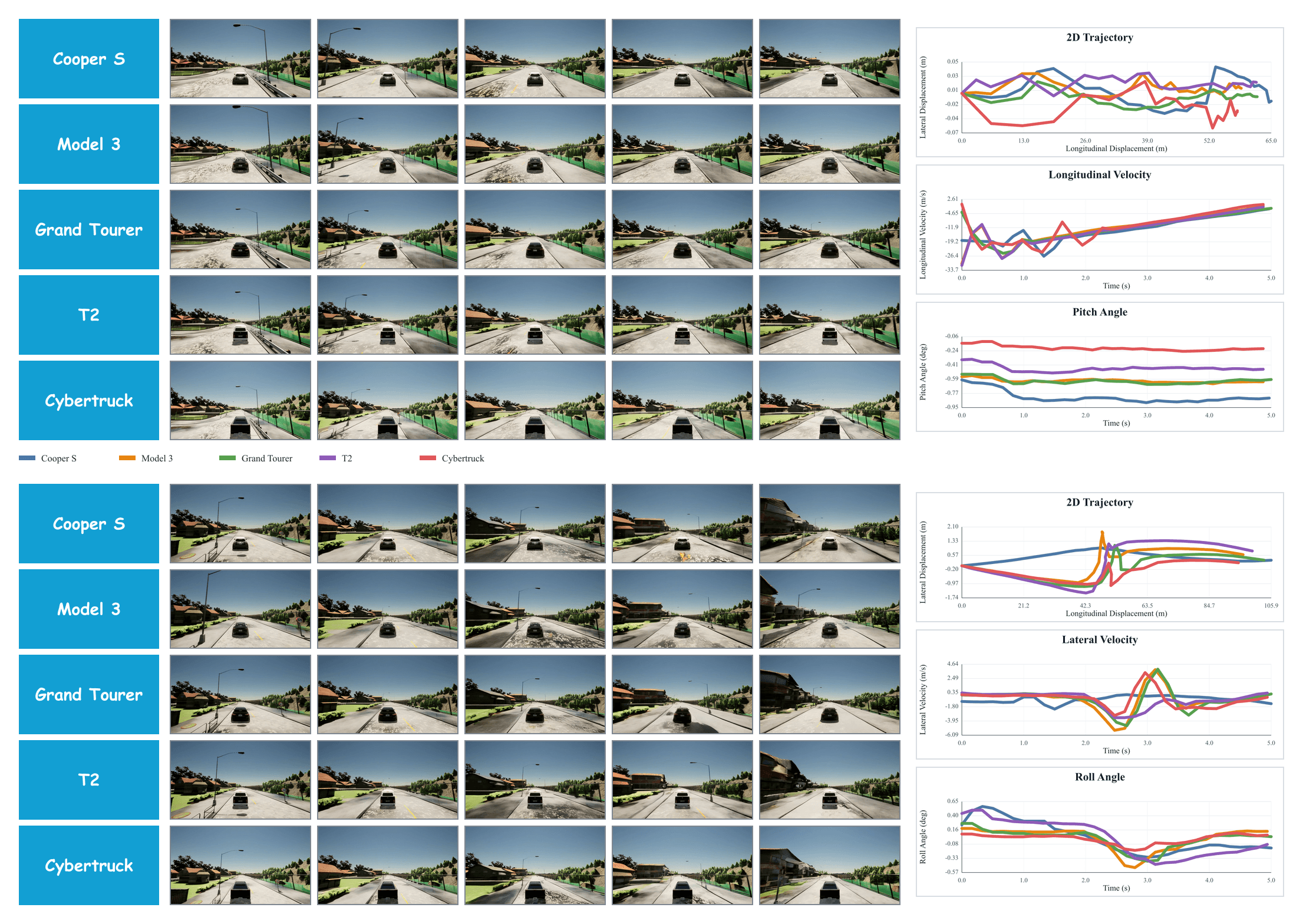}
    \caption{Qualitative vehicle-type sweep: Epona. Top: Emergency-braking scenario; Bottom: Single-lane-changing scenario.}
    \label{fig:app-vehicle-epona}
\end{figure*}

\begin{figure*}[p]
    \centering
    \includegraphics[width=\textwidth]{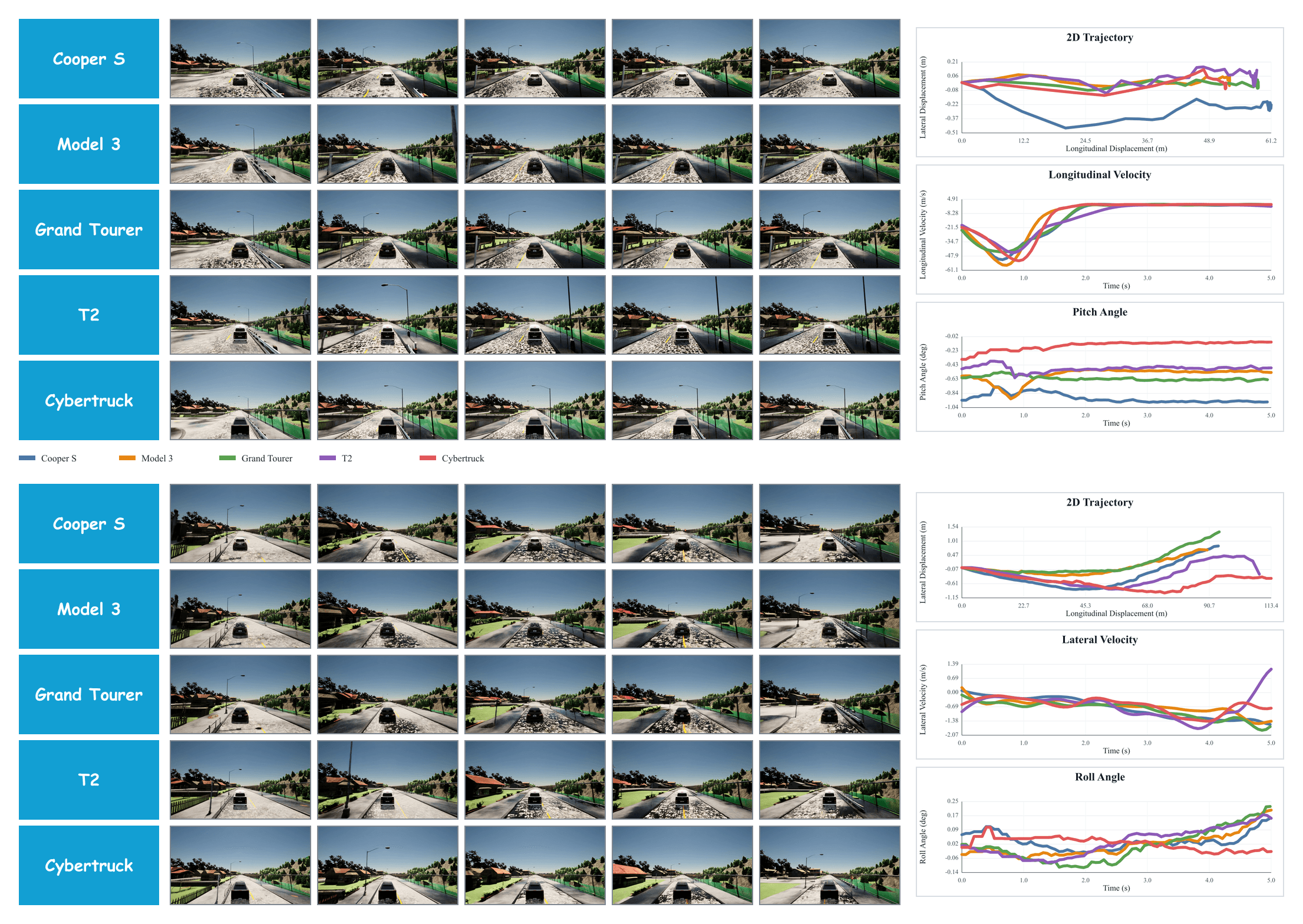}
    \caption{Qualitative vehicle-type sweep: Wan 2.2-I2V. Top: Emergency-braking scenario; Bottom: Single-lane-changing scenario.}
    \label{fig:app-vehicle-wan}
\end{figure*}

\begin{figure*}[p]
    \centering
    \includegraphics[width=\textwidth]{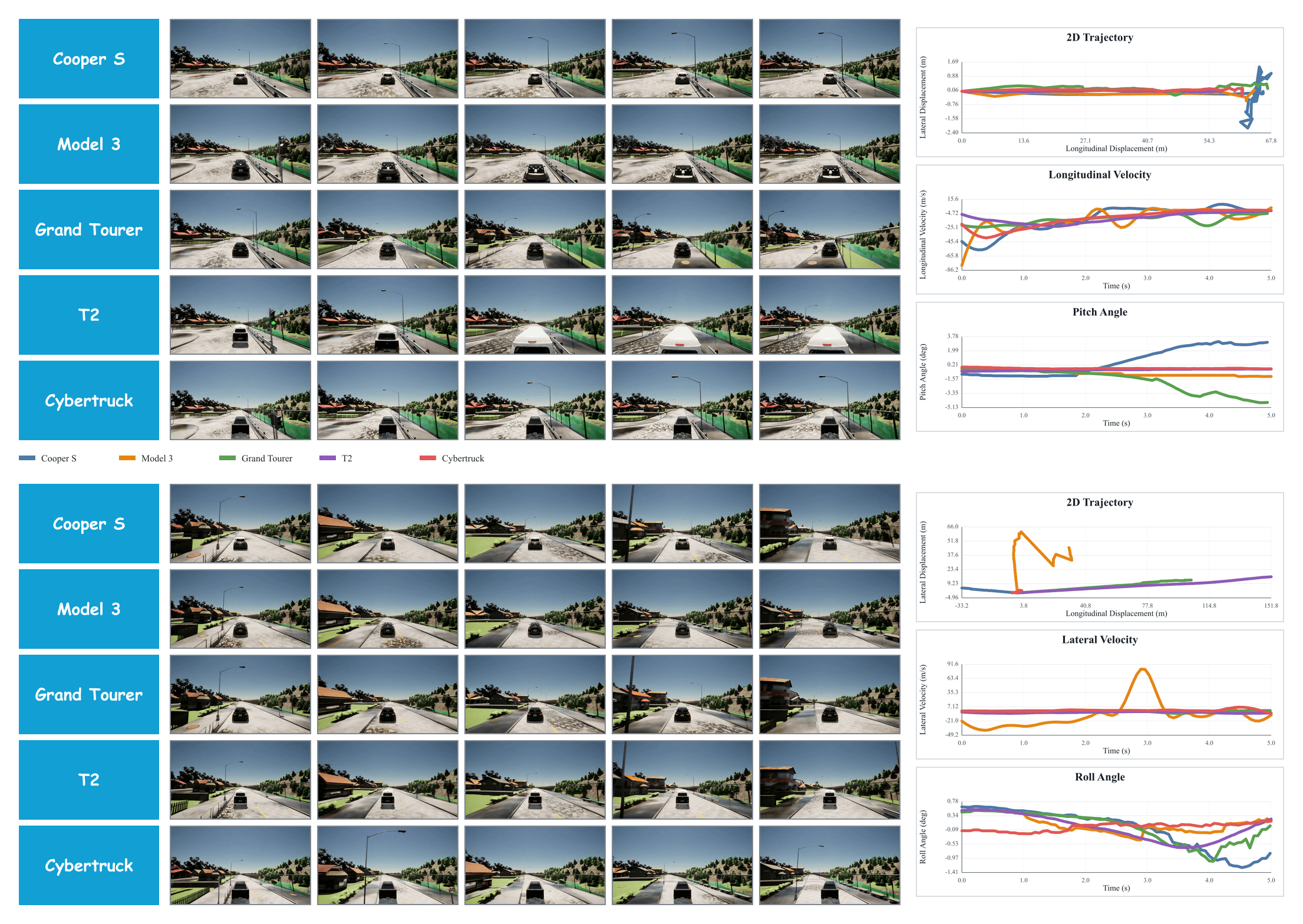}
    \caption{Qualitative vehicle-type sweep: HunyuanVideo 1.5-I2V. Top: Emergency-braking scenario; Bottom: Single-lane-changing scenario.}
    \label{fig:app-vehicle-hunyuan}
\end{figure*}

\begin{figure*}[p]
    \centering
    \includegraphics[width=\textwidth]{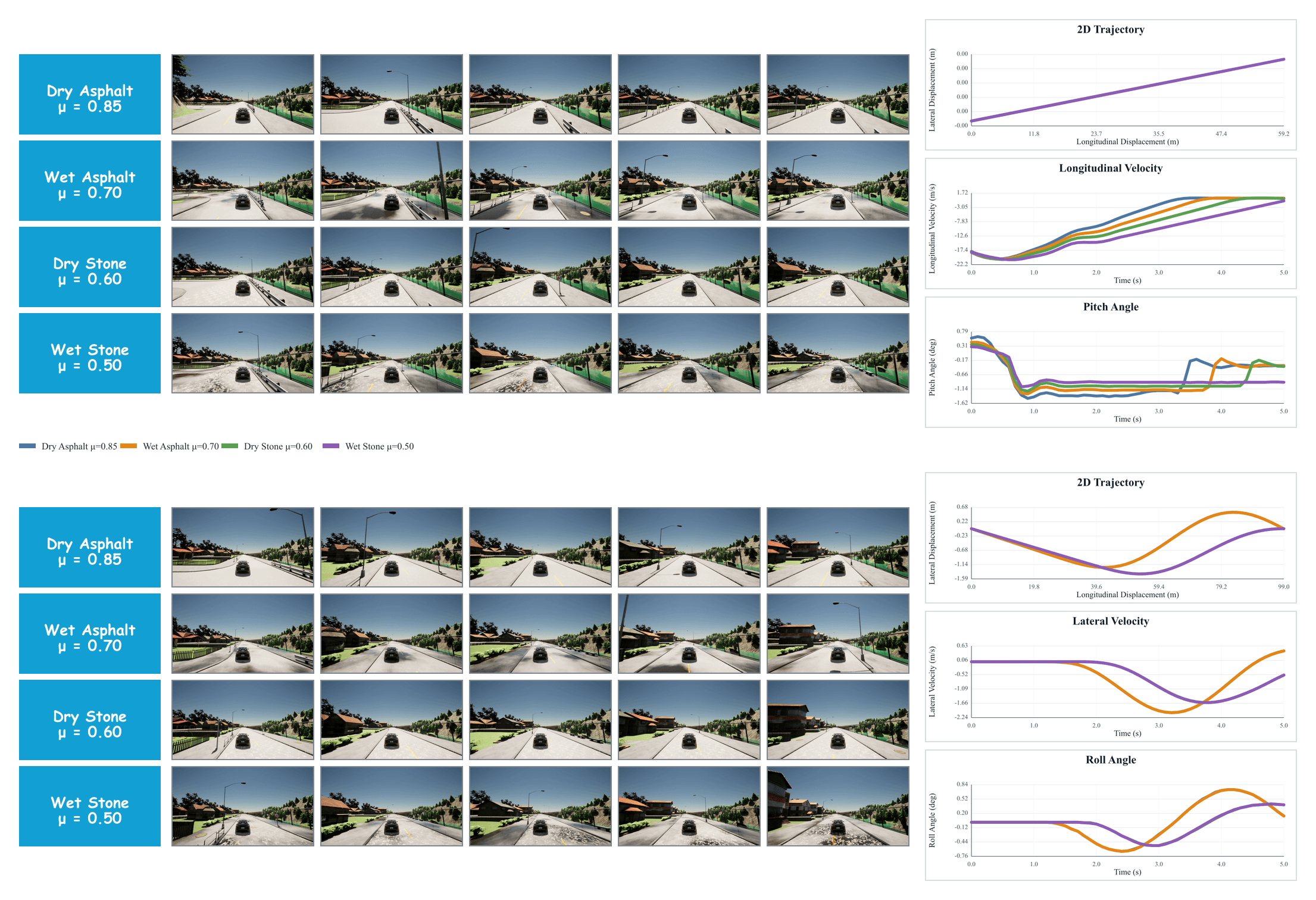}
    \caption{Qualitative tire-road-friction sweep: Ground Truth. Top: Emergency-braking scenario; Bottom: Single-lane-changing scenario.}
    \label{fig:app-friction-gt}
\end{figure*}

\begin{figure*}[p]
    \centering
    \includegraphics[width=\textwidth]{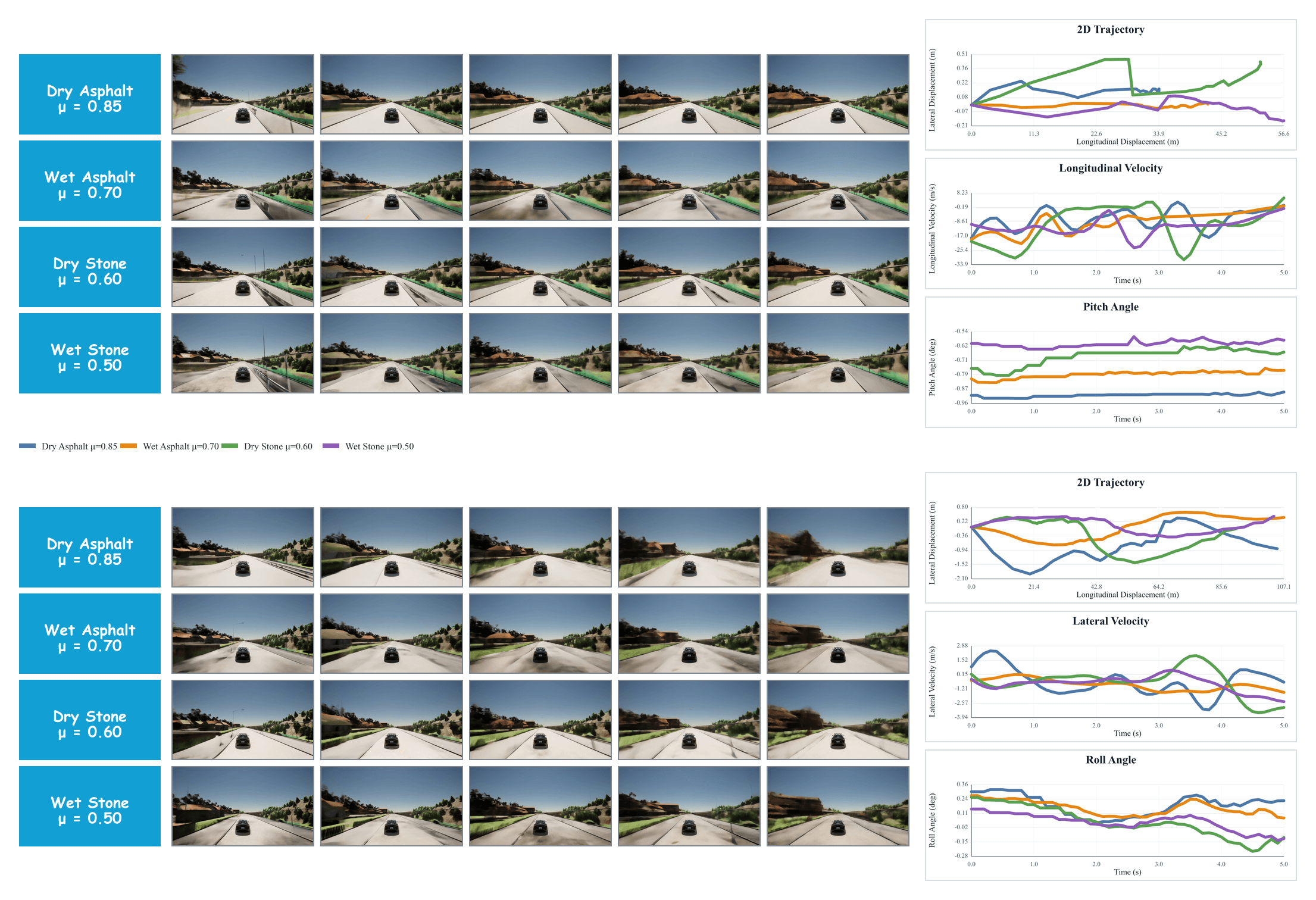}
    \caption{Qualitative tire-road-friction sweep: Vista. Top: Emergency-braking scenario; Bottom: Single-lane-changing scenario.}
    \label{fig:app-friction-vista}
\end{figure*}

\begin{figure*}[p]
    \centering
    \includegraphics[width=\textwidth]{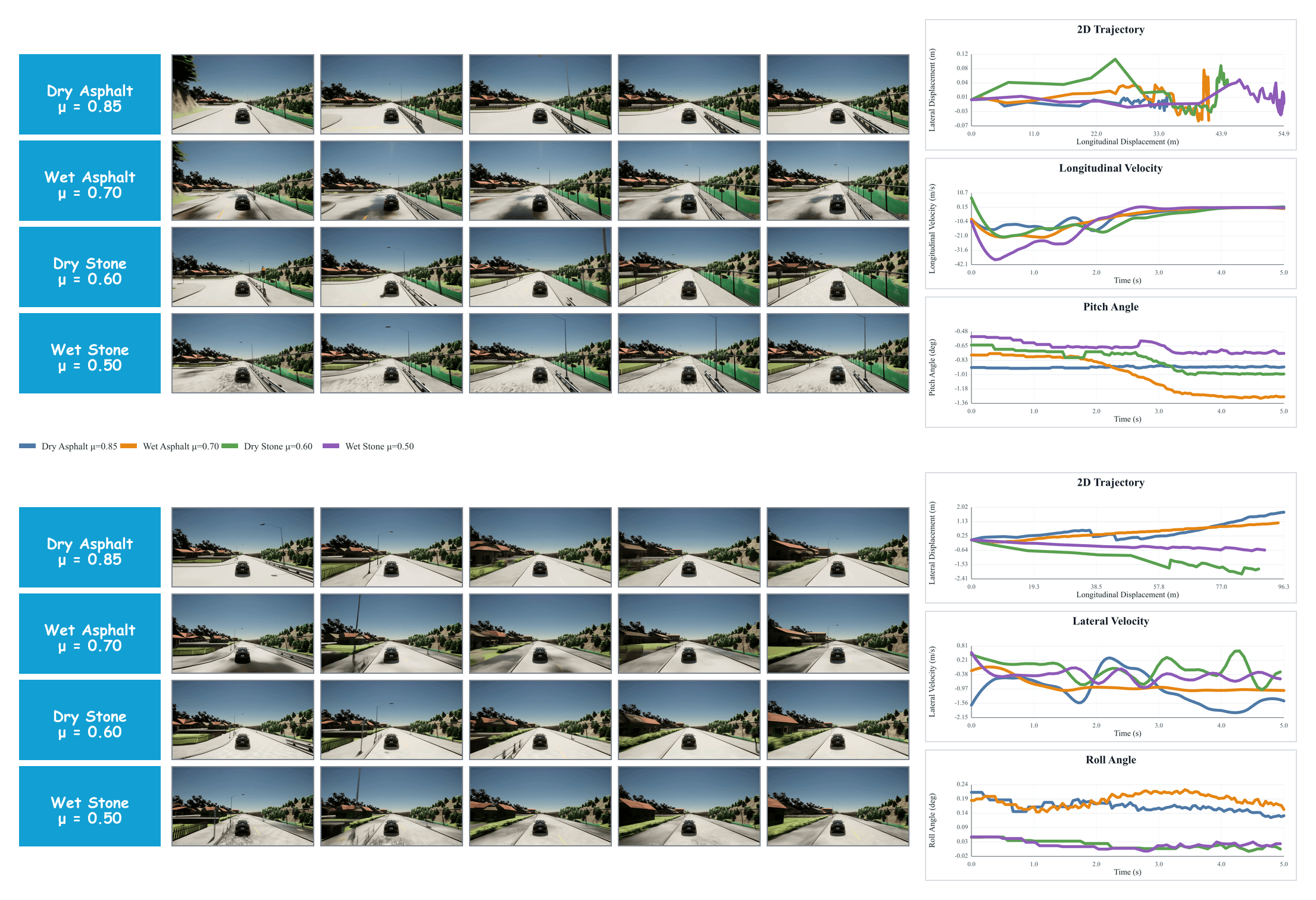}
    \caption{Qualitative tire-road-friction sweep: LTX-Video 2.5. Top: Emergency-braking scenario; Bottom: Single-lane-changing scenario.}
    \label{fig:app-friction-ltx}
\end{figure*}

\begin{figure*}[p]
    \centering
    \includegraphics[width=\textwidth]{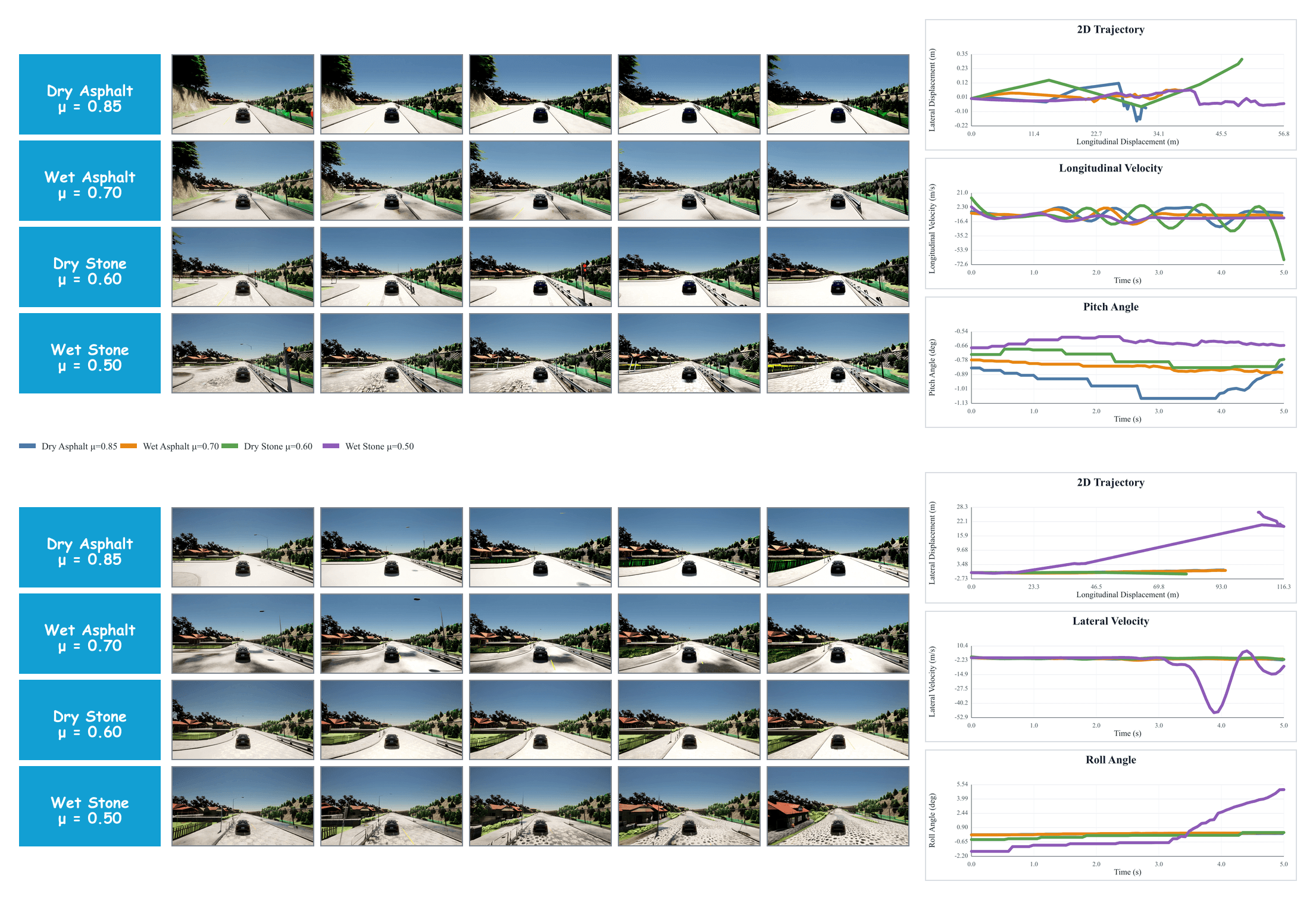}
    \caption{Qualitative tire-road-friction sweep: SkyReels-V2-I2V. Top: Emergency-braking scenario; Bottom: Single-lane-changing scenario.}
    \label{fig:app-friction-skyreels}
\end{figure*}

\begin{figure*}[p]
    \centering
    \includegraphics[width=\textwidth]{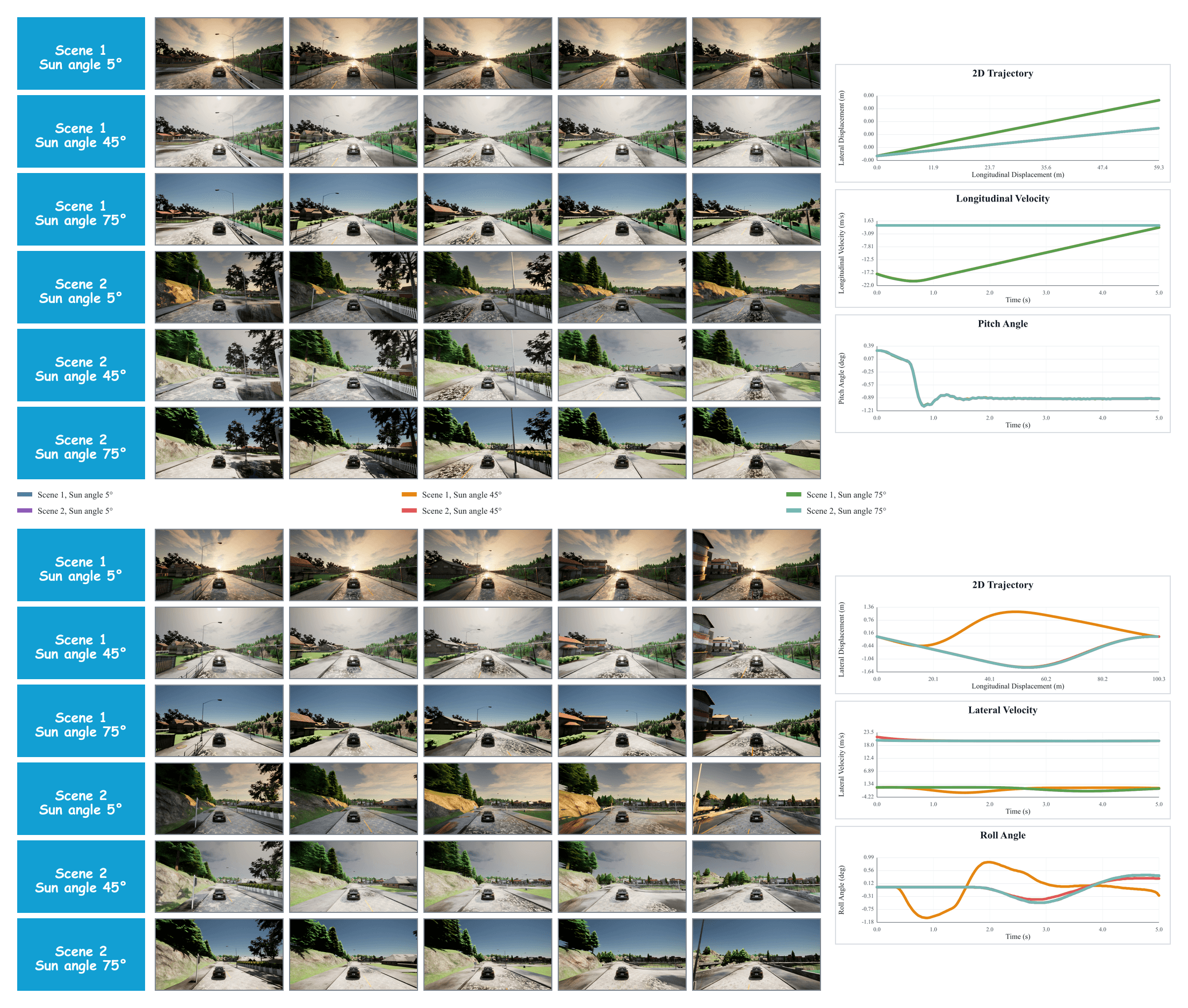}
    \caption{Qualitative environment sweep: Ground Truth. Top: Emergency-braking scenario; Bottom: Single-lane-changing scenario.}
    \label{fig:app-environment-gt}
\end{figure*}

\begin{figure*}[p]
    \centering
    \includegraphics[width=\textwidth]{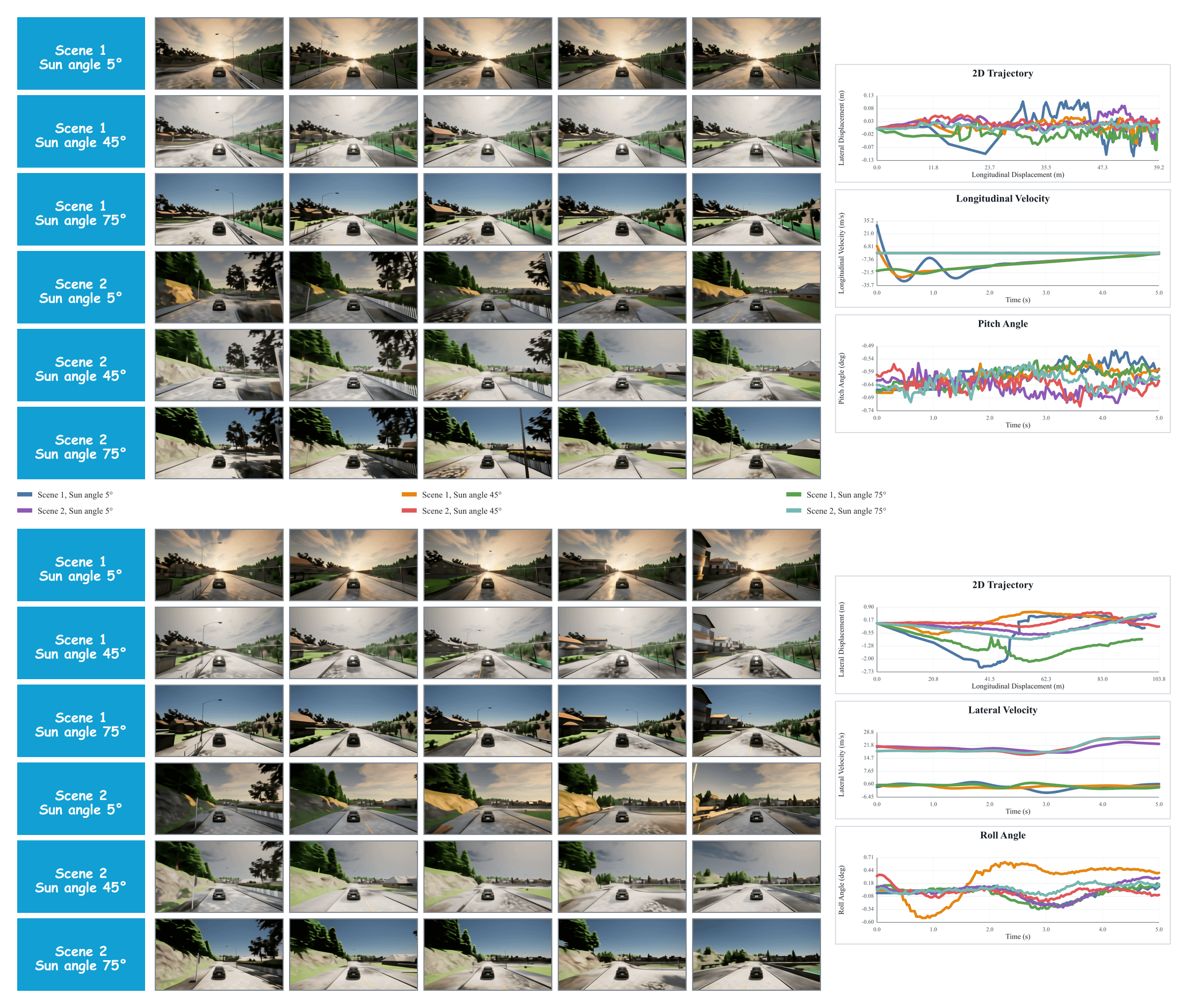}
    \caption{Qualitative environment sweep: DrivingWorld. Top: Emergency-braking scenario; Bottom: Single-lane-changing scenario.}
    \label{fig:app-environment-drivingworld}
\end{figure*}

\begin{figure*}[p]
    \centering
    \includegraphics[width=\textwidth]{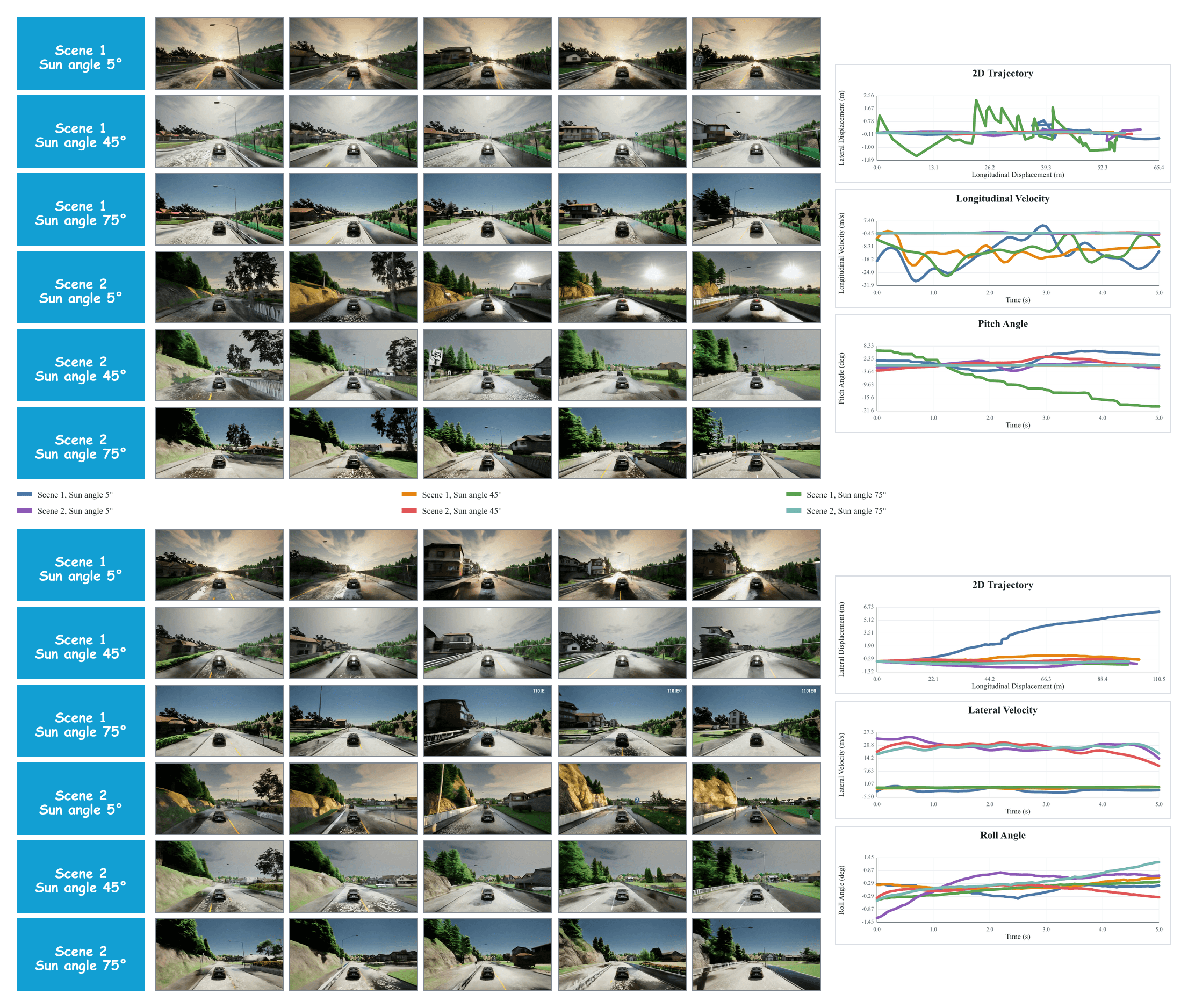}
    \caption{Qualitative environment sweep: CogVideoX 1.5. Top: Emergency-braking scenario; Bottom: Single-lane-changing scenario.}
    \label{fig:app-environment-cogvideox}
\end{figure*}

\begin{figure*}[p]
    \centering
    \includegraphics[width=\textwidth]{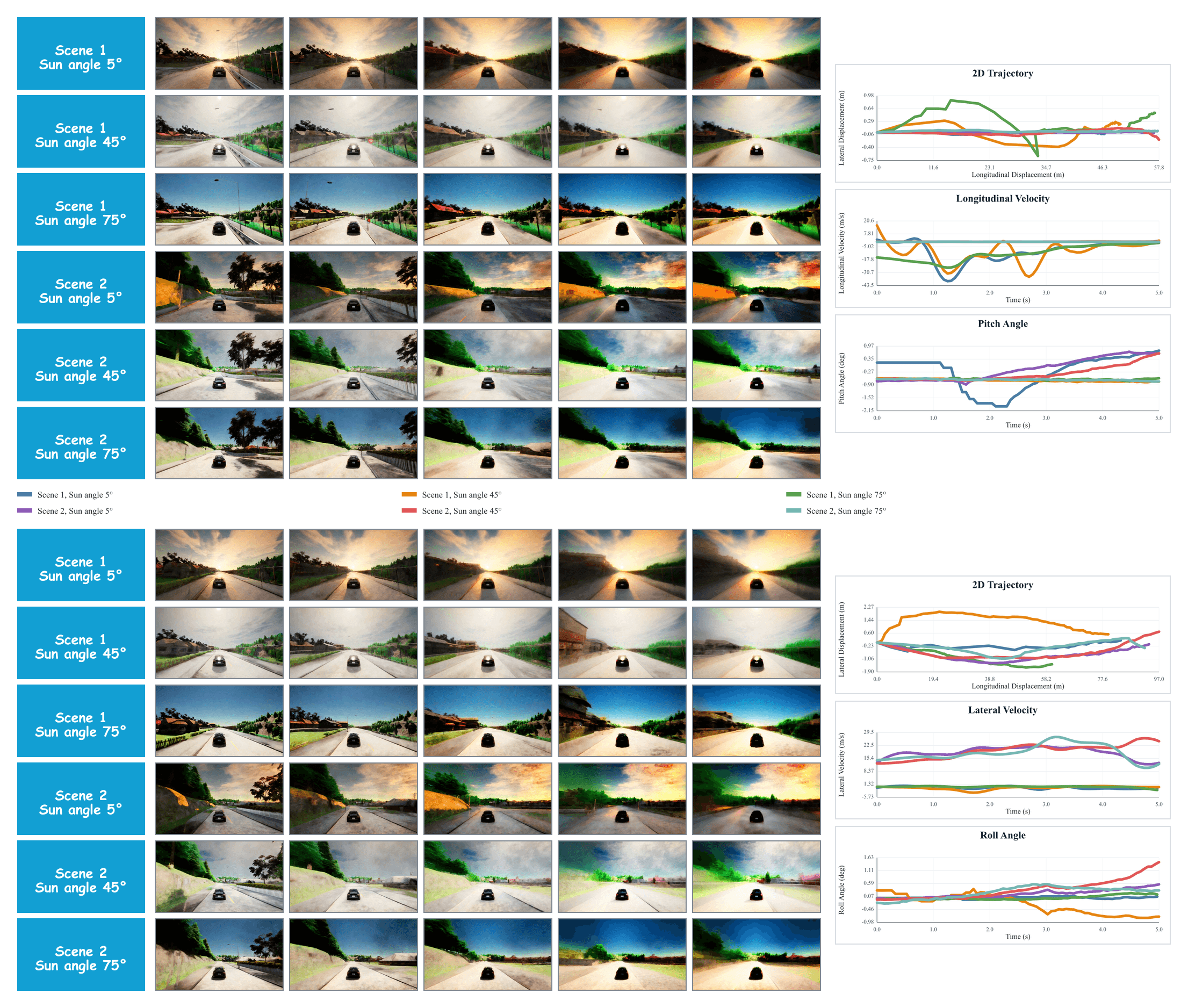}
    \caption{Qualitative environment sweep: GEM. Top: Emergency-braking scenario; Bottom: Single-lane-changing scenario.}
    \label{fig:app-environment-gem}
\end{figure*}

\subsubsection{Effect of the Target Speed}

The target-speed results in Figures \ref{fig:app-speed-gt}--\ref{fig:app-speed-cosmos-dreams} compare target speeds of $20$, $54$, $72$, and $108$~km/h. 
The ground truth shows distinct forward progress and braking transients across speeds, and higher-speed cases yield larger longitudinal displacements and more pronounced velocity changes. 
Lane-changing sequences similarly exhibit different lateral-motion and roll responses.
The model rollouts show that speed variation affects both visual quality and dynamic consistency. 
While lower-speed cases generally preserve the driving context, higher-speed rollouts more often accumulate appearance artifacts, lane-geometry drift, or inconsistent vehicle motion. 
The state curves reveal deviations in predicted deceleration and lateral response even when individual frames appear plausible.

\subsubsection{Effect of the Vehicle Type}

The vehicle-type results in Figures \ref{fig:app-vehicle-gt}--\ref{fig:app-vehicle-hunyuan} vary the ego vehicle among Cooper S, Model 3, Grand Tourer, T2, and Cybertruck. 
The ground-truth sequences maintain consistent road geometry and maneuver execution despite changes in vehicle scale and body shape.
Vehicle identity is a sensitive conditioning factor for the models, which must preserve vehicle shape, lane position, and scene context simultaneously. 
Some rollouts retain the intended vehicle appearance but show minor motion or camera inconsistency.
Others gradually alter vehicle geometry or develop visible artifacts. 
The state curves indicate whether these visual failures are accompanied by implausible motion predictions.

\subsubsection{Effect of the Tire-Road Friction Coefficient}

The tire-road-friction results in Figures \ref{fig:app-friction-gt}--\ref{fig:app-friction-skyreels} consider dry asphalt ($\mu=0.85$), wet asphalt ($\mu=0.70$), dry stone ($\mu=0.60$), and wet stone ($\mu=0.50$). 
The tire-road friction coefficient is explicitly provided as a conditioning input to each world model, together with the corresponding road condition. 
The ground truth provides a reference for the resulting braking and lane-changing dynamics.
This sweep evaluates whether the models can follow the friction-conditioned input in both visual generation and motion prediction. 
A successful rollout should reflect the specified surface condition while producing state trajectories consistent with its physical effect. 
The joint video-state presentation reveals whether a model uses the friction input coherently, rather than merely producing plausible road textures or generic vehicle motion.

\subsubsection{Effect of the Environment}

The environmental results in Figures \ref{fig:app-environment-gt}--\ref{fig:app-environment-gem} change the scene layout and sun angle, using $5^\circ$, $45^\circ$, and $75^\circ$ illumination. 
These settings alter shadows, brightness, contrast, and background structure while retaining the same maneuver.
Environmental variation primarily stresses long-horizon visual consistency. 
The generated sequences may exhibit exposure changes, unstable shadows, texture artifacts, or scene-geometry drift. 
The state curves reveal whether these visual deviations remain independent of the predicted motion or lead to trajectory and attitude errors, highlighting the need to jointly model appearance and ego dynamics across diverse environments.

\subsection{Details about Human Alignment}
\label{appendix:human}

\paragraph{Participants and ethics.}
We recruited 50 participants through internal volunteer sign-up.
All participants had normal vision and held a valid driving license, and none had been exposed to the evaluated samples or to the design of the VehDyn metrics.
The study was conducted in accordance with institutional guidelines, involved only the rating of pre-generated driving videos, and collected no sensitive personal information.
Participants were not informed of the world model, the post-training condition, or any automatic evaluation score associated with any sample.

\paragraph{Interface and rating procedure.}
Participants used a web-based interface that presented, for each item, the ground-truth clip and the generated clip side by side at the same resolution and playback rate, synchronized on a shared timeline with frame-by-frame scrubbing.
Beneath the two clips, the interface displayed the scenario labels shared by both (vehicle type, target speed, maneuver, and road condition) and, for the dynamic-consistency judgment only, a second row showing the ground-truth and generated clips of the matched sample that differs from the first in exactly one physical factor (target speed or tire-road friction coefficient).
Recovered state trajectories, automatic scores, and model identities were never shown.
The presentation order of items was randomized independently for each participant, and left/right placement of the two clips was fixed (ground truth left) so that participants could anchor on the reference.
For each item, participants assigned three scores on a [0,1] scale, one for each evaluation level.

\paragraph{Sample selection.}
Human evaluation covered all 12 world models and the three maneuvers.
Items were drawn from the test scenes by stratified sampling that balanced world model, maneuver family, vehicle type, road condition, and target speed, and that spanned low, medium, and high VehDyn-score ranges so that the correlation would not be inflated by rating only clearly successful or clearly failed generations.
In total 1,008 generated clips were rated, each by five participants, giving 5,040 ratings.
For the dynamic-consistency judgment, each item was paired with its matched sample along one randomly chosen factor (speed or friction), and both clips of the pair were rated in the same session.
Ground-truth clips were also included as items, paired with themselves, and were not identified as ground truth.

\paragraph{Scoring rubric.}
Participants rated each level according to the anchors in Table \ref{tab:human_rubric}.
The rubric asks for judgments that a viewer can make from the videos alone: whether the vehicle ends up where the reference vehicle does (trajectory alignment), whether it moves the way the reference vehicle moves along the way (kinematic consistency), and whether the generated motion changes between the two clips of a matched pair in the same direction and by a comparable amount as the reference motion changes (dynamic consistency).
The three levels were rated independently: a clip may follow the reference path closely while its body attitude is implausible, or reproduce one clip faithfully while failing to respond to the changed condition.

\begin{table}[ht!]
\centering
\caption{Scoring rubric for the human study. Each level is rated on $[0,1]$ in increments of $0.1$. TA: trajectory alignment; KC: kinematic consistency; DC: dynamic consistency.}
\label{tab:human_rubric}
\small
\renewcommand{\arraystretch}{1.15}
\setlength{\tabcolsep}{4pt}
\begin{tabularx}{\textwidth}{@{}c >{\raggedright\arraybackslash}X >{\raggedright\arraybackslash}X >{\raggedright\arraybackslash}X@{}}
\toprule
\textbf{Score} & \textbf{TA: does the vehicle follow the reference path?} & \textbf{KC: does the vehicle move like the reference vehicle?} & \textbf{DC: does the motion respond to the changed condition like the reference?} \\
\midrule
1.0 & Same lane position and heading throughout; stops (braking) or settles in the new lane (lane-changing) at the same point as the reference, within about one vehicle length. & Deceleration or lateral shift begins and ends at the same times as the reference; nose dip, body roll, and heading change match in magnitude and timing; no jitter or implausible sway. & Between the two clips, the generated braking distance, lane-changing duration, and body attitude change in the same direction as the reference and by a visibly comparable amount. \\
\addlinespace[2pt]
0.8--0.9 & Same path with a small lateral offset, or a stopping/settling point one to two vehicle lengths from the reference. & Correct sequence and magnitude of motion with a small timing offset, or a slightly under- or over-stated attitude response. & Correct direction of change for every response; the magnitude of the main response is somewhat under- or over-stated. \\
\addlinespace[2pt]
0.6--0.7 & Maneuver completed on a recognizably similar path, but with a lateral drift of up to about half a lane width or a stop two to four vehicle lengths early or late. & Maneuver recognizable and correctly ordered, but the speed profile or one attitude response is clearly wrong in magnitude. & Correct direction of change for the main response, but the magnitude is clearly wrong or a secondary response changes in the wrong direction. \\
\addlinespace[2pt]
0.4--0.5 & Maneuver completed, but the path drifts by about a lane width or the stop is several vehicle lengths early or late. & Maneuver recognizable, but the motion is visibly jerky, or two or more responses are wrong in magnitude or timing. & The response changes in the right direction for some quantities and not for others, or in the right direction but with negligible magnitude. \\
\addlinespace[2pt]
0.2--0.3 & Maneuver only partly executed (incomplete lane-changing, braking that does not stop the vehicle), or the path diverges over most of the clip. & Motion largely inconsistent with the reference (no deceleration where the reference brakes hard; roll or heading changes that the reference does not show). & The two generated clips are nearly identical although the reference clips differ clearly, or the change is in the wrong direction for the main response. \\
\addlinespace[2pt]
0.0--0.1 & Wrong maneuver, vehicle leaves the road, or the ego vehicle is not identifiable. & Motion physically implausible (teleportation, reversal, the vehicle body deforms) or absent. & The generated response changes in the opposite direction to the reference for the main response, or the clips are unusable for comparison. \\
\bottomrule
\end{tabularx}
\end{table}

\paragraph{Rating rules.}
Participants completed a written tutorial and three practice items with author-assigned calibration scores before the formal session.
The tutorial set five rules.
\begin{itemize}
\item The three levels are rated independently; one impression must not set all three scores.
\item Ratings concern the ego vehicle's motion only; rendering artifacts, background inconsistency, and image quality are ignored unless they make the vehicle's motion impossible to judge, in which case the item is flagged rather than scored.
\item Judgments are made over the whole clip rather than from the final frame; a lane change that reaches the correct lane by an implausible path does not receive full trajectory credit.
\item For dynamic consistency, the reference pair defines what a correct change looks like; the participant compares the change between the generated clips with the change between the reference clips, not either generated clip with its own reference.
\item The full scale is to be used, and when torn between two adjacent scores the lower one is chosen if any clear discrepancy remains.
\end{itemize}

\paragraph{Procedure per item.}
For each item, participants (i) played the ground-truth clip once to fix the reference maneuver, (ii) played the generated clip, using synchronized scrubbing where needed, and assigned the trajectory-alignment and kinematic-consistency scores, (iii) for the matched pair, played the two reference clips, then the two generated clips, and assigned the dynamic-consistency score, and (iv) optionally entered a one-line note for any score of 0.
The interface enforced that all three scores were entered before the next item was shown.

\paragraph{Data collection and aggregation.}
For each item and participant we recorded the anonymized item identifier, the three scores, the completion time, and any note.
The human score of item $i$ at level $\ell$ is the mean over its $K_i$ valid ratings,
\begin{equation}
H_{i,\ell} = \frac{1}{K_i}\sum_{k=1}^{K_i} h_{ik,\ell}, \qquad \ell \in \{\mathrm{TA}, \mathrm{KC}, \mathrm{DC}\},
\end{equation}
and the human score of model $m$ at level $\ell$ is the mean of $H_{i,\ell}$ over the items generated by $m$, computed per maneuver family and macro-averaged with equal weight, mirroring the aggregation of the automatic level scores.
The overall human score is the sum of the three level scores, on the same $[0,3]$ scale as the VehDyn score.

\paragraph{Analysis.}
Agreement between automatic and human scores was measured with Pearson correlation, Spearman rank correlation, the coefficient of determination $R^2$ of a least-squares fit, and the mean absolute error of that fit, at the model level for the four scores reported in the main text (Figure~\ref{fig4}(b)).
Because the automatic and human scores share the same $[0,1]$ scale at each level, the line $y=x$ is a meaningful reference and is drawn in the scatter plot.

\end{document}